\documentclass[twocolumn]{article}

\usepackage[margin=1in]{geometry}
\usepackage[utf8]{inputenc} 
\usepackage[T1]{fontenc}    
\usepackage{hyperref}       
\usepackage{url}            
\usepackage{booktabs}       
\usepackage{amsfonts}       
\usepackage{nicefrac}       
\usepackage{microtype}      
\usepackage{xcolor}         

\usepackage{enumitem}
\usepackage{enumerate}
\setlist[itemize]{leftmargin=0.75em}

\usepackage{color}
\usepackage{float}
\usepackage[style=base]{caption}
\usepackage{subcaption}
\usepackage{tikz}
\usepackage{tikz-cd}
\usepackage{pgfplots}
\usepackage{abstract}
\usepgfplotslibrary{groupplots}
\pgfplotsset{compat=1.18}

\usepackage{amsmath,amssymb,amsthm,mathtools,bbm}
\usepackage{physics}
\usepackage{bbold}
\usepackage{dsfont}
\usepackage{mathrsfs}
\usepackage[ruled,vlined,linesnumbered]{algorithm2e}
\usepackage[capitalize]{cleveref}
\usepackage{xfrac}
\usepackage{graphicx}
\usepackage{authblk}
\usepackage{natbib}

\usepackage{minitoc}

\DeclareMathOperator*{\erfc}{erfc}
\DeclareMathOperator*{\argmin}{arg\,min}

\newcommand{\sign}{\ensuremath{\operatorname{sign}}}

\newcommand{\advtrainingcost}{\ensuremath{\varepsilon_t}}
\newcommand{\advgencost}{\ensuremath{\varepsilon_g}}

\newcommand{\generr}{\ensuremath{E_{\mathrm{gen}}}}
\newcommand{\bounderr}{\ensuremath{E_{\mathrm{bnd}}}}
\newcommand{\trainerr}{\ensuremath{E_{\mathrm{train}}}}
\newcommand{\trainloss}{\ensuremath{\ell_{\mathrm{train}}}}
\newcommand{\advgenerr}{\ensuremath{E_{\mathrm{adv}}}}
\newcommand{\classpresgenerr}{\ensuremath{E_{\mathrm{CP}}}}
\newcommand{\usefulmetric}{\ensuremath{\mathcal{U}}}
\newcommand{\robustmetric}{\ensuremath{\mathcal{R}}}

\newcommand{\Sigmaw}{\ensuremath{\boldsymbol{\Sigma}_{\boldsymbol{w}}}}
\newcommand{\Sigmax}{\ensuremath{\boldsymbol{\Sigma}_{\boldsymbol{\boldsymbol{x}}}}}
\newcommand{\Sigmadelta}{\ensuremath{\boldsymbol{\Sigma}_{\boldsymbol{\delta}}}}
\newcommand{\Sigmaupsilon}{\ensuremath{\boldsymbol{\Sigma}_{\boldsymbol{\upsilon}}}}
\newcommand{\Sigmatheta}{\ensuremath{\boldsymbol{\Sigma}_{\boldsymbol{\theta}}}}

\newcommand{\w}{\ensuremath{\boldsymbol{\theta}}}
\newcommand{\x}{\ensuremath{\boldsymbol{x}}}
\newcommand{\wavec}{\ensuremath{\boldsymbol{\theta}^a}}
\newcommand{\what}{\ensuremath{\hat{\boldsymbol{\theta}}}}
\newcommand{\wstar}{\ensuremath{\boldsymbol{\theta}_0}}
\newcommand{\vecdelta}{\ensuremath{\boldsymbol{\delta}}}
\newcommand{\singleattack}{\ensuremath{\boldsymbol{v}}}

\newcommand{\datidx}{\ensuremath{i}}
\newcommand{\blockidx}{\ensuremath{\ell}}

\newcommand{\proxim}[3]{\ensuremath{\mathcal{P}_{#1}\qty[#2]\qty(#3)}}
\newcommand{\Dproxim}[3]{\ensuremath{\mathcal{P}^\prime_{#1}\qty[#2]\qty(#3)}}

\newcommand{\gaussdist}[2]{\ensuremath{\mathcal{N}\qty(#1, #2)}}

\newcommand{\EEb}[2]{\ensuremath{\mathbb{E}_{#1}\qty[#2]}}
\newcommand{\RR}{\ensuremath{\mathbb{R}}}

\newcommand{\Egen}{\ensuremath{E_{\mathrm{gen}}}}

\theoremstyle{plain}
\newtheorem{theorem}{Theorem}[section]
\newtheorem{proposition}[theorem]{Proposition}

\theoremstyle{definition}

\theoremstyle{remark}

\author[1]{Kasimir Tanner}
\author[1]{Matteo Vilucchio}
\author[2]{Bruno Loureiro}
\author[1]{Florent Krzakala}

\affil[1]{Information Learning and Physics Laboratory, \'Ecole Polytechnique F\'ed\'erale de Lausanne (EPFL)}
\affil[2]{Departement d'Informatique, \'Ecole Normale Sup\'erieure, PSL \& CNRS}

\title{A High Dimensional Statistical Model for Adversarial Training: Geometry and Trade-Offs}

\begin{document}
\maketitle
\begin{abstract}
This work investigates adversarial training in the context of margin-based linear classifiers in the high-dimensional regime where the dimension $d$ and the number of data points $n$ diverge with a fixed ratio $\alpha = n / d$. 
We introduce a tractable mathematical model where the interplay between the data and adversarial attacker geometries can be studied, while capturing the core phenomenology observed in the adversarial robustness literature. 
Our main theoretical contribution is an exact asymptotic description of the sufficient statistics for the adversarial empirical risk minimiser, under generic convex and non-increasing losses for a Block Feature Model. 
Our result allow us to precisely characterise which directions in the data are associated with a higher generalisation/robustness trade-off, as defined by a robustness and a usefulness metric. 
We show that the the presence of multiple different feature types is crucial to the high sample complexity performances of adversarial training.
In particular, we unveil the existence of directions which can be defended without penalising accuracy. 
Finally, we show the advantage of defending non-robust features during training, identifying a uniform protection as an inherently effective defence mechanism.
\end{abstract}

\section{INTRODUCTION}
 
The susceptibility of machine learning models to adversarial attacks --- subtle yet strategically crafted data perturbations --- has been an ongoing concern for various machine learning models, from linear classifiers to deep neural networks. 
In particular, this vulnerability is intrinsic to margin-based classifiers \citep{ilyas_adversarial_2019}.
Also in image classification, seemingly innocuous modifications, like tiny stickers on road signs, can dramatically mislead models that otherwise exhibit strong generalisation \citep{PavlitskaLambingZoellner2023_1000163579}.

The problem has been theoretically studied in linear models where solutions could be obtained analytically. Previous studies were focusing on the fundamental limits of the trade-off between adversarial and generalisation errors \citep{javanmard_precise_tradeoffs_2020} or on including a non trivial data covariance and comparing the performances of adversarial training to Bayes optimal errors \citep{taheri_asymptotic_2021}. 
 
Nonetheless a comprehensive understanding of how structure defines feature types --- robust, useful, or both --- affect model performance is still developing. 
The interplay of the data structure with attack and training geometries remains a particularly fertile ground for theoretical investigation, with potential implications 
for enhancing adversarial robustness and developing more effective protection methods. 

In this paper, we introduce a structured, high-dimensional model for studying adversarial classification under margin-based classifiers. Our \textbf{main contributions} are fourfold:
\begin{itemize}[noitemsep]
    \item We introduce a mathematically tractable model for investigating the interplay between the data, attack and defence geometries. Despite its simplicity, we show our model is rich enough to capture the key phenomenology observed in practical adversarial training setups.
    \item 
    We show that, in the high dimensional proportional limit (where the number of samples and covariate dimension diverge at fixed ratio), the relevant statistical properties of the adversarial empirical risk minimiser can be exactly characterised by a finite set of sufficient statistics.
    \item 
    Leveraging our high-dimensional characterisation, we are able to show the importance of distinguishing different features to 
    have
    different performances in the large sample complexity regime: 
    considering single block models leads to the same performances 
    for any kind of adversarial training.
    Additionally we also derive specific conditions under which defending non-robust features is beneficial in this prototypical data model.
    \item 
    Finally, building on our findings we investigate the interplay between data and attack geometry in the effectiveness of adversarial training. 
    In particular, we show how attack geometry's direction can be divided in two groups: directions leading to a 
    trade-off and directions that can be successfully defended without sacrificing accuracy. 
\end{itemize}
This manuscript is organised as follows. \cref{sec:setting-specification} introduces the data model and the margin-based adversarial training protocol. \cref{sec:technical-result} describes our main theoretical results, namely the asymptotic characterisation of the linear adversarial estimator. \cref{sec:tradeoffs-large-sample-complexity} discusses the implications of the main theoretical formulas. Finally, \cref{sec:additional-explorations} reports on the additional experiments. 
The code used to produce all the figures in this manuscript can be found at \href{https://github.com/IdePHICS/Adversarial-Setting}{github.com/IdePHICS/Adversarial-Setting}.

\subsection*{Related works}\label{sec:related-works}
\paragraph{Adversarial attacks ---} In the study of neural networks, the vulnerability to adversarial attacks is well-established, with early works like \citep{szegedy_intriguing_2014,goodfellow_explaining_2015,papernot2016limitations} uncovering this intriguing weakness. Adversarial Training, particularly through methods like projected gradient descent, has emerged as a leading defence strategy, as explored in-depth by \citet{madry_towards_2019}.

Understanding adversarial robustness has been a long standing challenge. \citet{schmidt_adversarially_2018} highlighted the necessity of reaching higher sample complexity in adversarial training to achieve the same generalisation performance as standard training.
\citet{ilyas_adversarial_2019} proposed that datasets contain predictive yet imperceptible features vulnerable to attacks. Additionally, \citet{tsilivis_can_2023} notices the difficulty of optimising a data-set to improve adversarial robustness. The adversarial setting has also been studied in the neural tangent kernel regime by \citet{tsilivis_what_2023}. 

The idea of a fundamental trade-off between adversarial robustness and standard accuracy has been noted in \cite{tsipras_robustness_2019,zhang2019theoretically,suggala2019revisiting}. 
Later, \citet{bhagoji2019lower,dan2020sharp,javanmard_precise_statistical_2022} have examined this trade-off in the case of Gaussian Mixture Models.
\citet{raghunathan2020understanding} show that a class of augmented estimators can have a worse generalisation error than the standard estimator. 
\citet{taheri_asymptotic_2021,javanmard_precise_tradeoffs_2020} contribute to this discourse by examining the impact of adversarial training on the interpolation threshold and double descent in the case of Gaussian data.
More recently \citet{roth_adversarial_2020,ribeiro2023regularization} have studied adversarial training as a form of data dependent regularisation.
The trade-off between robust and clean generalisation error that we are going to consider in this paper has been found fundamental in the design of algorithms such as TRADES \citep{zhang2019theoretically} and ARoW Regularisation \citep{yang_improving_2023}.

\paragraph{Exact asymptotics ---} Our main theoretical result pertains to an established literature employing techniques from high-dimensional probability \citep{thrampoulidis2015gaussian,pmlr-v40-Thrampoulidis15, doi:10.1073/pnas.1810420116, Dhifallah2020}, random matrix theory \citep{doi:10.1073/pnas.1307845110, 8683376, NEURIPS2020_a03fa308, mei_generalization_2022, xiao2022precise, pmlr-v202-schroder23a} and statistical physics \citep{aubin_generalization_2020, mignacco2020role, Gerace_2021, bordelon20a, vilucchio2024asymptotic, pmlr-v206-okajima23a, adomaityte_classification_2023, adomaityte2023highdimensional} to derive exact asymptotic results of high-dimensional statistical estimation problems. Of particular relevance to our work is \citet{Loureiro_2022}, who proved a formula for the sufficient statistics of general Gaussian Covariate models. While our work leverages their results, our formulas are more general, as they account for adversarial attacks and training. Moreover, in our proof we use a mapping to an Approximate Message Passing (AMP) for adversarial training on structured problems, which builds upon \citet{rangan_GAMP_2011, javanmard_montanari_gamp_2013, 9745052, Loureiro_gaussian_mixtures_2021, pmlr-v162-loureiro22a, 10.1093/imaiai/iaad020}.

The present work extends these previous analyses in three fundamental aspects. First, unlike the unstructured Gaussian setting of \citet{javanmard_precise_tradeoffs_2020}, our framework explicitly captures feature-dependent robustness through the Block Feature Model, allowing us to study how different features contribute to adversarial vulnerability. Second, we generalise the attack models of \citet{taheri_asymptotic_2021} by introducing structured attacks and defenses through arbitrary positive definite matrices, providing a more complete picture of adversarial training. Third, while \citet{zhang2019theoretically} provided finite-dimensional bounds, our analysis yields exact asymptotics in high dimensions.

\section{SETTING SPECIFICATION}
\label{sec:setting-specification}

Consider a supervised binary classification problem with training data \(\mathcal{D} = \qty{(\x_\datidx, y_\datidx)}_{\datidx=1}^{n}\in\mathbb{R}^{d}\times\{-1,+1\}\). In the following, we focus on a generalised estimation problem, where for each $i=1,\cdots,n$ we assume the covariates are independently drawn from \(\x_\datidx \sim \gaussdist{\boldsymbol{0}}{\Sigmax}\) and with labels 
\(y_{i} \sim \mathbb{P}(y|\wstar^{\top}\x_{i})\)
for a fixed parameter $\wstar\in\mathbb{R}^{d}$.\footnote{Note our results also hold under the assumption of \(\wstar \sim \gaussdist{\boldsymbol{0}}{\Sigmatheta}\).} Although our theoretical results in \cref{sec:technical-result} hold under a generic likelihood \(P(y|z)\), for concreteness the discussion in \cref{sec:tradeoffs-large-sample-complexity} will be mostly focused on the probit model \(\mathbb{P}\left(y|z\right) = \sfrac{1}{2}\erfc\left(-\sfrac{z}{\sqrt{2}\tau}\right)\), where the parameter $\tau > 0$ controls the noise level. In particular, note that for \(\tau\to 0\) we have \(\mathbb{P}(y|z) = \delta(y-\sign(z))\). 

Given the training data $\mathcal{D}$, our goal in the following is to investigate the capacity of margin-based linear classifiers \(\hat{y}(\what, \x) = \sign ( \what \cdot \x / \sqrt{d})\)
in robustly and efficiently classifying the data under adversarial attacks, where $\what = \what(\mathcal{D})$ is an estimator that has been learned from the training data. The setting introduced above is often refereed to as a \emph{teacher-student} setting, and is widespread in the high-dimensional statistics literature \citep{doi:10.1073/pnas.1810420116,taheri_asymptotic_2021,Clarte_uncertanty}. 

In the case where the data points are not attacked, the metric of interest is the (clean) \textit{generalisation error} defined as
\begin{equation}\label{eq:gen-error-definition}
    \generr = \EEb{y,\x}{
        \mathbb{1}(y \neq \hat{y} (\what, \x))
    } \,,
\end{equation}
where the expectation is taken over input-output pairs generated using the same \(\wstar\) as in training.

If the adversary can perturb the input sample \(\x_\datidx\) its objective is to find a perturbation \(\boldsymbol{\upsilon}_{\datidx}\) which leads to a wrongly classified sample, \textit{i.e.} \(y(\x_\datidx) \neq \hat{y}(\what, \x_\datidx + \boldsymbol{\upsilon}_\datidx)\).
We focus on allowed perturbations that must have a \(\Sigmaupsilon\)-induced norm smaller or equal than a fixed attack strength $\advgencost$, beyond which an attack could be identified --- \(\norm{\boldsymbol{\upsilon}}_{\Sigmaupsilon^{-1}} \leq \advgencost\).  
We will call $\Sigmaupsilon$ the \textit{attack matrix}. By normalising $\Sigmaupsilon$ we can interpret $\advgencost$ as the global strength of the attack or attack budget and $\Sigmaupsilon$ as the attack geometry.

This model generalises the case considered by \citet{javanmard_precise_tradeoffs_2020,taheri_asymptotic_2021} where they consider the cases of bounded attacks in \(\ell_2\) norm.

Given the previous adversarial constraints, 
we introduce the \textit{adversarial generalisation error}. This metric quantifies the student's performance under adversarial attack
\begin{equation}\label{eq:adv_gen_metric_definition}
    \advgenerr = \EEb{y,\x}{
        \max_{\norm{\vecdelta}_{\Sigmaupsilon^{-1}} \leq \advgencost} \mathbb{1}(y \neq \hat{y} (\what, \x + \vecdelta))
    } \,, 
\end{equation}
where the expectation is taken over pairs of input-output generated with the same \(\wstar\) used during training. 
Notably, this adversarial generalisation error is an extension of the standard generalisation error, with the latter being the special case where \(\generr = \advgenerr(\advgencost=0)\). 

As noted by \citet{zhang2019theoretically,yang_improving_2023} the adversarial generalisation error can be decomposed into a \(\advgenerr = \generr + \bounderr\), where we will refer to \(\bounderr\) as \textit{boundary error}. This metric measures the number of samples correctly classified but that are within attack range from the decision boundary. Explicitly one has 
\begin{equation}\label{eq:definition_adv_boundary_property}
    \bounderr 
    =  \EEb{y,\boldsymbol{x}}{ 
        {\textstyle \mathbb{1}(y = \hat{y}(\what ; \boldsymbol{x})) }
        \max_{\norm{\boldsymbol{\delta}}_{\Sigmaupsilon^{-1}} \leq \advgencost} 
        {\textstyle \mathbb{1}(y \neq \hat{y} (\what, \boldsymbol{x} + \boldsymbol{\delta}))}
    } .
\end{equation}

\subsection{Empirical Risk Minimisation and Adversarial Case}
To estimate the student vector that achieves the minimal adversarial error the most common way is to consider a convex surrogate empirical version of the adversarial error we are trying to minimise \citep{bach2024learning}.
We define the adversarial risk function as
\begin{equation}\label{eq:adversarial-training-problem}
     \sum_{\datidx = 1}^{n} 
    \max_{
        \norm{\vecdelta_\datidx}_{\Sigmadelta^{-1}} \leq \advtrainingcost 
    }
    g \qty(y_\datidx \frac{\w^\top \qty(\x_\datidx + \vecdelta_\datidx)}{\sqrt{d}}) 
    + r(\w) \,,
\end{equation}
where \(g\) is a convex loss, \(r(\w)\) is a convex regularisation term and \(\Sigmadelta\) is a positive definite matrix. We will call \(\Sigmadelta\) the \textit{defence matrix} and it will be normalised in the same way as \(\Sigmaupsilon\).

By considering any decreasing convex loss \(g\), we can simplify the inner maximisation problem leading to an equivalent form of the risk
\begin{equation}\label{eq:modified-minimisation-problem}
    \sum_{\datidx = 1}^{n} 
    g \qty(
        y_\datidx \frac{\w^\top \x_\datidx}{\sqrt{d}} 
        - \advtrainingcost \frac{\sqrt{\w^\top \Sigmadelta \w}}{\sqrt{d}} 
    ) 
    + r(\w) \,.
\end{equation}
Minimising the risk function provides an estimate \(\what(\mathcal{D})\) for the student weights.
While our framework is versatile enough to accommodate various convex regularisation functions
for the rest of the paper we set \(r(\w) = \frac{\lambda}{2} \norm{\w}_2^2\).
 
Our analysis will be carried out in the \textit{high-dimensional proportional limit}. Specifically, we investigate settings where both the dimension \(d\) and the number of training samples \(n\) are large \(d,n \to \infty\), whilst maintaining a fixed sample complexity \(\alpha := n / d\).

\subsection{Block Feature Data Model}\label{sec:strong-weak-feature-model}

In \cite{tsipras_robustness_2019,ilyas_adversarial_2019,tsilivis_what_2023} the discourse is centred on distinguishing between \textit{useful} and \textit{robust} features.
The usefulness of a feature is a measure of how much that specific feature correlates with the output that we want to predict. The robustness is a measure of the same correlation after an attack is performed on the data point.
The proposed view is that adversarial vulnerability increases when the classification is based on useful but non-robust features. 

In the context of linear models considered in this manuscript we define \textit{usefulness} \(\usefulmetric_{\wstar}\) and \textit{robustness} \(\robustmetric_{\wstar}\) respectively as
\begin{align}
    \usefulmetric_{\wstar}
    &= \frac{1}{\sqrt{d}} \mathbb{E}_{\x, y}[y \wstar^\top \x ] \,, \label{eq:mt-usefulness-def} \\
    \robustmetric_{\wstar} &= \frac{1}{\sqrt{d}} \EEb{\x, y}{
        \inf_{\norm{\vecdelta}_{\Sigmaupsilon^{-1}}  \leq \advgencost} y \wstar^\top ( \x + \vecdelta)
    }\,. \label{eq:mt-robustness-def}
\end{align}
These two metrics capture the relationship between the learning task ($\wstar$) and the data model, focusing on how informative are the features ($\usefulmetric_{\wstar}$) and how they remain informative under worst-case input perturbations ($\robustmetric_{\wstar}$).

Following the interpretation of linear models in \citet[Sec.~1.2]{hastie2022surprises} we see the importance of defining different types of features with different properties and introducing features that are more/less easily attack-able.

\paragraph{The Block Feature Model} 
While our theoretical results hold for a wide array of data models, we want to study adversarial robustness in a reductionist spirit. We introduce the \textit{Block Feature Model (BFM)}, which allows to systematically vary feature usefulness and robustness to understand their impact on (adversarial) generalisation metrics.

For a given dimension \(d\) we define \(k \leq d\) blocks with individual sub-dimensions \(\qty{d_\blockidx}_{\blockidx=1}^{k}\) satisfying \(\sum_{\blockidx=1}^{k} d_\blockidx = d\). 
This allows us to write the quantities of interest as 
\begin{equation}
\begin{aligned}
    \Sigmax &= \operatorname{blockdiag}\qty( \psi_1 \mathbb{1}_{d_1}, \dots, \psi_k \mathbb{1}_{d_k} ) \,, \\
    \Sigmadelta &= \operatorname{blockdiag}\qty( \Delta_1 \mathbb{1}_{d_1}, \dots, \Delta_k \mathbb{1}_{d_k} ) \,, \\
    \Sigmaupsilon &= \operatorname{blockdiag}\qty( \Upsilon_1 \mathbb{1}_{d_1},  \dots, \Upsilon_k \mathbb{1}_{d_k} ) \,, \\
    \Sigmatheta &= \operatorname{blockdiag}\qty( t_1 \mathbb{1}_{d_1}, \dots, t_k \mathbb{1}_{d_k} ) \,,
\end{aligned}
\end{equation}
where each \(\psi_\blockidx, \Delta_\blockidx, t_\blockidx\) and \(\Upsilon_\blockidx\) is greater than zero for each \(\blockidx=1,\dots k\) to preserve the positive definiteness.  Notice, this model can easily be extended to allow for power-law distributions on the eigenvalues by considering \(d_\blockidx = 1\) and \(\psi_\blockidx = \blockidx^{-\beta_\psi}\) for \(\beta_\psi > 1\). 

The parameters of the BFM have direct interpretations: \(\psi_\blockidx\) characterises the variance of features in block \(\blockidx\), \(\Delta_\blockidx\) describes how sensitive these features are to attacks, \(\Upsilon_\blockidx\) determines the defence strategy for the block, and \(t_\blockidx\) captures the learning task's importance on that given feature block. 
This parametrisation naturally emerges when considering the training of neural networks in specific regimes \citep{chizat_lazy_2019} and additionally the features will have distinct robustness properties \citep{tsipras_robustness_2019}.

The BFM allows for the expression of artificial datasets capturing the simplest definitions of structure, whilst capturing the intricacies of realistic power-law data, as can be found in 
real images \citep{wainwright_scale_1999,simoncelli_natural_2001}. 

Additionally, we define the \textit{Strong Weak Feature Model (SWFM)} as a special case of the BFM. In the SWFM, we only consider two blocks \(k=2\) where the relative sizes are tuned as \(d_\blockidx / d \to \phi_\blockidx \in (0,1)\). 
Under this data model we will also speak of the usefulness and robustness of a single block of features, where we generalise \cref{eq:mt-usefulness-def,eq:mt-robustness-def} just for a subset of the features.

\section{MAIN TECHNICAL RESULTS : EXACT ASYMPTOTICS}\label{sec:technical-result}

The core technical result is a rigorous, closed-form characterisation of the properties of the estimator for the previously described model, and the corresponding training and generalisation errors in the high-dimensional limit.

\textbf{Assumptions} 
The first assumption that we consider is that all matrices to have a well defined spectral distribution in the high dimensional limit. 
We will consider \(\Sigmax = \mathrm{S}^\top \operatorname{diag}(\omega_i) \mathrm{S}\), \(\zeta_i = \operatorname{diag}(\mathrm{S} \Sigmadelta \mathrm{S}^\top)_{i}\) and \(\upsilon_i = \operatorname{diag}(\mathrm{S} \Sigmaupsilon \mathrm{S}^\top )_i\).
Next we assume that \(\wstar^\top \Sigmax \wstar / d\) converges to a given value \(\rho\) in the limit and that the entries of \(\bar{\boldsymbol{\theta}} = \mathrm{S} \Sigmax^\top \wstar / \sqrt{\rho}\) converge as well to a limiting distribution.
Finally we assume that in the high dimensional limit the spectral distributions for the matrices and the distributions of the elements of the vectors just defined converge jointly to a p.d.f., \textit{i.e.} \(\nicefrac{1}{d} \sum_{i=1}^{d} \delta(\omega - \omega_i)\delta(\bar{\theta} - \bar{\theta}_i) \delta(\zeta - \zeta_i)\delta(\upsilon - \upsilon_i) \to \mu(\omega, \bar{\theta}, \zeta, \upsilon)\).

These assumptions are standard in high-dimensional statistics and naturally extend previous frameworks to our adversarial setting. The existence of well-defined spectral distributions, appearing in works like~\citet{doi:10.1073/pnas.1810420116,mei_generalization_2022}, is necessary to characterise the asymptotic behaviour of the data model and training procedure. The convergence of \(\wstar^\top \Sigmax \wstar / d\) and \(\bar{\boldsymbol{\theta}}\) ensures the signal strength remains controlled as dimensionality increases, following similar conditions in~\citet{Loureiro_2022}. Finally, the joint convergence assumption of the spectral distributions, as used in~\citet{dhifallah2020precise}, guarantees that correlations between data, attack, and defence geometries are well-behaved in the high-dimensional limit.

Under the previous assumptions and the model of~\cref{sec:setting-specification} we can characterise the behaviour of the clean and adversarial generalisation error of the ERM estimator
$\what(\mathcal{D})$ that minimises the risk in~\cref{eq:adversarial-training-problem}.

\begin{theorem} \label{thm:main-theorem-saddle-point-eqs}
For the ERM estimator of the risk function with \(\ell_2\) regularisation \(r(\w) = \frac{\lambda}{2} \norm{\w}_2^2\) and \(\lambda \geq 0\), under the data model defined in \cref{sec:setting-specification} and in the high dimensional proportional limit, the generalisation error \(\generr\) and the boundary error \(\bounderr\) defined in \cref{eq:gen-error-definition,eq:definition_adv_boundary_property} concentrate to
\begin{align}
    \generr &= \frac{1}{\pi} \arccos \qty( 
    {
        m / \sqrt{ (\rho + \tau^2) q}
    } 
    ) \label{eq:gen-error-theorem-overlap} \,, \\
    \bounderr &= \!
    \int_{0}^{\advgencost \frac{\sqrt{A}}{\sqrt{q}}} \!\!\!\!
    {\textstyle
        \erfc\left( \frac{ -\frac{m}{\sqrt{q}} \nu}{\sqrt{2 \qty(\rho + \tau^2 - m^2 / q)}} \right) 
        \!\! \frac{e^{-\frac{\nu^2}{2}}}{\sqrt{2\pi}} \dd{\nu}
    }
    \label{eq:adv-error-theorem-overlap} \,, 
\end{align}
and the adversarial generalisation error concentrates to \(\advgenerr = \generr + \bounderr\).

The values of \(m\) and \(q\) are the solutions of a system of eight self-consistent equations for the unknowns \((m, q, V, P, \hat{m}, \hat{q}, \hat{V}, \hat{P})\).
The first four equations are dependant on the loss function \(g\) and the adversarial training strength \(\advtrainingcost\) and read
\begin{equation}\label{eq:saddle-point-channel-main-thm}
    \begin{cases}
        \hat{m} = \alpha \mathbb{E}_{\xi}\left[
            \int_{\RR} \dd{y} \partial_\omega \mathcal{Z}_0 f_g(y, \sqrt{q} \xi, P)
        \right] \\
        \hat{q} = \alpha \mathbb{E}_{\xi}\left[
            \int_{\RR} \dd{y} \mathcal{Z}_0 f_g^2(y, \sqrt{q} \xi, P)
        \right] \\
        \hat{V} = -\alpha \mathbb{E}_{\xi}\left[
            \int_{\RR} \dd{y} \mathcal{Z}_0 \partial_\omega f_g(y, \sqrt{q} \xi, P)
        \right] \\
        \hat{P} = - \frac{\advtrainingcost}{2 \sqrt{P}} \alpha \mathbb{E}_{\xi} \qty[ 
            \int_{\RR} \dd{y} y \mathcal{Z}_0 f_g (y, \sqrt{q} \xi, P)
        ] 
    \end{cases} \,,
\end{equation}
where \(\xi \sim \gaussdist{0}{1}\) and 
\(\mathcal{Z}_{0} = \nicefrac{1}{2} \erfc(\nicefrac{- y \omega}{ \sqrt{2(V + \tau^2)}})\) and \(f_{g}(y, \omega, V, P) = \qty(\mathcal{P}(\omega) - \omega) / V\), where \(\mathcal{P}\) is the following proximal operator
\begin{equation}
    \mathcal{P} (\omega) = \min_x \qty[\frac{(x - \omega)^2}{2 V} + g(yx - \advtrainingcost \sqrt{P})] \,.
\end{equation}
The second set of equation depend on the spectral distribution of the matrices \(\Sigmax, \Sigmadelta\) and on the limiting distribution of the elements of \(\bar{\boldsymbol{\theta}}\). The equations read
\begin{equation}\label{eq:saddle-point-prior-main-thm}
    \begin{cases}
        m = \mathbb{E}_{\mu} \qty[ 
            \frac{\hat{m} \bar{\theta}^2}{\lambda + \hat{V} \omega + \hat{P} \delta} 
        ] \\
        q = \mathbb{E}_{\mu} \qty[ 
            \frac{
                \hat{m}^2 \bar{\theta}^2 \omega + \hat{q} \omega^2
            }{
                (\lambda + \hat{V} \omega + \hat{P} \delta)^2
            } 
        ] \\
        V = \mathbb{E}_{\mu} \qty[ 
            \frac{\omega}{\lambda + \hat{V} \omega + \hat{P} \delta } 
        ] \\
        P = \mathbb{E}_{\mu} \qty[ 
            \zeta \frac{
                \hat{m}^2 \bar{\theta}^2 + \hat{q} \omega^2
            }{
                (\lambda + \hat{V} \omega + \hat{P} \delta)^2
            } 
        ] 
    \end{cases}\,.
\end{equation}
The value of \(A\) can be obtained from the solution of the same system of self consistent equations as
\begin{equation}
    A = \! \mathbb{E}_{\mu} \! \qty[ 
    {\textstyle
        \upsilon \frac{\hat{m}^2 \bar{\theta}^2 \omega + \hat{q} \omega^2}{(\lambda + \hat{V} \omega + \hat{P} \delta)^2} 
    }
    ] \,.
\end{equation}
\end{theorem}

The proof is based on rephrasing the minimisation of the risk as a constrained optimisation problem for which we develop a GAMP algorithm. This algorithm, upon convergence, minimises the risk and has the advantage that his performance can be asymptotically described by a low dimensional description called state evolution.
The details are provided in \cref{sec:proofs-appendix,sec:error-metrics-app}. 

The separation of the equations into two parts, one depending on loss function and the other one on regularisation is common to many high-dimensional analyses of convex estimation problems.
The difference is that the adversarial setting introduces two additional parameter \(P, \hat{P}\) without changing this structure.

The parameters \(m,q,P\) and \(A\) are the values to which the following quantities concentrate in high-dimension
\begin{equation}\label{eq:interpretation-overlaps}
\begin{aligned}
    m &= {\textstyle \mathbb{E}_{\mathcal{D}}\qty[\frac{1}{d} \wstar^\top \Sigmax \what] }\,, & \  
    q &= {\textstyle \mathbb{E}_{\mathcal{D}}\qty[\frac{1}{d} \what^\top \Sigmax \what] }\,, \\
    P &= {\textstyle \mathbb{E}_{\mathcal{D}}\qty[\frac{1}{d} \what^\top \Sigmadelta \what] } \,, &\ 
    A &= {\textstyle \mathbb{E}_{\mathcal{D}}\qty[\frac{1}{d} \what^\top \Sigmaupsilon \what] } \,.
\end{aligned}
\end{equation}
These order parameters have the following interpretations \(m\) measures the alignment between the estimator and the true parameter in the geometry of the data, \(q\) quantifies the magnitude of the estimator in the same geometry, while \(P\) and \(A\) measure the defence and attack strengths respectively in their corresponding geometries.

The previous quantities, because of the average over the dataset of \(\what\), are not directly accessible if not after a \(d\) dimensional minimisation. We rephrased the problem into finding the solution of a low-dimensional system of coupled equations which can be solved efficiently.

\section{TRADE-OFFS IN THE LARGE SAMPLE COMPLEXITY REGIME}\label{sec:tradeoffs-large-sample-complexity}

\begin{figure*}[t!]
    \centering
    \includegraphics[width=0.875\textwidth]{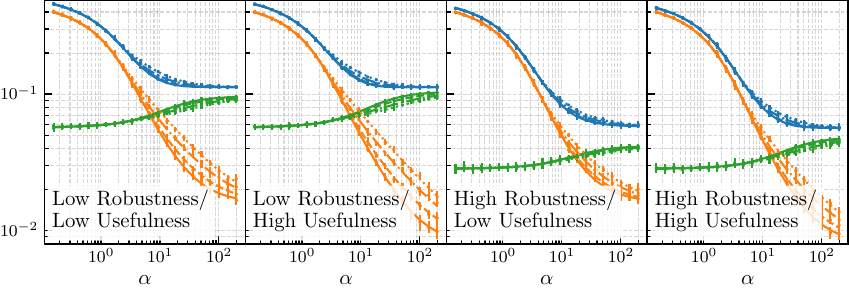}
    \raisebox{0.2\height}{
    \includegraphics[width=0.10\textwidth]{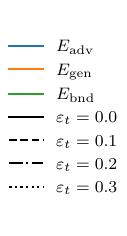}
    }
    \caption{%
        Error metrics as a function of the sample complexity \(\alpha\) for different combinations of high/low robustness and high/low usefulness.
        We see a good agreement between the theory (lines) and finite size simulations (error bars) already for \(d=1000\). Settings in \cref{sec:settings-figure-2}.
    }
    \label{fig:feature_comparison}
\end{figure*}

\paragraph{Introduction: Adversarial Trade-Off}
From the decomposition of the adversarial error we can simplify its form to get
\begin{equation}\label{eq:simplified-adv-error-formula}
    \advgenerr = 
    {
        \generr(\vartheta, \usefulmetric_{\wstar}) + 
        \int_{0}^{\advgencost \varkappa } 
        f(\xi; \vartheta, \usefulmetric_{\wstar}) \dd{\xi}
    } \,,
\end{equation}
where we introduce the variable \(\vartheta = m / \sqrt{\rho q}\) and \(\varkappa = \sqrt{A} / \sqrt{q}\). 
\(\vartheta\) is the cosine of the angle between the teacher weights \(\wstar\) and the student estimate \(\what\) in the geometry of \(\Sigmax\) and \(\varkappa\) is the norm of \(\what\) under the attack matrix.
The function \(f(\xi;\vartheta)\) is positive \(\forall \vartheta, \forall \xi \in [0,+\infty)\) and it is strictly increasing in \(\vartheta\) for any fixed \(\xi \in [0,+\infty)\). 

On the one hand, \(\generr\) is a monotonically decreasing function of \(\vartheta\). In other words, to improve generalisation error, it is best to align well with \(\wstar\). 
On the other hand, \(\bounderr\) is an increasing function of \(\vartheta\).
For a fixed attack strength the boundary error decreases choosing a \(\what\) that aligns less with \(\wstar\) and more in the directions where the attack is weak.
As a result, when we optimise the student vectors \(\what\) to a lower generalisation error, we increase the boundary error, and vice versa.
To minimise the adversarial error \(\advgenerr\), we must find a balance between these competing objectives.
This behaviour is common for boundary based classifiers \citep{tanay_boundary_2016}.

\subsection{Building Non-Robust, but Useful Features}\label{sec:all-features}

We notice that the values for \(\generr\) and \(\bounderr\) change by varying the usefulness and robustness of the features for fixed types of attacks. 
Intuitively, we have that the more usefulness one has the less generalisation error one makes, indeed we can write a lower bound for the generalisation error
\begin{equation}\label{eq:bound-gen-error}
    \generr
    \geq
    \frac{1}{\pi} \arccos( \sqrt{\frac{\pi}{2 \rho}} \usefulmetric_{\wstar} ) \,. 
\end{equation}
We note that \(\rho\) and \(\usefulmetric_{\wstar}\) only depend on \(\Sigmax\) and \(\wstar\).

Robustness only affects the boundary error. 
High robustness implies less sensibility to adversarial attacks: robust features have less samples within an attack range of the student decision boundary. 
The highest value that the boundary error can achieve is limited by both the robustness and the usefulness as
\begin{align} \label{eq:bound-boundary-error}
    \bounderr \leq & \  
    2 \mathrm{T}\qty( \advgencost \mathcal{A} \, \mathcal{B}, \mathcal{A}^{-1} ) - \frac{1}{\pi} \arctan( \mathcal{A}^{-1} ) \nonumber\\ 
    & - \frac{1}{\pi} \erf\qty( \frac{\advgencost \mathcal{B}}{\sqrt{2}} ) \erfc\qty( \frac{\advgencost \mathcal{A} \, \mathcal{B}}{\sqrt{2}} ) \,,
\end{align}
where \(\mathcal{B} = \max_i \sqrt{(\Sigmaupsilon)_{ii} / (\Sigmax)_{ii}}\), \(\mathcal{A} = \sqrt{\pi} \usefulmetric_{\wstar} / \sqrt{2 \rho}\) and \(\mathrm{T}\) is the Owen function. This previous bound is a decreasing function of the robustness.

We are particularly interested in studying the effects of the interplay between \(\generr\) and \(\bounderr\) in the large sample complexity regime.
We show that the adversarial error always goes to a constant in the high \(\alpha\) regime, \textit{i.e.} \(\lim_{\alpha \to \infty} \partial_\alpha \advgenerr = 0\)
where the limit \(\alpha \to \infty\) is taken after the high-dimensional proportional limit. Additionally, we show that the constant to which \(\advgenerr\) converges, can be zero only if \(\tau = \advgencost = 0\). The details are provided in \cref{sec:large-alpha-limit-of-adverr}.

\begin{figure*}[t]
    \centering
    \includegraphics[width=0.58\textwidth]{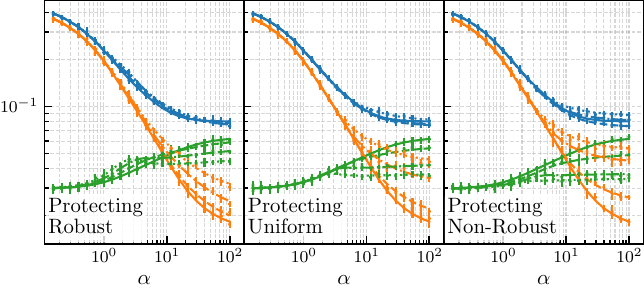}
    \raisebox{0.18\height}{
    \includegraphics[width=0.081\textwidth]{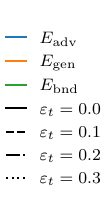}
    }
    \includegraphics[width=0.28\textwidth]{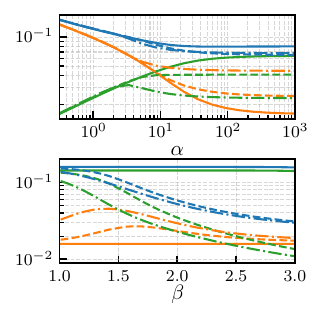}
    \caption{%
        \textbf{(Left)} 
        Error metric as a function of sample complexity for different choices of \(\Sigmadelta\). 
        The model considered is a SWFM: both blocks have the same usefulness and they differ in robustness.
        Each defence strategy leads to a different asymptotic values in the large sample complexity regime for the same values of \(\advtrainingcost\).
        \textbf{(Right, Top)} 
        Error metric as a function of \(\alpha\) for a power-law BFM with \(\Sigmaupsilon = \Sigmadelta = \mathbf{1}\).
        \textbf{(Right, Bottom)} 
        Error metrics as a function of the power law exponent of the data \(\beta\), i.e. \((\Sigmax)_{ii} = i^{-\beta}\) and for fixed \(\Sigmaupsilon = \Sigmadelta = \mathbf{1}\). We fixed the same usefulness for all values of \(\beta\).
        A higher \(\beta\) increases the percentage of non-robust features increasing the number of features protected by $\Sigmadelta$. 
        Settings in \cref{sec:settings-figure-3}.  
    }
    \label{fig:defence_sweep}
    \label{fig:power-law-behaviour}
\end{figure*}

One could expect that by performing a correct cross validation of the hyper parameters \(\advtrainingcost\) and \(\Sigmadelta\), one could achieve an improvement in the value of \(\advgenerr\) in the regime of data abundance. We show that this is not the case for too simple data models.
\begin{proposition}
Under the same setting as \cref{thm:main-theorem-saddle-point-eqs} and considering a BFM with a single type of feature, i.e. \(k=1\) one has that \(\forall \advgencost, \advtrainingcost \geq 0\) for \(\alpha\) big enough 
exist two positive numbers $M_1, M_2$ such that
\begin{equation}
\begin{aligned}
    \abs{
        \advgenerr(\advgencost, \advtrainingcost) - \advgenerr(\advgencost,  \advtrainingcost = 0)
    } & < M_1 / \alpha \,, \\
    \abs{
        \generr(\advtrainingcost) - \generr(\advtrainingcost = 0)
    } & < M_2 / \alpha \,,
\end{aligned}
\end{equation}
where \(\advgenerr(\advgencost, \advtrainingcost)\) and \(\generr(\advtrainingcost)\) define the adversarial and generalisation error of \(\what\) trained with \(\advtrainingcost\) and evaluated for \(\advgencost\).
\end{proposition}

The proof is based on the asymptotic expansion of the result of \cref{thm:main-theorem-saddle-point-eqs} and it is presented in \cref{sec:single-block-model-high-alpha}. 
Thus we proved that in the large sample complexity regime, there is no benefit in adversarial training and indeed the values of \(\advgenerr,\generr\) are universal for any \(\advtrainingcost\) chosen in training.
Note that the previous proposition also covers the setting of \cite{javanmard_precise_tradeoffs_2020}.

\Cref{fig:feature_comparison} shows the dependency of \(\generr\), \(\bounderr\) and \(\advgenerr\) as a function of the sample complexity \(\alpha\) for the different combination of usefulness and robustness. 
We see that the plateau's value of the generalisation error mainly depends on usefulness (higher usefulness/lower plateau) while the one of boundary error mainly depends on robustness (higher robustness/lower plateau). Additionally, we also see that the effect of adversarial training (distance between curves) gets smaller as \(\alpha\) increases.

\subsection{Directional Defences \& Structured Data}\label{sec:features-to-defend}

After the analysis of previous section, we proceed to study the effects that defending different features through the choice of \(\Sigmadelta\) has on adversarial performances.
We consider the following three defence strategies: \textit{defending the robust features}, \textit{uniformly defend all the features} and \textit{defend the non-robust features}.

\Cref{fig:defence_sweep} (Left) presents the comparison between the three different defence matrices for a SWFM with one robust block of features and one not.
The different normalised defence matrices \(\Sigmadelta\) change the relative protection between the two blocks of features, while \(\Sigmaupsilon = \mathbb{1}\) for all cases.
In the high \(\alpha\) region of the curves we see that the more we protect the non robust features the more \(\generr\) increases for the same value of \(\advtrainingcost\). 
Simultaneously we have that the more we protect non-robust features the more \(\bounderr\) decreases.
The value of \(\advgenerr\) does not always decrease or increase by protecting more the non-robust features, hinting at the trade off between \(\generr\) and \(\bounderr\).

A similar behaviour can be found for datasets like CIFAR10 \citep{krizhevsky_learning_nodate} and FashionMNIST \citep{xiao2017fashion} as we explore in \cref{sec:real_data_experiments}. We find a choice of \(\Sigmadelta\) with entries proportional to the inverse of the eigenvalue and uniform entries effective.

\begin{figure*}[t]
    \centering
    \raisebox{0.0045\height}{
        \includegraphics[width=0.278\textwidth]{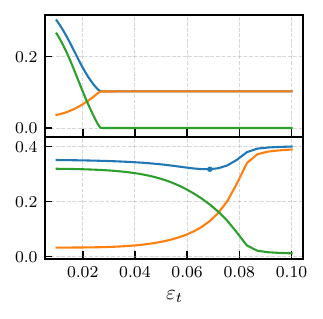}
    }
    \raisebox{0.09\height}{
        \includegraphics[width=0.09\textwidth]{mt_imgs/DefenceSweep/vertical_legend.pdf}
    }
    \raisebox{0.0495\height}{ 
        \includegraphics[width=0.252\textwidth]{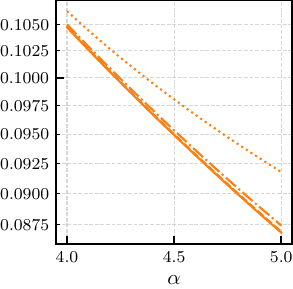}
    }
    \raisebox{0.0495\height}{
        \includegraphics[width=0.252\textwidth]{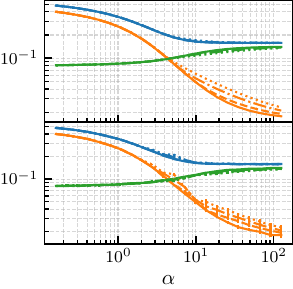}
    }
    \caption{%
        \textbf{(Left)}
        We show the adversarial errors as a function of the adversarial defence strength in the case \(\alpha \gg 1\).
        \textit{(Top)}
        Defending features that are on average orthogonal to the teacher vector. 
        \textit{(Bottom)}
        Defending the features that are on average parallel to the teacher vector. 
        \textbf{(Center)}
        Adversarial training is not just an \(\ell_2\)-regularisation, we show the value of the generalisation error \(\generr\) as a function of the sample complexity where the parameter \(\lambda\) is chosen such that it minimises the generalisation error.      
        \textbf{(Right)}
        Learning curves for \textit{(Top)} adversarial training in \cref{eq:modified-minimisation-problem} and \textit{(Bottom)} the equivalent problem in \cref{eq:equivalent-adversarial-regularisation}.  
        Settings in \cref{sec:settings-figure-4}.
    }
    \label{fig:regularisation-equivalence}
    \label{fig:optimal_defense}
\end{figure*}

We are able to provide an analytical description of this phenomena under the data model considered.
\begin{proposition}\label{prop:defending-less-robust}
Consider the SWFM defined in \cref{sec:strong-weak-feature-model}
where the defence matrix is 
\(
    \Sigmadelta = \operatorname{blockdiag}\qty( 
        (\Delta_1 + \delta_1 \varrho ) \mathbb{1}_{d_1}, (\Delta_2 + \delta_2 \varrho ) \mathbb{1}_{d_2} 
    )
\), 
with \(\varrho\) the parameter that makes the defence matrix change.
Assume also that \(\psi_1 > \psi_2\), \(\Delta_2 \psi_1 \geq \Delta_1 \psi_2\) and \(\Upsilon_i = 1\).\footnote{This assumption corresponds to saying that the first set of component is more robust than the second and that the defence's effect is greater in the more robust subspace or at least equal between the two. Additionally the attacker does not distinguish the two subspaces.}
In the \(\alpha\to \infty\) (taken after the \(n,d\to\infty\)) there exists \(\kappa > 0\) such that \(\forall \delta_1 > \kappa\), \(\delta_2 = -\delta_1\) one has that
\begin{equation}
\begin{aligned}
    \bounderr(\varrho) &= \bounderr^0 + \bounderr^1 \varrho + \order{\varrho^2} \,, \\
    \generr(\varrho) &= \generr^0 + \generr^1 \varrho + \order{\varrho^2} \,,
\end{aligned}
\end{equation}
where \(\generr^1 > 0\), \(\bounderr^1 < 0\) and \(\bounderr^0,\generr^0\) are the errors when \(\varrho =0\).
Additionally, this leads to an improved value of \(\advgenerr\) at order \(\varrho\) iff the following condition is satisfied
\begin{equation}
    {\textstyle
        \frac{\advgencost}{\sqrt{2}}
        \erfc\qty( -\frac{\vartheta_0 u_0 \advgencost}{ \sqrt{2-2 \vartheta_0 ^2} }) 
        < \frac{e^{-\frac{\vartheta_0^2 u_0^2 \advgencost ^2}{2 \left(1-\vartheta_0 ^2\right) } }}{\sqrt{\pi} \sqrt{1-\vartheta_0^2}} \,,
    }
\end{equation}
where \(\vartheta_0 = m_0 / \sqrt{\rho q_0}\) and \(u_0 = \sqrt{A_0} / \sqrt{q_0}\) the solution of the problem with \(\varrho = 0\). 
Notice that for \(\advgencost\) small enough this condition is always verified.
\end{proposition}
We prove this proposition in \cref{sec:change-defence-direction} by expanding in \(\varrho\) the large \(\alpha\) equations of \cref{thm:main-theorem-saddle-point-eqs}.

A similar phenomenology of protection of weak features can be seen for a power-law BFM in \cref{fig:power-law-behaviour} (Right, Top).
Adversarial training (\(\advtrainingcost > 0\)) decreases \(\bounderr\) and increases \(\generr\) in the high sample complexity regime, while the adversarial error does not always decrease.

In \cref{fig:power-law-behaviour} (Right, Bottom) we present the behaviour of (\(\advgenerr, \generr, \bounderr\)) as a function of the power law exponent \(\beta\) of \(\Sigmax\). 
Heavier tails (lower \(\beta\)) increase the total number of robust features and thus more features are not sensible to the attack. Conversely, weaker tails (bigger \(\beta\)) reduce the number of robust features. 

To summarise, 
we see that the uniform defence strategy ($\Sigmadelta = \mathbb{1}$) performs well as it successfully defends the non-robust features, which are naturally prioritised when acting on all of them equally. The robust features are less affected by the perturbation as they have a bigger margin.

\section{ADDITIONAL EXPLORATIONS}\label{sec:additional-explorations}

\paragraph{Tradeoff directions and innocuous directions}

We now investigate the effect that different types of geometries have on the trade-off between \(\generr\) and \(\bounderr\).  
Depending on the attack geometry \(\Sigmaupsilon\) one can choose different defence geometries \(\Sigmadelta\) and ask if and for which, protection without trade-off is possible. 
Any attack matrix \(\Sigmaupsilon\) eigenvalues can be split into 
directions orthogonal to the teacher and directions aligned with the teacher. 
\Cref{fig:optimal_defense} (Left) considers the effect of the adversarial training strength \(\advtrainingcost\) on the errors for different choices of matrices \(\Sigmadelta = \Sigmaupsilon\). 
In the the top we consider matrices whose biggest eigenvalues are orthogonal to the teacher vector and in the bottom one matrices where there is a leading eigenvector in the direction of the teacher.

We start by noticing that there is a qualitative difference between the two cases. 
If the attack focuses on features not important for the learning task (orthogonal to the teacher vector) the effect of the attack can be neutralised by choosing an \(\advtrainingcost\) big enough. In this case, even if the features are just orthogonal on average, we see that \(\advgenerr\) is decreasing as \(\advtrainingcost\) decreases and for \(\advtrainingcost\) big enough we have that \(\advgenerr = \generr\) with \(\bounderr = 0\).
The features that are important for the learning task (aligned with \(\wstar\)) are the ones causing the trade-off. In this case there exists a single value of \(\advtrainingcost\) such that is minimal and thus cross validation over this hyperparmeter is necessary to obtain optimal performances.

\paragraph{Interpreting Adversarial Training as a Data-Dependent Regularisation}
Recently, \citet{ribeiro2023regularization} show that for certain problems adversarial training can be exactly expressed as a data dependent regularisation.
We investigate this claim within the framework of our model. 
In \cref{fig:regularisation-equivalence} (Center) we show the generalisation error as a function of sample complexity for a BFM where each matrix is chosen as the identity and \(\lambda\) as been optimised to obtain minimal \(\generr\). 
We see that the performance still depend on the value \(\advtrainingcost\) used during training. 
In this sense there is a difference between two \(\what\) obtained for different values of \(\advtrainingcost\) that cannot be eliminated by a careful choice of the regularisation strength.

In \cref{sec:equivalent-problem} we show how, from a careful series expansion around the minimiser, one can approximate the minimisation problem in \cref{eq:modified-minimisation-problem} with another minimisation problem without adversarial attacks but with explicit regularisation.
The form for the approximate loss in the case of small \(\advtrainingcost\) is
\begin{equation}\label{eq:equivalent-adversarial-regularisation}
    \sum_{\datidx = 1}^{n} 
    g \Big( y_\datidx \frac{\w^\top \x_\datidx}{\sqrt{d}} \Big) 
    + \tilde{\lambda}_1 \sqrt{\w^\top \Sigmadelta \w} + \tilde{\lambda}_2 \w^\top \Sigmadelta \w
\end{equation}
where \(\tilde{\lambda}_1\) and \(\tilde{\lambda}_2\) depend on the model's parameters and perturbed margins of the points that shift sign under perturbation. 

In \cref{fig:regularisation-equivalence} (Right) we compare the generalisation performances obtained by minimising the true adversarial problem at the top with the performances of minimising the approximate problem. We see a qualitative match between the two even if the problem in \cref{eq:equivalent-adversarial-regularisation} is less numerically stable.

The effective regularisation  
is equivalent to a directional \(\sqrt{\ell_2} + \ell_2\) regularisation. 
The \(\sqrt{\ell_2}\) term indicates a regularisation linearly proportional to the norm of the student vector but that does not favour sparsity as \(\ell_1\) regularisation. 


\subsection*{Acknowledgements}
We thank Lenka Zdeborov\'a for fruitful discussions and insightful ideas regarding a class-preserving error, Lucas Clarte for useful discussions about the relevant literature, Guillaume Dalle for help in with the numerical implementation, Nikolaos Tsilivis for the discussion during the Cargese 2023 Workshop {\it Statistical Physics and Machine Learning back together again}, Pierre Mergny for the always helpful clarifications about Random Matrix Theory, Paul Krzakala for pointing relevant literature on adversarial training, Julia Kempe for the fruitful discussions and Vittorio Erba for rereading of the manuscript. BL acknowledges support from the \textit{Choose France - CNRS AI Rising Talents} program, and FK from the Swiss National Science Foundation grant SNFS OperaGOST  (grant number $200390$).

\bibliographystyle{unsrtnat}
\bibliography{arxiv}

\onecolumn
\newpage
\appendix
\part*{Supplementary Materials}%

In \cref{sec:proofs-appendix} we present a rigorous proof of the main theoretical result introduced in the main body of the paper. This section aims to provide a derivation of the main result based on the body of literature on the use of AMP and CGMT. 
In \cref{sec:error-metrics-app} we derive the theoretical formulas for the the error metrics in terms of the overlap solutions of the fixed-point equations presented in \cref{sec:technical-result}. Again based on the previous literature we explain the derivation of these formulas and clarify explicitly how the new error metric of the class-preserving error is derived.
In \cref{sec:large_alpha} we expand the fixed-point equations from \cref{sec:technical-result} in the high sample complexity regime, offering insights into how the solutions behave as the complexity increases. The expansion provides important implications for practical applications as it studies the limiting performances that can be reached even with infinite number of data.
Additionally we provide further results related to the adversarial generalisation error, including its relation to Owen's T function.
In \cref{sec:equivalent-problem} we show how to rewrite the adversarial problem in terms of a data-dependent regularisation, providing an alternative formulation that aids in better understanding the adversarial setup and its effect on the bias it gives to the solution.
In \cref{sec:fast-gradient-method} we consider another type of attack and provide a similar high-dimensional description of it.
In \cref{sec:best-choice-attack-defence} we explore the relationship between the attack geometry and the defense geometry, providing geometric interpretations and insights into how different adversarial attack strategies can influence and be mitigated by corresponding defense mechanisms.
In \cref{sec:real_data_experiments}, we demonstrate the robustness metrics and various defense strategies on the Cifar-10 \cite{krizhevsky_learning_nodate} and FashionMNIST \cite{xiao2017fashion} datasets. Detailed experimental results highlight the performance and effectiveness of the proposed methods.
\cref{sec:figures-setting-general} provides a comprehensive explanation of the figure settings used in the main text is provided here, ensuring reproducibility.
In \cref{sec:replica-computation} we derive in detail the same result presented in \cref{sec:proofs-appendix} with the use of the statistical physics's replica method. This offers a different perspective on the problem, serving as a complementary approach to the formal proof.

\section{HIGH-DIMENSIONAL ASYMPTOTICS}\label{sec:proofs-appendix}

In this section, we present a comprehensive proof of the fixed-point equations outlined in~\cref{thm:main-theorem-saddle-point-eqs}. Our approach leverages recent advancements in high-dimensional statistics and convex optimization to establish the asymptotic behavior of adversarial training in the high-dimensional limit.

The proof unfolds in three main stages
\begin{description}
    \item[Problem Reformulation] We begin by recasting the original optimization problem into an equivalent form with additional constraint variables $P$ and $\hat{P}$. This reformulation allows us to apply techniques such as the mapping to a GAMP algorithm.
    \item[Algorithmic Interpretation] We demonstrate that the reformulated problem can be solved by a Generalized Approximate Message Passing (GAMP) algorithm. This connection not only provides an algorithmic perspective but also establishes the optimality of the GAMP solution at convergence.
    \item[Asymptotic Analysis] Building upon the well-established literature about high-dimensional asymptotics \citep{Donoho_AMP_compress_sensing_2010,Krzakala_2012,miolane2021distribution,Loureiro_2022}, we derive the low-dimensional asymptotics of our system. This step is crucial in bridging the gap between finite-dimensional system describing our system and the $d$-dimensional inital problem.
\end{description}
For readers well-versed in statistical physics, we provide an alternative derivation using the replica method in~\cref{sec:replica-computation}.

\subsection{Notations and Definitions}

In this paper, we extensively employ the concepts of Moreau envelopes and proximal operators, pivotal elements in convex analysis frequently encountered in recent works on high-dimensional asymptotic of convex problems \cite{boyd_vandenberghe_2004,parikh_boyd_proximal_algo}. For an in-depth analysis of their properties, we refer the reader to the cited literature. Here, we briefly outline their definition and the main properties for context.

The Moreau envelope and the proximal operator associated to a scalar function \(f: \RR \to \RR\) are defined as
\begin{equation}\label{eq:moreau-proximal-definitions}
    \mathcal{M}_{V f(\cdot)} (\omega) = \min_{x} \qty[\frac{(x - \omega)^2}{2 V} + f(x)] \,,\quad \mathcal{P}_{V f(\cdot)} (\omega) = \argmin_{x} \qty[\frac{(x - \omega)^2}{2 V} + f(x)]\,,
\end{equation}
where the \((\cdot)\) indicates the variable considered if the function is of more variables. Generally one can consider the Moreau envelope or the proximal with respect to just one of the inputs of a function depending on more variables.

In the rest of the paper we will use the following properties of the Moreau and Proximal that can be found in \cite{parikh_boyd_proximal_algo}. We will be using the \textit{envelope theorem} which states
\begin{equation}
    \partial_{\omega} \mathcal{M}_{V g(y,\cdot)}(\omega)= V^{-1}\left(\omega-\mathcal{P}_{V g(y,\cdot)}(\omega)\right) \,.
\end{equation}

Additionally we will use the two following results
\begin{equation}
    \mathcal{M}_{V f(\cdot + u)} (\omega) 
    = \mathcal{M}_{V f(\cdot)} (\omega + u) \,, \quad 
    \mathcal{P}_{V f(\cdot + u)} (\omega) 
    = u + \mathcal{P}_{V f(\cdot)} (\omega + u) \,.
\end{equation}

\subsection{Approximate Message Passing}

In the landscape of high-dimensional statistical inference, Approximate Message Passing (AMP) algorithms have emerged as a cornerstone for efficiently solving problems like compressed sensing \cite{Donoho_AMP_compress_sensing_2010,Krzakala_2012}. 
At their hearts AMP algorithms are iterative schemes, inspired by the ISTA algorithm \cite{ista_2009}, that in addition leverage the statistical properties of high-dimensional random matrices to remove correlation at each step. 
A key feature of AMP algorithms is their connection with state evolution (SE), a powerful analytical tool that tracks the evolution of the AMP algorithm's performance across iterations. 
State evolution provides a set of deterministic equations that accurately predict the algorithm's behaviour in the limit of large system sizes, thus offering deep insights into the convergence properties and asymptotic accuracy of AMP algorithms. This method has been extensively used to understand problems like the learning of Gaussian mixtures \cite{Loureiro_gaussian_mixtures_2021} or learning curves of ensembling methods \cite{pmlr-v162-loureiro22a}.
This kind of connection and algorithm has been also know by statistical physicist \cite{zdeborova_inference_2016}.

In the case of estimation of an i.i.d. random vector observed through a linear transform followed by a component-wise, probabilistic (possibly nonlinear) measurement channel, an optimal algorithm, called generalised approximate message passing (GAMP) and its respective SE has been introduced by \cite{rangan_GAMP_2011}. 
The main result that associates the evolution of GAMP to the SE has been proven later by \cite{javanmard_montanari_gamp_2013}. 
Thus the idea is that the fixed point equations could be seen as the state evolution equations for a specific Generalised AMP algorithm that minimises the equivalent minimisation problem in \cref{eq:app:proof-equivalent-problem}. This way of proving the result also gives the advantage of defining an algorithm that, upon convergence, returns the vector \(\what\).

We will now define in general a GAMP sequence and state here the result about the general state evolution.
Consider a sequence Gaussian matrices $A(n)_{ij} \sim \mathcal{N}(0,1)$, with $i \in \qty{1, \dots n}, j \in \qty{1, \dots d}$, with i.i.d. Gaussian entries. For each $n,d \in \mathbb{N}$, consider two sequences of pseudo-Lipschitz functions
\begin{equation}
    \left\{\boldsymbol{h}_t: \mathbb{R}^{n  } \rightarrow \mathbb{R}^{n  }\right\}_{t \in \mathbb{N}} \quad\left\{\boldsymbol{e}_t: \mathbb{R}^{d  } \rightarrow \mathbb{R}^{d  }\right\}_{t \in \mathbb{N}}
\end{equation}
and recursively define $\boldsymbol{u}^i \in \RR^d$ and $\boldsymbol{v}^i \in \RR^n$ as
\begin{equation}\label{eq:app:general-gamp}
\begin{aligned}
    & \boldsymbol{u}^{t+1}=A^{\top} \boldsymbol{h}_t\left(\boldsymbol{v}^t\right)-\boldsymbol{e}_t\left(\boldsymbol{u}^t\right)\left\langle \boldsymbol{h}_t^{\prime}\right\rangle^{\top} \\
    & \boldsymbol{v}^t=A \boldsymbol{e}_t\left(\boldsymbol{u}^t\right)-\boldsymbol{h}_{t-1}\left(\boldsymbol{v}^{t-1}\right)\left\langle \boldsymbol{e}_t^{\prime}\right\rangle^{\top}
\end{aligned}
\end{equation}
where we define the Onsager terms as
\begin{equation}
    \left\langle\boldsymbol{h}_t^{\prime}\right\rangle=\frac{1}{d} \sum_{i=1}^n \frac{\partial \boldsymbol{h}_t^i}{\partial \boldsymbol{v}_i}\left(\boldsymbol{v}^t\right) \in \mathbb{R}
    \quad
    \left\langle\boldsymbol{e}_t^{\prime}\right\rangle=\frac{1}{d} \sum_{i=1}^d \frac{\partial \boldsymbol{e}_t^i}{\partial \boldsymbol{u}_i}\left(\boldsymbol{u}^t\right) \in \mathbb{R}
\end{equation}
The previous recursive relation is defined for a suitable initial condition that has a well-defined high-dimensional limit.

For any $d$ (and respective $n$) we have a series of vectors defined iteratively through~\cref{eq:app:general-gamp}.

We then define the state evolution from two sets of 
\begin{equation}
\begin{aligned}
& Q_{t+1, s}=Q_{s, t+1}=\lim _{d \rightarrow \infty} \frac{1}{d} \mathbb{E}\left[\boldsymbol{e}_s\left(\hat{Z}^s\right)^{\top} \boldsymbol{e}_{t+1}\left(\hat{Z}^{t+1}\right)\right] \in \mathbb{R} \\
& \hat{Q}_{t+1, s+1}=\hat{Q}_{s+1, t+1}=\lim _{d \rightarrow \infty} \frac{1}{d} \mathbb{E}\left[\boldsymbol{h}_s\left(Z^s\right)^{\top} \boldsymbol{h}_t\left(Z^t\right)\right] \in \mathbb{R}
\end{aligned}
\end{equation}
where $\left(Z^0, \ldots, Z^{t-1}\right) \sim \mathcal{N}\left(0,\left\{Q_{r, s}\right\}_{0 \leqslant r, s \leqslant t-1} \otimes I_n\right),\left(\hat{Z}^1, \ldots, \hat{Z}^r\right) \sim \mathcal{N}\left(0,\left\{\hat{Q}_{r, s}\right\}_{1 \leqslant r, s \leqslant t} \otimes I_d\right)$. Where the initial values are coherent with the ones of the iterates in~\cref{eq:app:general-gamp}.

\begin{theorem}[\citet{javanmard_montanari_gamp_2013}]
In the setting of the previous paragraph, for any sequence of pseudo-Lipschitz functions $\phi_n : (\RR^n \times \RR^d)^t \to \RR$ we have that
\begin{equation}
    \phi_n\left(\boldsymbol{u}^0, \boldsymbol{v}^0, \boldsymbol{u}^1, \boldsymbol{v}^1, \ldots, \boldsymbol{v}^{t-1}, \boldsymbol{u}^t\right) \stackrel{\mathrm{P}}{\sim} \mathbb{E}\left[\phi_n\left(\boldsymbol{u}^0, Z^0, \hat{Z}^1, Z^1, \ldots, Z^{t-1}, \hat{Z}^t\right)\right]
\end{equation}
where $\left(Z^0, \ldots, Z^{t-1}\right) \sim \mathcal{N}\left(0,\left\{Q_{r, s}\right\}_{0<r, s \leqslant t-1} \otimes I_n\right),\left(\hat{Z}^1, \ldots, \hat{Z}^t\right) \sim \mathcal{N}\left(0,\left\{\hat{Q}_{r, s}\right\}_{1 \leq r, s \leqslant t} \otimes I_n\right)$.
\end{theorem}

The previous theorem states the fact that for the limit of any pseudo lipshitz function of the iterates, and thus of the limiting value, one has that its behaviour concentrates to the expectation of gaussian variables with specific covariances.

\subsection{Assumptions}

For our results to hold we will need all the assumptions required in \cite{Loureiro_2022} for the application of their results on the saddle point equations. Our assumptions are thus all of the ones contained in \cite[Appendix~B.1]{Loureiro_2022}. Additionally we need
\begin{description}
   \item[(A1)] The attack and defence geometry \(\Sigmadelta\) and \(\Sigmaupsilon\) should be \(\Sigmaupsilon, \Sigmadelta \succ 0\).  The spectral distributions of the matrices \(\Sigmadelta,\Sigmaupsilon\) converge to distributions such that the overlaps 
   \begin{equation}
        P = \frac{1}{d} \mathbb{E}_{\mathcal{D}}\qty[\what^\top \Sigmadelta \what] \,, \quad 
        A = \frac{1}{d} \mathbb{E}_{\mathcal{D}}\qty[\what^\top \Sigmaupsilon \what] \,, \quad 
        F = \frac{1}{d} \mathbb{E}_{\mathcal{D}}\qty[\wstar^\top \Sigmaupsilon \what] \,,
   \end{equation}
   are well-defined. Additionally, the maximum singular values of them are bounded with high probability when \(n, p \to \infty\).
   \item[(A2)] The values \(\zeta_i = \operatorname{diag}(\mathrm{S} \Sigmadelta \mathrm{S}^\top)_{i}\), \(\upsilon_i = \operatorname{diag}(\mathrm{S} \Sigmaupsilon \mathrm{S}^\top )_i\) and \(\mathbf{f}_i = (\mathrm{S} \Sigmaupsilon^\top \wstar / \sqrt{\rho})_i \), where additionally \(\rho = \wstar^\top \Sigmax \wstar / d\), \(a = \wstar^\top \Sigmaupsilon \wstar / d\)
   and \(\bar{\boldsymbol{\theta}} = \mathrm{S} \Sigmax^\top \wstar / \sqrt{\rho}\) should have jointly a well defined limit when \(n, p \to \infty\).
   More formally
   \begin{equation}
       \frac{1}{d} \sum_{i=1}^d \delta \qty(\omega - \omega_i) \delta\qty(\bar{\theta} - \bar{\theta}_i) \delta\qty(\zeta - \zeta_i) \delta(\upsilon - \upsilon_i) \delta(f - f_i) \to \mu
   \end{equation}
   should converge for \(n, p \to \infty\).
   \item[(A3)] The choice of the matrices \(\Sigmadelta\) and \(\Sigmaupsilon\) should be independent of the teacher vector \(\wstar\).
\end{description}

\subsection{Reformulation of the problem}

Our analysis begins with the adversarial training problem introduced in~\cref{eq:adversarial-training-problem} of the main text. For clarity, we restate it here
\begin{equation}\label{eq:adversarial-training-problem-appendix}
    \sum_{\datidx = 1}^{n} 
    \max_{
        \norm{\vecdelta_\datidx}_{\Sigmadelta^{-1}} \leq \advtrainingcost 
    }
    g \qty(y_\datidx \frac{\w^\top \qty(\x_\datidx + \vecdelta_\datidx)}{\sqrt{d}}) 
    + r(\w) \,,
\end{equation}
where we remind that the regularisation function \(r : \RR^{d} \rightarrow \RR \) is convex and the loss function \(g : \RR \rightarrow \RR\) is non-increasing, meaning that \(x_1 \leq x_2\) implies \(g(x_1) \geq g(x_2)\).

The non-increasing property of \(g\) allows us to simplify the inner maximization, leading to
\begin{equation}\label{eq:app:proof-equivalent-problem}
    \sum_{\datidx = 1}^{n} 
    g \qty(
        y_\datidx \frac{\w^\top \x_\datidx}{\sqrt{d}} 
        - \advtrainingcost \frac{\sqrt{\w^\top \Sigmadelta \w}}{\sqrt{d}} 
    ) 
    + r(\w) \,.
\end{equation}

We can introduce two constrains to rewrite the minimisation as
\begin{equation}
    \sum_{\datidx = 1}^{n} 
    g \qty(
        y_\datidx,  \frac{\w^\top \x_\datidx}{\sqrt{d}} 
        - y_\datidx \advtrainingcost \sqrt{P} 
    )
    + r(\w) \quad \text{ such that } \quad d P = \w^\top \Sigmadelta \w \,, 
\end{equation}

The Lagrangian form of this problem reads
\begin{equation}\label{eq:lagrangian-problem-appendix}
    \mathcal{L}(\w, \boldsymbol{z}, \boldsymbol{s}, P, \hat{P}) =
    g\qty(\boldsymbol{z} - \advtrainingcost \sqrt{P}  \boldsymbol{y}) + 
    r(\w) +
    \boldsymbol{s} \qty(\frac{1}{\sqrt{d}} \boldsymbol{X} \w - \boldsymbol{z}) + \hat{P} \qty(\w^\top \Sigmadelta \w - d P)
\end{equation}
where we have simplified the notations by introducing
\begin{equation}
    g\qty(\boldsymbol{z} - \advtrainingcost \sqrt{P} \boldsymbol{y}) = 
    \sum_{\datidx = 1}^{n} 
    g \qty(
        y_\datidx, 
        z_\datidx - \advtrainingcost \sqrt{P}
    ) \,,
\end{equation}
and defined the feature matrix \(\boldsymbol{X} \in \RR^{n \times d}\).

\subsection{Fixed Point Equations for the constrain variables}

We can decompose this problem into separate optimizations over $\w$, $\boldsymbol{z}$, and $P$
To obtain the equations for \(P,\hat{P}\) we go back to the complete optimisation problem 
\begin{equation}
\begin{aligned}
    \sup_{\boldsymbol{s}, \hat{P}} \inf_{\w, \boldsymbol{z}, P} &\ \mathcal{L}(\w, \boldsymbol{z}, \boldsymbol{s}, P, \hat{P}) = \sup_{\boldsymbol{s}, \hat{P}} \Bigg[ 
        \inf_{\w} \qty[
            r(\w) + \hat{P} \w^\top \Sigmadelta \w + \frac{1}{\sqrt{d}} \boldsymbol{s}^\top \boldsymbol{X} \w
        ] + \inf_{\boldsymbol{z}, P} \qty[ 
        g\qty(\boldsymbol{z} - \advtrainingcost \sqrt{P} \boldsymbol{y}) - \boldsymbol{s}^\top \boldsymbol{z} - d \hat{P} P
    ] \Bigg]  \\
    &= \sup_{\boldsymbol{s}, \hat{P}} \Bigg[ 
        \inf_{\w} \qty[f(\w) + \hat{P} \w^\top \Sigmadelta \w + \frac{1}{\sqrt{d}} \boldsymbol{s}^\top \boldsymbol{X} \w] + 
        \inf_{\boldsymbol{z}} \qty[g(\boldsymbol{z}) - \boldsymbol{s}^\top \boldsymbol{z}] + \inf_{P} \qty[\advtrainingcost \sqrt{P} \boldsymbol{s}^\top \boldsymbol{y} - d \hat{P} P ]
    \Bigg]
\end{aligned}
\end{equation}
We also consider the gradients with respect to the new variables
\begin{align}\label{eq:app:gradients-proof}
    \pdv{P} &= \frac{\advtrainingcost}{2\sqrt{P}} \boldsymbol{s}^\top \boldsymbol{y} - d \hat{P} \,, 
    & \quad \pdv{\hat{P}} &= \w^\top \Sigmadelta \w - d P \,, 
\end{align}

\subsection{Generalised AMP mapping of our problem}

\begin{algorithm}
\SetAlgoLined
\DontPrintSemicolon
\caption{
    Adversarial Generalised Approximate Message Passing (advGAMP)
}\label{algo:adversarial-gamp}
\KwIn{Matrix \(\boldsymbol{X} \in \RR^{n \times d}\), functions \(r(\w) : \RR^d \to \RR, g(\boldsymbol{z}) : \RR^n \to \RR\).}
\KwOut{An estimate \(\w \in \RR^d\)}
\(t \gets 0\)\;

Initialise \(\w^t \in \RR^d, \boldsymbol{\tau}_w^t \in \RR_+^d\)\;
\(\boldsymbol{s}^{t-1} \gets \boldsymbol{0} \in \RR^n\)\;
\(\boldsymbol{F} \gets \boldsymbol{X} \odot \boldsymbol{X}\)\;

\Repeat{Termination condition}{
    \tcp{Output node update}
    \(\boldsymbol{\tau}_\omega^t \gets \boldsymbol{F} \boldsymbol{\tau}_w^t\)\;
    
    \(\boldsymbol{\omega}^t \gets \boldsymbol{X} \w^t - \boldsymbol{s}^{t-1} \odot \boldsymbol{\tau}_\omega^t\)\;

    \(P^t \gets \frac{1}{d} \w^t \Sigmadelta \w^t\)\label{algo-line:p-update}\;
    
    \(\boldsymbol{z}^t \gets \proxim{\boldsymbol{\tau}_p^t}{g(\cdot; P^t)}{\boldsymbol{\omega}^t} \)\; 
    
    \(\boldsymbol{\tau}_z^t \gets \boldsymbol{\tau}_p^t \odot \Dproxim{\boldsymbol{\tau}_p^t}{g(\cdot; P^t)}{\boldsymbol{\omega}^t}\) \;
    
    \(\boldsymbol{s}^t \gets (\boldsymbol{z}^t - \boldsymbol{\omega}^t) \oslash \boldsymbol{\tau}_p^t\)\;
    
    \(\boldsymbol{\tau}_s^t \gets (\boldsymbol{1} - \boldsymbol{\tau}_z^t \oslash \boldsymbol{\tau}_p^t) \oslash \boldsymbol{\tau}_p^t\)\;

    \tcp{Input node update}
    \(\hat{P}^t \gets \frac{\advtrainingcost}{2\sqrt{P^t}} \frac{1}{d} \boldsymbol{s}^\top \boldsymbol{y}\)\label{algo-line:phat-update} \;
    
    \(\boldsymbol{\tau}_r^t \gets \boldsymbol{1} \oslash (\boldsymbol{F}^\top \boldsymbol{\tau}_s^t)\) \;
    
    \(\boldsymbol{r}^t \gets \w^t + \boldsymbol{\tau}_r^t \odot \x^\top \boldsymbol{s}^t\) \;
      
    \(\w^{t+1} \gets \mathcal{P}_{\boldsymbol{\tau}_r^t}[r(\cdot; \hat{P}^t)] (r^t)\) \;
    
    \(\boldsymbol{\tau}_x^{t+1} \gets \boldsymbol{\tau}_r^t \odot \mathcal{P}^\prime_{\boldsymbol{\tau}_r^t}[r(\cdot, \hat{P}^t)] (r^t)\) \;
}
\end{algorithm}

In \cref{algo:adversarial-gamp}, we present the advGAMP that will be the object of the study of this subsection. The notation \(\odot\) represents the component-wise product and \(\oslash\) the component-wise division. After an initialisation, we have to update the variables alternating by output channel variables and input channel variables until some convergence condition of the variables is reached.

This algorithm can be seen as a specialisation of a GAMP algorithm where the denoising function change at each iteration because dependant on the constants \(P,\hat{P}\) that are updated in \cref{algo-line:p-update,algo-line:phat-update}. 

Enforcing these constraints at each step guarantees that the \(\hat{P}\) (respectively \cref{algo-line:phat-update}) variable will be such that they minimise the Lagrangian with respect to the primal variables. Additionally for the way the \(P\) variables are updated in \cref{algo:adversarial-gamp} (respectively \cref{algo-line:p-update}) it is guaranteed that the definition is satisfied. 
With these considerations in mind one can prove that upon convergence the $\w$ found solves the problem in~\cref{eq:app:proof-equivalent-problem}. The steps are based on interpreting the algorithm as a ADMM in a similar fashion as in \citet[Theorem~1]{rangan_gamp_admm}.

We are left with proving that the low dimensional equations are the same presented in the main text in the next subsection.

\subsection{Mapping of Saddle point Equations}

The SE equations corresponding to the GAMP algorithm defined in \cite{rangan_GAMP_2011,rangan_gamp_admm} correspond exactly the the equations in \cref{eq:final-saddle-point-eqs-appendix-proof} with the dependence on the parameters \(P,\hat{P}\).
To prove the state evolution form for the new variables one can apply Theorem~1 in \cite{javanmard_montanari_gamp_2013} to the update for \(P,\hat{P}\) in~\cref{eq:app:gradients-proof} and obtain the results in \cref{eq:app:P-update-saddle-point,eq:app:P-hat-update-saddle-point}.

To prove the form of the fixed point equations we could also map them to already proven set of fixed point equations proved in \cite{Loureiro_2022}. 

If we consider the values of the overlaps to be values we have that one can prove the equations for \(m,q,V\) and \(\hat{m},\hat{q},\hat{V}\) as being a case of a specific loss for \citet[Theorem~1]{Loureiro_2022}. For each fixed value of \(P,\hat{P}\) which will be specified afterwards. The mapping from the notation of this current paper to the other one involves a different loss function and regularisation
\begin{equation}
    g(z) \leftrightarrow g\qty(z - \advtrainingcost \sqrt{P}) \,, \qquad r(\w) \leftrightarrow r(\w) + \hat{P} \w^\top \Sigmadelta \w \,,
\end{equation}
where the first ones are the notations used in \cite{Loureiro_2022} and the second ones the one used in this paper. For fixed values of \(P,\hat{P}\) this can be seen by comparing the Lagrangian formulation in \cref{eq:lagrangian-problem-appendix} to the one in Eq.~(B.61) of \cite{Loureiro_2022}.
If we consider that the regularisation is \(\ell_2\) we have that we can apply the equations with the effective regularisation that is \(\w^\top((\lambda/ 2)\mathbb{1} + \hat{P} \Sigmadelta)\w\), under the previously stated assumptions
\begin{equation}\label{eq:final-saddle-point-eqs-appendix-proof}
\begin{aligned}
    \hat{m} &= \alpha \mathbb{E}_{\xi}\left[
        \int_{\RR} \dd{y} \partial_\omega \mathcal{Z}_0 f_g(\sqrt{q} \xi, P)
    \right] & \quad 
    m &= \mathbb{E}_{\mu} \qty[ 
        \frac{\hat{m} \bar{\theta}^2}{\lambda + \hat{V} \omega + \hat{P} \delta} 
    ] \\
    \hat{q} &= \alpha \mathbb{E}_{\xi}\left[
        \int_{\RR} \dd{y} \mathcal{Z}_0 f_g^2(\sqrt{q} \xi, P)
    \right] & \quad 
    q &= \mathbb{E}_{\mu} \qty[ 
        \frac{
            \hat{m}^2 \bar{\theta}^2 \omega + \hat{q} \omega^2
        }{
            (\lambda + \hat{V} \omega + \hat{P} \delta)^2
        } 
    ] \\
    \hat{V} &= -\alpha \mathbb{E}_{\xi}\left[
        \int_{\RR} \dd{y} \mathcal{Z}_0 \partial_\omega f_g(\sqrt{q} \xi, P)
    \right] & \quad 
    V &= \mathbb{E}_{\mu} \qty[ 
        \frac{\omega}{\lambda + \hat{V} \omega + \hat{P} \delta } 
    ]
\end{aligned}
\end{equation}
where we have the same definitions for \(\mathcal{Z}_0\) and \(f_g\) as in the \cref{thm:main-theorem-saddle-point-eqs}.

At optimality we want the gradients in \cref{eq:app:gradients-proof} to be equal to zero. Thus these equations should be considered as equalities to zero also in the limit. To find the limiting form of these equations we would like to apply Theorem~5 of \cite{Loureiro_2022}.
We can start from the condition for the gradients of the dual variables. The function satisfies the assumptions of the theorem and thus can be applied. To obtain the specific form 
\begin{equation}\label{eq:app:P-update-saddle-point}
    P = \mathbb{E}\qty[
        \zeta \frac{
            \hat{m}^2 \bar{\theta}^2 + \hat{q} \omega
        }{
            (\lambda + \hat{V} \omega + \hat{P} \delta )^2
        } 
    ] 
\end{equation}

For the other two equations we can remember that at optimality the value of \(\boldsymbol{s}\) is connected to the proximal operator and the function \(f_g\). We have that these functions are pseudo-Lipschitz and thus satisfy the assumptions of the previously applied theorem. Thus the optimality conditions read in the limit are
\begin{equation}\label{eq:app:P-hat-update-saddle-point}
\begin{aligned}
    \hat{P} &= \alpha \frac{\advtrainingcost}{2 \sqrt{P}} \mathbb{E}_{\xi}\qty[ \int \dd{y} \mathcal{Z}_0 y f_g (\sqrt{q} \xi, P, \advtrainingcost) ] \,, 
\end{aligned}
\end{equation}
thus proving the set of equations.

\subsection{Form for the overlaps \(A\) and \(F\)}

Similarly to before we want to find the high dimensional form for
\begin{equation}
    A = \frac{1}{d} \what^\top \Sigmaupsilon \what \,, \quad F = \frac{1}{d} \what^\top \Sigmaupsilon \wstar
\end{equation}

Again we can leverage \citet[Theorem~5]{Loureiro_2022} or \citet[Theorem~1]{javanmard_montanari_gamp_2013} to obtain in both cases
\begin{equation}
    A = \mathbb{E}_{\mu} \qty[ \upsilon \frac{\hat{m}^2 \bar{\theta}^2 \omega + \hat{q} \omega^2}{(\lambda + \hat{V} \omega + \hat{P} \delta )^2} ] \,, \quad 
    F = \mathbb{E}_{\mu} \qty[ \frac{\hat{m}  f  \bar{\theta}}{\lambda + \hat{V} \omega + \hat{P} \delta }]
\end{equation}
as explained before.

\subsection{Interpretation of the Result}

The parameters \(m, q, P, N, A\) and \(F\) appearing in the previous sections have a simple interpretation, they are the values around which the teacher-student and student-student overlaps concentrate in high dimension 
\begin{equation}\label{eq:app:interpretation-overlaps}
\begin{aligned}
    m &= \frac{1}{d} \mathbb{E}_{\mathcal{D}}\qty[\wstar^\top \Sigmax \what] \,, & \  q &= \frac{1}{d} \mathbb{E}_{\mathcal{D}}\qty[\what^\top \Sigmax \what]  \,, &\
    P &= \frac{1}{d} \mathbb{E}_{\mathcal{D}}\qty[\what^\top \Sigmadelta \what] \,, \\ 
    A &= \frac{1}{d} \mathbb{E}_{\mathcal{D}}\qty[\what^\top \Sigmaupsilon \what] \,, &\  F &= \frac{1}{d} \mathbb{E}_{\mathcal{D}}\qty[\wstar^\top \Sigmaupsilon \what] \,. & &
\end{aligned}
\end{equation}

The overlap \(m\) describes the angle between student estimate \(\what\) and the ground truth teacher vector \(\wstar\), the overlap \(q\) represents the data-weighted norm of the weights, \(P\) quantifies the norm of the weights scaled by the defence direction, \(A\) quantifies how much the student lies in the attack geometry and \(F\) quantifies the overlap between teacher and student in the attack geometry.
Note that contrary to \cref{eq:saddle-point-channel-main-thm,eq:saddle-point-prior-main-thm}, these expressions cannot be used to  efficiently obtain sufficient statistics as they depend on average over dataset realisation of the trained weights \(\what\).

It is important to note that all the summary statistics involved in the statement of the theorem are finite-dimensional as the dimension increases, and therefore the result is a fully asymptotic characterisation, in the sense that it does not involve any high-dimensional object. With this theorem, we can avoid solving \cref{eq:modified-minimisation-problem} (a high dimensional problem) and instead solve \cref{eq:saddle-point-channel-main-thm,eq:saddle-point-prior-main-thm} (eight dimensional problem): all quantities of interest can be expressed of scalar parameters/sufficient statistics that concentrate in the high-dimensional limit.

\section{ERROR METRICS}\label{sec:error-metrics-app}

As is common in machine learning, we want to see how the model trained performs on different metrics. This section is devoted to defining the metrics of our interest and expressing them as a function of the overlaps solutions of the fixed-point equations in \cref{thm:main-theorem-saddle-point-eqs}. In general, we distinguish between the strength of the training attack and indicate it as \(\advtrainingcost\) and the strength of the actual attacker considered in generalisation, which we call \(\advgencost\).
In each subsection, we provide the formula to compute the value from the overlap solution of \cref{thm:main-theorem-saddle-point-eqs} or directly from a data set generated with the distribution explained in \cref{sec:setting-specification}.

Some of these metrics will be computed using the method of ``local fields'' where one can suppose the jointly Gaussian behaviour for \((\what^\top \x / \sqrt{d}, \wstar^\top \x / \sqrt{d})\) with mean zero and covariance \(\boldsymbol{\sigma} = \bigl( \begin{smallmatrix}\rho & m \\ m & q\end{smallmatrix}\bigr)\). We will refer to the following probability distribution
\begin{equation}\label{eq:local-field-defintion-appendix}
    \dd{\mu}(\nu, \lambda) = \frac{1}{2\pi \sqrt{\det \boldsymbol{\sigma}}} \exp\qty( -\frac{1}{2} 
    \begin{pmatrix}
       \nu \\
       \lambda
     \end{pmatrix}^\top \boldsymbol{\sigma}^{-1} \begin{pmatrix}
       \nu \\
       \lambda
     \end{pmatrix} 
    )
\end{equation}
This method is explained at length in \cite{Clarte_uncertanty} and applies also in the case considered here. For other computations we refer to the places where the computation can be found in detail.

\subsection{Generalisation}

In machine learning, particularly in the context of adversarial training, we are concerned with how well our model, referred to as the student model, can make correct predictions on data that has not been altered or perturbed. We quantify this ability using a metric called the generalisation error. The generalisation error, denoted by \(\generr\), is the expected value over all possible data points of whether the model's prediction \(\hat{y}\) is incorrect. When the model's predictions are based on the estimated parameter vector \(\what\), the generalisation error is mathematically expressed as
\begin{equation}\label{eq:standard-generalisation-error-erm}
    \generr = \EEb{y, \x}{\mathbb{1}[y \neq \hat{y}(\hat{\w} ; \x)]}
\end{equation}
This can also be represented in terms of overlaps using
\begin{equation}\label{eq:standard-generalisation-error-overlap}
    \generr = 
    \int \dd{y} \dd{\mu}(\nu, \lambda) P(y \mid \lambda) \mathbb{1}\qty[y \neq f(\nu)] = 
    \frac{1}{\pi}\arccos\qty(\frac{m}{\sqrt{ (\rho + \tau^2 ) q}})
\end{equation}
where this integral can be simplified further in the case of simple models, \textit{e.g.} the case of a noiseless channel.
The final form in the case of in the noiseless case has been presented in \cite{aubin_generalization_2020} Appendix~II. For the noisy case the derivation can be found in \cite{Clarte_uncertanty,clarte_double_descent}. 
The general form can be found in \cite[Appendix~D]{Gerace_2021}.

However, in the case of adversarial learning, we're not only interested in the generalisation error under normal conditions but also under adversarial attacks. 
The adversarial generalisation error measures the model's robustness against such attacks by evaluating the probability of misclassification when the input data is perturbed within a certain norm bound determined by \(\advgencost\) and the covariance matrix \(\Sigmaupsilon\). 

Formally, the adversarial generalisation error is given by
\begin{equation}\label{eq:adversarial-generalisation-error-erm}
    \advgenerr = 
    \EEb{y, \x}{\max_{
        \norm{\vecdelta}_{\Sigmaupsilon^{-2}} \leq \advgencost
    } \mathbb{1}[y \neq \hat{y}(\hat{\w}(\alpha) ; \x + \vecdelta)]} 
\end{equation}
where the inner maximisation, for a fixed choice of \(\x\) and \(\what\) can be  solved explicitly.

Again by using the idea of local fields one can compute the value of the adversarial error as a function of the overlaps. The definitions of the overlap parameters \(A, F\) is fixed after training and thus it doesn't get included in the average. We have that the form of the error is 
\begin{equation}\label{eq:adversarial-generalisation-error-overlap}
\begin{aligned}
    \advgenerr =& \int_{0}^{\infty} \erfc\left(\frac{\frac{m}{\sqrt{q}} \xi}{\sqrt{2\left(\rho + \tau^2 - m^2 / q\right)}}\right) \frac{e^{-\frac{\xi^2}{2}}}{\sqrt{2 \pi}} \dd{\xi} \\
    +& \int_{0}^{\advgencost \frac{\sqrt{A}}{\sqrt{q}}} \erfc\left(\frac{ -\frac{m}{\sqrt{q}} \xi}{\sqrt{2\left(\rho + \tau^2 - m^2 / q\right)}}\right) \frac{e^{-\frac{\xi^2}{2}}}{\sqrt{2 \pi}} \dd{\xi} \\ 
    =& \Egen + 
    \int_{0}^{\advgencost \frac{\sqrt{A}}{\sqrt{q}}} \erfc\left(\frac{ -\frac{m}{\sqrt{q}} \xi}{\sqrt{2\left(\rho + \tau^2 - m^2 / q\right)}}\right) \frac{e^{-\frac{\xi^2}{2}}}{\sqrt{2 \pi}} \dd{\xi}
\end{aligned}
\end{equation}

One could modify the equation even more to relate it to the Owen's T function as we show in \cref{sec:adversarial-owen-function}.

\subsection{Training}

The training error reflects the model's performance on the dataset it was trained on. The goal during the training phase is to minimise this error and we can expect to reach an optimal zero training error for the estimator \(\what\) on noise less dataset. The training error is expressed as
\begin{equation}\label{eq:training-error-erm}
    \trainerr = \sum_{\datidx=1}^{n} \mathbb{1}\qty[ \hat{y}(\what; \x_\datidx) \neq y_\datidx ]
\end{equation}
where the sum is over the same dataset used to find \(\what\).

Following \citet[Appendix~D.2]{Gerace_2021}, one can find the following form of the training error as a function of overlaps as
\begin{equation}\label{eq:training-error-overlap}
    \trainerr = \frac{1}{2} \mathbb{E}_{\xi}\left[\int \dd{y} \mathcal{Z}_0 \qty(y, \sqrt{\eta} \xi, \rho-\rho\eta) \mathbb{1}\qty[ \sign\qty( \mathcal{P}_{V g(\cdot; y, P) }(\sqrt{q} \xi) ) \neq y ] \right] 
\end{equation}
where \(\eta = \frac{m}{\sqrt{\rho q}}\).

Arguably more interesting than the training error is the training loss since it is a part of the actual risk to minimise. This metric also brings more information than the training error since usually the losses used for classification are also sensitive to the norm of the solution and not only the direction of \(\what\). The definition as a function of the predicted ERM weights is
\begin{equation}\label{eq:training-loss-erm}
    \trainloss = \sum_{\datidx=1}^{n} g\qty(y_\datidx \frac{\what^\top \x_\datidx}{\sqrt{d}})
\end{equation}
and as a function of overlaps
\begin{equation}\label{eq:training-loss-overlap}
    \trainloss = \mathbb{E}_{y, \xi} \left[
        \mathcal{Z}_0 \qty(y, \sqrt{\eta} \xi, \rho-\rho\eta) \,
        g\left( y, \mathcal{P}_{V g(\cdot; y, P) }(\sqrt{q} \xi), \advtrainingcost \sqrt{P}  \right) 
    \right] 
\end{equation}

\subsection{Expression for Teacher Usefulness and Robustness}

The equations for the usefulness and the robustness can be found using the same approach as before. We remember assumption \textbf{(A1)} and \textbf{(A4)} of \cite{Loureiro_2022} in Appendix~B for the assumptions on the teacher. We explicitly have that
\begin{equation}
\begin{aligned}
    \usefulmetric_{\wstar} &= \frac{1}{\sqrt{d}} \mathbb{E}_{\x, y}[y \wstar^\top \x ] = \sqrt{\frac{2}{\pi}} \frac{\rho}{\sqrt{\rho + \tau^2}} \,, \\
    \robustmetric_{\wstar} &= \frac{1}{\sqrt{d}} \EEb{\x, y}{
        \inf_{\norm{\vecdelta}_{\Sigmaupsilon^{-2}}  \leq \advgencost} y \wstar^\top ( \x + \vecdelta)
    }  \\
    &= \frac{1}{\sqrt{d}} \mathbb{E}_{\x, y}[y \wstar^\top \x ] - \frac{\advgencost}{\sqrt{d}} \mathbb{E}\qty[\sqrt{\w^\top \Sigmaupsilon \w}] = \usefulmetric_{\wstar} - \advgencost \sqrt{a} \,. 
\end{aligned}
\end{equation}
and in the end we have integrated using the local fields method in the case of averaged teacher.

\section{LARGE SAMPLE COMPLEXITY ASYMPTOTICS}
\label{sec:large_alpha}

In this Appendix, we provide expansion of the fixed point equations in \cref{thm:main-theorem-saddle-point-eqs}.
In practical terms, the high sample complexity regime is challenging to access due to the computational resources it demands to simulate the ERM. By providing a theoretical expansion, we offer a lens through which the system's behaviour under high sample complexity can be understood and predicted at less computational cost. 

\subsection{Self Consistent equations in the large \texorpdfstring{$\alpha$}{sample complexity} limit}

We consider the following behaviour as a function of \(\alpha\) for the order parameters
\begin{equation}\label{eq:high_alpha_expansion_tau_small_epsilon_greater_zero}
\begin{aligned}
    & q \underset{\alpha \rightarrow \infty}{=} q_0\,,& & \hat{q} \underset{\alpha \rightarrow \infty}{=} \hat{q}_0 \alpha\,, \\
    & m \underset{\alpha \rightarrow \infty}{=} m_0\,, & & \hat{m} \underset{\alpha \rightarrow \infty}{=} \hat{m}_0 \alpha\,, \\
    & V \underset{\alpha \rightarrow \infty}{=} \frac{V_0}{\alpha}\,, & & \hat{V} \underset{\alpha \rightarrow \infty}{=} \hat{V}_0 \alpha\,, \\
    & P \underset{\alpha \rightarrow \infty}{=} P_0\,, & & \hat{P} \underset{\alpha \rightarrow \infty}{=} \hat{P}_0 \alpha\,, \\
    & A \underset{\alpha \rightarrow \infty}{=} A_0\,, & & \hat{A} \underset{\alpha \rightarrow \infty}{=} 0\,, \\
    & F \underset{\alpha \rightarrow \infty}{=} F_0 \,, & & \hat{F} \underset{\alpha \rightarrow \infty}{=} 0 \,,
\end{aligned}
\end{equation} 
where the sub scripted quantities are independent of \(\alpha\). We remark that this scaling is the same found in \cite{vilucchio2024asymptotic}.

These scaling assumptions break down whenever $\advtrainingcost = \tau = 0$, in this case, the first row of overlaps except $V$ scale linearly in $\alpha$. This case is already studied in \cite{aubin_generalization_2020}. 
This specific scaling ansatz is justified not only because of consistency of the saddle point equations but also because it is the one found empirically in the study of the phenomena.

The study of the proximity operator for the logistic loss function \(f(v) : v \mapsto \log(1+\exp(-v))\) has been studied in \cite[Section~3.3]{briceno2019random}. Explicitly from \cite[Propositon~2]{briceno2019random} one has that
\begin{equation}
    \mathcal{P}_{Vf(\cdot)} (\omega) = \omega + \mathbf{W}_{\exp(-\omega)} (V \exp(-\omega))
\end{equation}
where specifically \(\mathbf{W}\) is the generalised Lambert function that satisfies
\begin{equation}
    (\forall \bar{v} \in \mathbb{R})(\forall v \in \mathbb{R})(\forall r \in] 0,+\infty[) \quad \bar{v}(\exp (\bar{v})+r)=v \quad \Leftrightarrow \quad \bar{v}=\mathbf{W}_r(v) .
\end{equation}

Translating this result for our loss function \(g(y,\cdot; \advtrainingcost, P)\) gives us the following proximal operator
\begin{equation}
    \mathcal{P}_{V g(\cdot; y, P)}(\sqrt{q} \xi) = \sqrt{q} \xi + y \mathbf{W}_{\exp ( - y \sqrt{q} \xi + \advtrainingcost \sqrt{P} )} \qty( V \exp ( - \sqrt{q} \xi + \advtrainingcost \sqrt{P} )) 
\end{equation}

To simplify the proximal operator we start by its formulation then expand 
\begin{equation}
    \mathcal{P}_{V g(\cdot; y, P)}(\sqrt{q} \xi) \underset{\alpha \rightarrow \infty}{=} \sqrt{q_0} \xi + \frac{c(\xi)}{\alpha} \,, \qquad 
    c(\xi) = y V_0 \frac{\exp( -y \sqrt{q_0} \xi + \advtrainingcost \sqrt{P_0} ) }{ 1 + \exp( -y \sqrt{q_0} \xi + \advtrainingcost \sqrt{P_0} ) } \,.
\end{equation}

With this, we can simplify the channel and the prior equations as follows where we remember that \(\eta_0 = m_0^2 / q_0\) and \(\hat{\eta} = \hat{m}^2 / \hat{q}\)
\begin{equation}\label{eq:channel-saddle-point-equation-large-alpha}
    \begin{cases}
        \hat{m}_0   &= \frac{1}{V_0} \mathbb{E}_{y, \xi}\left[
            \partial_\omega \mathcal{Z}_0(y, \sqrt{\eta_0} \xi, \rho - \eta_0) 
            c(\xi) 
        \right] \\
        \hat{q}_0 &= \frac{1}{V_0^2} \mathbb{E}_{y, \xi}\left[
            \mathcal{Z}_0(y, \sqrt{\eta_0} \xi, \rho - \eta_0) 
            c(\xi)^2 
        \right]\\
        \hat{V}_0 & = \mathbb{E}_{y, \xi}\left[
            \mathcal{Z}_0(y, \sqrt{\eta_0} \xi, \rho - \eta_0)
            \partial^2 g (y, \sqrt{q_0} \xi, P_0, \advtrainingcost) )
        \right] \\ 
        \hat{P}_0 &= \frac{\advtrainingcost }{\sqrt{N_0} V_0} \mathbb{E}_{y, \xi}\left[
            \mathcal{Z}_0(y, \sqrt{\eta_0} \xi, \rho - \eta_0)
            y c(\xi)
        \right] 
    \end{cases}
\end{equation}
\begin{equation}\label{eq:prior-saddle-point-equation-large-alpha}
    \begin{cases}
        m_0 & =  \frac{1}{d} \tr \qty[ 
            \hat{m}_0 \Sigmax^{\top} \boldsymbol{\theta}_0 \boldsymbol{\theta}_0^{\top} \Sigmax \boldsymbol{\Lambda}_0^{-1} 
        ] \\
        q_0 & =  \frac{1}{d} \tr \qty[
            \hat{m}_0^2 \Sigmax^{\top} \boldsymbol{\theta}_0 \boldsymbol{\theta}_0^{\top} \Sigmax \Sigmax \boldsymbol{\Lambda}_0^{-2}
        ] \\
        V_0 & =  \frac{1}{d} \tr \qty[
            \Sigmax \boldsymbol{\Lambda}_0^{-1}
        ] \\
        P_0 & =  \frac{1}{d} \tr \qty[
            \hat{m}_0^2 \Sigmax^{\top} \boldsymbol{\theta}_0 \boldsymbol{\theta}_0^{\top} \Sigmax \Sigmadelta \boldsymbol{\Lambda}_0^{-2}
        ] 
    \end{cases}
\end{equation}
with 
\(\boldsymbol{\Lambda}_0 = \lambda_1 \mathbb{1} + \hat{V}_0 \Sigmax + \hat{P}_0 \Sigmadelta\) where \(\lambda = \lambda_1 \alpha\). 
Additional explanation on this behaviour can be found in \citet[Appendix~D]{vilucchio2024asymptotic}.

The additional values are
\begin{equation}
    A_0 = \frac{1}{d} \tr \qty[
        \hat{m}_0^2 \Sigmax^{\top} \boldsymbol{\theta}_0 \boldsymbol{\theta}_0^{\top} \Sigmax \Sigmaupsilon \boldsymbol{\Lambda}_0^{-2}
    ] \,, \qquad
    F_0 = \frac{1}{d} \tr \qty[
        \hat{m}_0^2 \Sigmax^{\top} \boldsymbol{\theta}_0 \boldsymbol{\theta}_0^{\top} \Sigmaupsilon \boldsymbol{\Lambda}_0^{-1}
    ] \,.
\end{equation}

This formulation can be evaluated using the set of equations in \cref{thm:main-theorem-saddle-point-eqs} upon convergence.

\subsection{Large \texorpdfstring{$\alpha$}{sample complexity} limit of \texorpdfstring{$\advgenerr$}{the adversarial test error}}\label{sec:large-alpha-limit-of-adverr}

Using above simplifications, we can simplify the following quantities in the large sample complexity limit.

We can write the following
\begin{equation}
\begin{aligned}
    \lim_{\alpha \to \infty} \frac{m}{\sqrt{ (\rho + \tau^2 ) q}} & = 
    \frac{  
        \tr \qty[  \Sigmax^{\top} \boldsymbol{\theta}_0 \boldsymbol{\theta}_0^{\top} \Sigmax \boldsymbol{\Lambda}_0^{-1} ]  
    }{  
        \sqrt{  (\wstar^\top \Sigmax \wstar + \tau^2)  \tr \qty[  \Sigmax^{\top} \boldsymbol{\theta}_0 \boldsymbol{\theta}_0^{\top} \Sigmax  \Sigmax  \boldsymbol{\Lambda}_0^{-2} ] }   
    }
\end{aligned}
\end{equation}
and the overlap ratio determining how strong an attack is, is given by
\begin{equation}
\begin{aligned}
    \lim_{\alpha \to \infty} \frac{\sqrt{A}}{\sqrt{q}} &= 
    \frac{
        \sqrt{
            \tr \qty[  \Sigmax^{\top} \boldsymbol{\theta}_0 \boldsymbol{\theta}_0^{\top} \Sigmax  \Sigmaupsilon \boldsymbol{\Lambda}_0^{-2} ]  
        }
    }{ 
        \sqrt{
            \tr \qty[ \Sigmax^{\top} \boldsymbol{\theta}_0 \boldsymbol{\theta}_0^{\top} \Sigmax  \Sigmax \boldsymbol{\Lambda}_0^{-2} ] 
        }  
    } 
\end{aligned}
\end{equation}

Notably, the previous equations no longer depend on \(\hat{q}_0\) and \(\hat{m}_0\), with this the adversarial generalisation error is characterised by the following equation, which is a function of the limiting values we just defined.
Nonetheless the previous equation still depends on the values of \(\hat{V}_0, \hat{P}_0\) through \(\boldsymbol{\Lambda}_0\).
The values are constants evaluated at the fixed point of \cref{eq:channel-saddle-point-equation-large-alpha,eq:prior-saddle-point-equation-large-alpha}.

We also want to show that the plateau is reached with a derivative that is going to zero. To do so we first look at the derivative of the angle variables \(m/\sqrt{\rho q}\) and \(\sqrt{A} / \sqrt{q}\).

Specifically we have that
\begin{equation}
\begin{aligned}
    \frac{m}{\sqrt{\rho q}} = & 
    \frac{ 
        \tr \qty[ \hat{m}_0 \Sigmax^{\top} \boldsymbol{\theta}_0 \boldsymbol{\theta}_0^{\top} \Sigmax \boldsymbol{\Lambda}_0^{-1} ] 
    }{
        \sqrt{\wstar^\top \Sigmax \wstar \tr \qty[  \qty( \hat{m}_0^2 \Sigmax^{\top} \boldsymbol{\theta}_0 \boldsymbol{\theta}_0^{\top} \Sigmax + \frac{\hat{q}_0}{\alpha}  \Sigmax ) \Sigmax \boldsymbol{\Lambda}_0^{-2}  ] }  
    } \\
    \frac{\sqrt{A}}{\sqrt{q}} =& 
    \frac{ 
        \sqrt{
            \tr \qty[ \qty( \hat{m}_0^2 \Sigmax^{\top} \boldsymbol{\theta}_0 \boldsymbol{\theta}_0^{\top} \Sigmax + \frac{\hat{q}_0}{\alpha}  \Sigmax ) \Sigmaupsilon \boldsymbol{\Lambda}_0^{-2} ]  
        }
    }{
        \sqrt{ \tr \qty[ \qty( \hat{m}_0^2 \Sigmax^{\top} \boldsymbol{\theta}_0 \boldsymbol{\theta}_0^{\top} \Sigmax + \frac{\hat{q}_0}{\alpha}  \Sigmax ) \Sigmax \boldsymbol{\Lambda}_0^{-2}  ] }  
    }  \\
\end{aligned}
\end{equation}

To compute the derivative, we use the additivity of the trace. Note that we compute the derivative with respect to a scalar that is not dependent on the trace itself. Thus, every derivative in both terms will lead to a factor of the form \(\frac{c}{\alpha^2} \tr(M)\), which goes to zero as \(\alpha \to \infty\).

Thus the adversarial generalisation error approaches a constant in the large sample complexity limit
\begin{equation}
\begin{aligned}
    \partial_{\alpha} \advgenerr \underset{\alpha \rightarrow \infty}{=} & 0 , \quad \advgenerr \underset{\alpha \rightarrow \infty}{=} & cst.\
\end{aligned}
\end{equation}

\subsection{Specific Case of Single Block Model}\label{sec:single-block-model-high-alpha}

In this subsection we consider the case where it only exists one kind of feature and we show that this will lead to results that are independent of adversarial training in the \(\alpha \to \infty\) limit.

Specifically we consider the fact of having a Block Feature Data Model as explained in \cref{sec:strong-weak-feature-model} with a single block. We note that the data model of \cite{javanmard_precise_tradeoffs_2020} fall under this category.

In this specific case the self consistent equations in \cref{eq:prior-saddle-point-equation-large-alpha} simplify as
\begin{equation}
\begin{aligned}
    m_0 &= \frac{\hat{m}_0 \psi^2 t}{\lambda_1 + \hat{V}_0 \psi} \,, \quad &
    q_0 &= \frac{\hat{m}_0^2 \psi^3 t}{(\lambda_1 + \hat{V}_0 \psi)^2} \,, \quad &
    V_0 &= \frac{1}{\lambda_1 + \hat{V}_0 \psi} \,, \quad  \\
    P_0 &= \frac{\hat{m}_0^2 \psi^2 \Delta t}{(\lambda_1 + \hat{V}_0 \psi)^2} \,, \quad &
    A_0 &= \frac{\hat{m}_0^2 \psi^2 t \Upsilon}{(\lambda_1 + \hat{V}_0 \psi)^2} \,, &
    F_0 &= \frac{\hat{m}_0 \psi t \Upsilon}{\lambda_1 + \hat{V}_0 \psi} \,.
\end{aligned}
\end{equation}

From the analysis done in the previous section we have that
\begin{equation}
    \lim_{\alpha \to \infty} \frac{m}{\sqrt{(\rho + \tau^2) q}} = \frac{\psi^2 t}{\sqrt{(\psi t + \tau^2) \psi^3 t}} \,, \quad 
    \lim_{\alpha \to \infty} \frac{\sqrt{A}}{\sqrt{q}} = \frac{\psi^2 t \Upsilon}{\sqrt{\psi^2 t \psi^3 t}} \,.
\end{equation}
We notice that these equations are independent from the value of \(\advtrainingcost\).

By writing the system of saddle point equations to next leading order, one could solve for the next leading order of the overlap parameters \((m_1, q_1, V_1, P_1)\). 
In general these do depend on the value of \(\advtrainingcost\) but by the integer scaling of all the analytic functions we have that the overlap parameters at that order differ of an order \(1/ \alpha\). And the same order of difference is there between the terms of \(\generr\) and \(\bounderr\).

We have that the second leading order can be bounded by a constant times \(1/\alpha\) thus the claim follows.

\subsection{Change of Defence Direction}\label{sec:change-defence-direction}

We can rewrite the adversarial error as
\begin{equation}
    \advgenerr = 
    {
        \frac{1}{\pi} \arccos\qty(
            \frac{\vartheta}{\sqrt{ (1 + \frac{\tau^2}{\rho}) }}
        ) + 
        \int_{0}^{\advgencost u} 
        \erfc\qty(
            \frac{ - \vartheta \nu}{\sqrt{2 \qty((1 + \frac{\tau^2}{\rho}) - \vartheta^2 )}} 
        ) \frac{e^{-\frac{1}{2}\nu^2}}{\sqrt{2 \pi}} \dd{\nu}
    }
\end{equation}
where \(\vartheta = m/\sqrt{\rho q}\) and \(u = \sqrt{A} / \sqrt{q}\) as there is no dependance on the result of the saddle point equations.

And the formal power series is given by
\begin{equation}\label{eq:first-order-expansion-gen-error}
    \generr = \frac{1}{\pi} \arccos\qty(
        \frac{\vartheta(0)}{\sqrt{ (1 + \frac{\tau^2}{\rho}) }}
    ) -
    \frac{\vartheta^\prime(0)}{\pi  \sqrt{1-\vartheta (0)^2}} \varrho + \order{\varrho^2}
\end{equation}
\begin{equation}\label{eq:first-order-expansion-boundary-error}
\begin{aligned}
    \bounderr &= \int_{0}^{\advgencost u(0)} 
    \erfc\qty(
        \frac{ - \vartheta(0) \nu}{\sqrt{2 \qty((1 + \frac{\tau^2}{\rho}) - \vartheta(0)^2 )}} 
    ) \frac{e^{-\frac{1}{2}\nu^2}}{\sqrt{2 \pi}} \dd{\nu} \\
    &+ \advgencost \frac{ e^{-\frac{1}{2} u(0)^2 \advgencost^2} }{\sqrt{2 \pi}} u^\prime(0) 
    \erfc\qty(
        -\frac{\vartheta(0) u(0) \advgencost }{\sqrt{2-2 \vartheta(0)^2}}
    ) \varrho - \frac{
        e^{-\frac{u(0)^2 \advgencost ^2}{2 (1 - \vartheta(0)^2)}} - 1
    }{\pi \sqrt{1-\vartheta(0)^2}} \theta^\prime(0) \varrho + \order{\varrho^2}
\end{aligned}
\end{equation}

We now need to consider the sign of the derivatives of \(u^\prime(0)\) and \(\vartheta^\prime(0)\). The main point is that one can expand to get the derivatives as a function of the solution.

We consider the case of two features as in \cref{prop:defending-less-robust}. We also suppose that we have an attack that is \(\Sigmaupsilon = \mathbb{1}\). We also suppose that \(\psi_1 > \psi_2\) without loss of generality.
\begin{equation}
    \frac{m^2}{\rho q} = \frac{
        \qty(\phi_1 \psi_1^2 b + \phi_2 \psi_2^2 a )^2
    }{
        (\phi_1 \psi_1 + \phi_2 \psi_2) (\phi_1 \psi_1^3 b^2 + \phi_2 \psi_2^3 a^2 ) 
    } 
    \qquad
    \frac{A}{\sqrt{q}} = \frac{
        \phi_1 \psi_1^2 b^2 + \phi_2 \psi_2^2 a^2 
    }{
        \sqrt{\phi_1 \psi_1^3 b^2 + \phi_2 \psi_2^3 a^2}
    }
\end{equation}
where \(a = \lambda + \hat{V} \psi_1 + \hat{P} (1 + \delta_1 \varrho) + \hat{N} \) and \(b = \lambda + \hat{V} \psi_2 + \hat{P} (1 + \delta_2 \varrho) + \hat{N} \).

By the use of the implicit function theorem \citep{krantz2002implicit} one has that 
the any overlap parameter $u$ solution in the limit \(\varrho \to 0\) can be written as
\(u = u_0 + u_1 \varrho + \order{\varrho^2}\), where $u_0$ is the same overlap that is solution for \(\varrho = 0\). 
Thus we can expand and thus at first order the following
\begin{equation}
    \frac{m}{\sqrt{\rho q}} = \frac{m_0}{\sqrt{\rho q_0}} + 
    \frac{
        \mathcal{N}_\vartheta
   }{
        \mathcal{D}_\vartheta
   } \varrho + \order{\varrho^2} \,, \quad 
   \frac{A}{\sqrt{q}} = \frac{A_0}{\sqrt{q_0}} + 
    \frac{
        \mathcal{N}_u
   }{
        \mathcal{D}_u
   } \varrho + \order{\varrho^2} \,, \quad 
\end{equation}
where we have that both \(\mathcal{D}_\vartheta > 0\) and \(\mathcal{D}_u > 0\) independently of the values \( \psi_1, \psi_2 >0\) and \(\delta_1, \delta_2\). 
Additionaly we have
\begin{equation}
\begin{aligned}
    \mathcal{N}_\vartheta &= \hat{P}_0 t_1 t_2 \psi_1^2 \psi_2^2 (\Delta_2 \psi_1-\Delta_1 \psi_2) (-\Delta_1 \hat{P}_0 \hat{V}_1 \psi_2+\Delta_2 \hat{P}_0 \hat{V}_1 \psi_1+\Delta_1 \hat{P}_1 \hat{V}_0 \psi_2+\Delta_2(-\hat{P}_1) \hat{V}_0 \psi_1) \\
    &-\hat{P}_0 t_1 t_2 \psi_1^2 \psi_2^2 (\Delta_2 \psi_1-\Delta_1 \psi_2) \left(\Delta_1 \hat{P}_0^2+\hat{P}_0 \hat{V}_0 \psi_1\right)\delta_2 \\
    &+ \hat{P}_0 t_1 t_2 \psi_1^2 \psi_2^2 (\Delta_2 \psi_1-\Delta_1 \psi_2) \left(\Delta_2 \hat{P}_0^2+\hat{P}_0 \hat{V}_0 \psi_2\right) \delta_1
\end{aligned}
\end{equation}
and
\begin{equation}
\begin{aligned}
    \mathcal{N}_u &= t_1 t_2 \psi_1^2 \psi_2^2 (\psi_2-\psi_1) (\Delta_1 \hat{P}_0+\hat{V}_0 \psi_1) (\Delta_2 \hat{P}_0+\hat{V}_0 \psi_2) (\Delta_1 \hat{P}_0 \hat{V}_1 \psi_2-\Delta_2 \hat{P}_0 \hat{V}_1 \psi_1-\Delta_1 \hat{P}_1 \hat{V}_0 \psi_2+\Delta_2 \hat{P}_1 \hat{V}_0 \psi_1) \\
    &- t_1 t_2 \psi_1^2 \psi_2^2 (\psi_1-\psi_2) (\Delta_1 \hat{P}_0+\hat{V}_0 \psi_1) \left(\Delta_1 \hat{P}_0^2+\hat{P}_0 \hat{V}_0 \psi_1\right) (\Delta_2 \hat{P}_0+\hat{V}_0 \psi_2) \delta_2 \\
    &+ t_1 t_2 \psi_1^2 \psi_2^2 (\psi_1-\psi_2) (\Delta_1 \hat{P}_0+\hat{V}_0 \psi_1) (\Delta_2 \hat{P}_0+\hat{V}_0 \psi_2) \left(\Delta_2 \hat{P}_0^2+\hat{P}_0 \hat{V}_0 \psi_2\right) \delta_1
\end{aligned}
\end{equation}

We see that for the term \(\mathcal{N}_u\) we have that the sign of the coefficients in front of \(\delta_1\) and \(\delta_2\) have the same sign as \(\pm(\psi_1 - \psi_2)\) respectively.

For the therm \(\mathcal{N}_\vartheta\) the analysis is a little bit more complicated. 

In the case where \(\Delta_1 = \Delta_2 = 1\) we see from this that \(\psi_1 \gtrless \psi_2\) and we chose \(\delta_1 \lessgtr \delta_2\) we have that both of the numerators are positive.
\begin{equation}
    \Delta_2 \psi_1 - \Delta_1 \psi_2 \geq 0 
\end{equation}
if this condition is satisfied then the claim holds, otherwise not. Take into consideration that as long as \(\Delta_2 \geq \Delta_1\) this is also the case.

\subsection{Adversarial Error and Owen's T function}\label{sec:adversarial-owen-function}

We demonstrate that the adversarial error from \cref{eq:adversarial-generalisation-error-overlap} can be written in terms of Owen's T function, we begin by explicitly stating the adversarial error. We believe that given the usual definition of the Owen's T function this is a interesting geometrical approach.

We start from the definition of the adversarial error
\begin{equation}\label{eq:scalar_case_adversarial_generalization_error}
\begin{aligned}
    \advgenerr(\advgencost) &= \int_{0}^{\infty} \erfc\left(\frac{\frac{m}{\sqrt{q}} \xi}{\sqrt{2\left(\rho + \tau^2 - m^2 / q\right)}}\right) \frac{e^{-\frac{\xi^2}{2}}}{\sqrt{2 \pi}} \dd{\xi} \\
    &+ \int_{0}^{\advgencost \frac{\sqrt{A}}{\sqrt{q}}} \erfc\left(\frac{ -\frac{m}{\sqrt{q}} \xi}{\sqrt{2\left(\rho + \tau^2 - m^2 / q\right)}}\right) \frac{e^{-\frac{\xi^2}{2}}}{\sqrt{2 \pi}} \dd{\xi}
\end{aligned}
\end{equation}
where we did a change of variable with an opposite sign and we changed the order of integration. 

Now we would like to use some identities to deal with the non-adversarial part of the generalisation error. We use the formula
\begin{equation}
\int_0^{\infty} \erfc(a x) e^{b^2 x^2} d x=\frac{1}{2 \sqrt{\pi} b} \ln \left[\frac{a+b}{a-b}\right], \quad b \text { may be complex, } \quad|\arg a|<\frac{\pi}{4}
\end{equation}

Finally, we obtain an expression for the adversarial generalisation error as a function of the standard generalisation error and an integral for the boundary error. 
\begin{equation}\label{eq:adversarial-error-vanilla-sum}
    \advgenerr = \frac{1}{\pi} \arccos \qty(\frac{m}{\sqrt{(\rho + \tau^2 ) q}}) + \int_{0}^{\advgencost  \frac{\sqrt{A}}{\sqrt{q}}} \erfc\left(\frac{ -\frac{m}{\sqrt{q}} \xi}{\sqrt{2\left(\rho + \tau^2 - m^2 / q\right)}}\right) \frac{e^{-\frac{\xi^2}{2}}}{\sqrt{2 \pi}} \dd{\xi}
\end{equation}

We would like to simplify this part more and obtain a relation to the Owen T function definition \cite{owenT56}. From \cite{Ng1969ATO, Korotkovintegralserrorfunction} we use the following identity
\begin{equation}
    \int_0^b e^{-x^2} \erfc(a x) d x= \frac{1}{2 \sqrt{\pi} a} \qty[-4 \pi \sqrt{a^2} T\left(\sqrt{2} \sqrt{a^2} b, \frac{1}{\sqrt{a^2}}\right)+\pi a \erf(b) \erfc(a b) + 2 a \cot ^{-1}(a)]
\end{equation}
where we are using Owen's T function which is defined as
\begin{equation}
    T(h, a)=\frac{1}{2 \pi} \int_0^a \frac{e^{-\frac{1}{2} h^2\left(1+x^2\right)}}{1+x^2} \dd{x} \quad(-\infty<h, a<+\infty) .
\end{equation}

By performing a change of variables \(\xi^\prime = \xi / \sqrt{2}\) we can simplify further the second term of \cref{eq:adversarial-error-vanilla-sum} as
\begin{equation}
\begin{aligned}
    \bounderr &= 2 T\qty( 
        \frac{ m }{\sqrt{(\rho + \tau^2 ) q - m^2}} \advgencost \frac{\sqrt{A}}{\sqrt{q}}, \frac{\sqrt{(\rho + \tau^2 ) q - m^2}}{m} 
    ) \\  
    &+ \frac{1}{2} 
    \erf\qty(\advgencost \frac{\sqrt{A}}{\sqrt{2 q}}) 
    \erfc\qty(\frac{-m }{\sqrt{2((\rho + \tau^2 ) q - m^2)}} \advgencost \frac{\sqrt{A}}{\sqrt{q}}) \\
    &+ \frac{1}{\pi} \cot^{-1}\qty(\frac{-m}{\sqrt{(\rho + \tau^2 ) q - m^2}})  
\end{aligned}
\end{equation}

Given that the adversarial test error in the case where it is not attacked is equal to the standard test error, it should be possible to simplify the expression for the owen's T function.
Indeed, the following property \cite{owenT80} holds
\begin{equation}
    T(0, a) = \frac{1}{2 \pi} \arctan (a) = \frac{1}{2 \pi} \arccos\qty(\frac{a}{\sqrt{1 + a^2}})
\end{equation}
and this leads us to the unperturbed version of the generalisation error.

\subsection{Simplified Expressions for Generalisation and Boundary Errors}\label{sec:simplified_gen_boundary_error}
We can rewrite the key quantity describing generalisation error as follows
\begin{equation}
    \frac{m_0}{\sqrt{(\rho + \tau^2) q_0}} = \frac{m_0}{\sqrt{\rho q_0}} \frac{1}{\sqrt{1 + \frac{\tau^2}{\rho}}} \,.
\end{equation}
Whenever \(\vartheta = \frac{m_0}{\sqrt{\rho q_0}} \to 1\) when \(\alpha \to \infty\), we can leverage, that the angle between teacher and student is zero, and the actual vectors differ only up to a constant, that is \(\w = c \wstar\).

\begin{equation}
    \generr \underset{\alpha \rightarrow \infty}{=} \frac{1}{\pi} \arccos( \vartheta \sqrt{\frac{\pi}{2}} \frac{\usefulmetric_{\wstar}}{\sqrt{\rho}}) = \frac{1}{\pi} \arccos( \frac{\sqrt{\rho}}{\sqrt{\rho + \tau^2}} )  
\end{equation}

Furthermore, the quantity \(\frac{\sqrt{A}}{\sqrt{q}}\) converges to \(\frac{a}{\sqrt{\rho}}\) by definition.
Writing the boundary error as a function of \(\usefulmetric_{\wstar} - \robustmetric_{\wstar} \), leads thus
\begin{equation}
    \bounderr = \int_{0}^{\advgencost \frac{1}{\sqrt{q}} (\usefulmetric_{\wstar} - \robustmetric_{\wstar})} f(\xi; \vartheta, \frac{\usefulmetric_{\wstar}}{\sqrt{\rho}}) \dd{\xi} 
    \underset{\alpha \rightarrow \infty}{=} \int_{0}^{\advgencost \frac{a}{\sqrt{\rho n}}} f(\xi; \frac{1}{\sqrt{1 + \frac{\tau^2}{\rho}}}) \dd{\xi} \,.
\end{equation}
We can simplify this expression further under the assumption of a uniform attack in a BFM with a singular block. In this case, the boundary term can be simplified and we obtain
\begin{equation}
    \bounderr \underset{\alpha \rightarrow \infty}{=}\int_{0}^{\advgencost \frac{1}{\sqrt{\psi_\ell}}} f(\xi; \frac{\sqrt{\rho}}{\sqrt{\rho + \tau^2}} ) \dd{\xi} \,.
\end{equation}
In this simple setting, the variance of the data influences inverse proportionally the integration bound.

\section{ADVERSARIAL PROBLEM AS AN APPROXIMATE DATA-DEPENDANT REGULARISATION}
\label{sec:equivalent-problem}

\begin{figure}[t]
    \centering
    \includegraphics[width=0.475\textwidth]{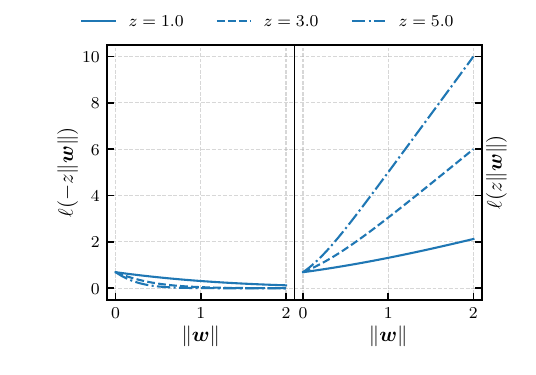}
    \caption{  Behaviour of the loss function for positive and negative margins as a function of the norm of the student estimate. A negative margin favours a zero norm solution.}
    \label{fig:loss_expansion}
\end{figure}

In other settings, adversarial training has been studied as a form of data dependent regularisation \cite{roth_adversarial_2020,ribeiro2023regularization}. In this Appendix, we provide an approximately equivalent loss function that can also be understood as a data dependent regularisation.
Here by data dependant regularisation we mean an explicit term in the minimisation function that is separate from the loss function term, even if it can still depend on the dataset $\mathcal{D}$.

Recall the minimisation problem in \cref{eq:modified-minimisation-problem} is
\begin{equation}
    \sum_{\mu = 1}^{n} 
    g \qty(
        y^\mu \frac{\w^\top \x^\mu}{\sqrt{d}} 
        - \advtrainingcost \frac{\sqrt{\w^\top \Sigmadelta \w}}{\sqrt{d} } 
    ) 
    + \frac{\lambda}{2} \norm{\w}_2^2
\end{equation}
where we specified the \(\ell_2\) regularisation. Also in this Appendix we will use \(g\) to specifically indicate the logistic loss, the same analysis could be performed for other losses similarly. 

We want to split the sum into a first part, where the shifted margin is positive and a second part, where the shifted margin is negative 
\begin{equation}
    \sum_{\mu = 1}^{n_1} 
    g \qty(
        y^\mu \frac{\w^\top \x^\mu}{\sqrt{d}} 
        - \advtrainingcost \frac{\sqrt{\w^\top \Sigmadelta \w}}{\sqrt{d} }  
    ) 
    + \frac{\lambda}{2} \norm{\w}_2^2
    + \sum_{\mu = n_1 + 1}^{n} 
    g \qty(
        y^\mu \frac{\w^\top \x^\mu}{\sqrt{d}} 
        - \advtrainingcost \frac{\sqrt{\w^\top \Sigmadelta \w}}{\sqrt{d} } 
    ) \,.
\end{equation}
Here, \(n_1\) is the number of points that have shifted positive margin. We call a margin positive, when \(y^\mu \frac{\w^\top \x^\mu}{\sqrt{d}} - \advtrainingcost \frac{\sqrt{\w^\top \Sigmadelta \w}}{\sqrt{d} }   > 0\). We expand the term with the negative margin around zero for the logistic loss to obtain
\begin{equation}\label{eq:perturbed_loss}
    \sum_{\mu = 1}^{n_1} 
    g \qty(
        \dots
    ) 
    + \frac{\lambda}{2} \norm{\w}_2^2
    - \frac{1}{2} \sum_{\mu = n_1 + 1}^{n} 
    \qty[
        y^\mu \frac{\w^\top \x^\mu}{\sqrt{d}} 
        - \advtrainingcost \frac{\sqrt{\w^\top \Sigmadelta \w}}{\sqrt{d} } 
    ]
    + \frac{1}{8} \sum_{\mu = n_1 + 1}^{n} 
    \qty[
        y^\mu \frac{\w^\top \x^\mu}{\sqrt{d}} 
        - \advtrainingcost \frac{\sqrt{\w^\top \Sigmadelta \w}}{\sqrt{d} }  
    ]^2
\end{equation}
The expansion around zero is justified, as a negative margin favours a zero norm solution, which we show in \cref{fig:loss_expansion}.

A first approximation is to not keep the mixed terms and resum all of them to obtain
\begin{equation}
    \sum_{\mu = 1}^{n_1} 
    g \qty(\dots) 
    + \sum_{\mu = n_1 + 1}^{n}
    g \qty( 
        y^\mu \frac{\w^\top \x^\mu}{\sqrt{d}} 
    ) 
    + \frac{\lambda}{2} \norm{\w}_2^2
    + (n - n_1) g \qty( 
        \frac{\advtrainingcost }{\sqrt{d}} \sqrt{\w^\top \Sigmadelta \w}
    ) 
\end{equation}
Note that one can also consider the case where \(\advtrainingcost\) is small and thus keep just the first terms in \(\advtrainingcost\) to get something like the following
\begin{equation}
\begin{aligned}
    \sum_{\mu = 1}^{n_1} 
    g \qty(\dots) 
    &+ \sum_{\mu = n_1 + 1}^{n}
    g \qty( 
        y^\mu \frac{\w^\top \x^\mu}{\sqrt{d}} 
    ) 
    + \frac{\lambda}{2} \norm{\w}_2^2
    + (n - n_1) \frac{\advtrainingcost}{2}
    \frac{\sqrt{\w^\top \Sigmadelta \w}}{\sqrt{d}} \\
    &- \frac{\advtrainingcost}{4} \frac{\sqrt{\w^\top \Sigmadelta \w}}{\sqrt{d}}  \sum_{\mu = n_1 + 1}^{n} y^\mu \frac{\w^\top \x^\mu}{\sqrt{d}} \\
    &+\frac{\advtrainingcost^2}{8} \frac{\w^\top \Sigmadelta \w}{d} (n - n_1)\,.
\end{aligned}
\end{equation}
In both cases in the limit \(\advtrainingcost\to 0\) there are two types of regularisation, the first is an \(\sqrt{\ell_2}\) regularisation and the other is a negative shifted margin dependent \(\ell_2\) regularisation.

In our setting we suppose that the coefficient for the second term is positive as we are considering Gaussian data and in high dimension most of the data lies very close to the boundary.

The loss can then be written as
\begin{equation}\label{eq:equivalent-adversarial-regularisation-appendix}
    \sum_{\datidx = 1}^{n} 
    g \qty( y_\datidx \frac{\w^\top \x_\datidx}{\sqrt{d}} ) 
    + \tilde{\lambda}_1 \sqrt{\w^\top \Sigmadelta \w} + \tilde{\lambda}_2 \w^\top \Sigmadelta \w \,.
\end{equation}
where the value for \(\tilde{\lambda}_1\) and \(\tilde{\lambda}_2\) can be read from the previous expansion.

\section{DEFENDABLE ATTACKS AND INEVITABLE TRADE-OFFS}\label{sec:best-choice-attack-defence}

In this Appendix we want to study the relation between the attack geometry \(\Sigmaupsilon\) and the defence geometry \(\Sigmadelta\). We assume to know the attack geometry. If we can successfully defend against a geometry, without incurring a trade-off, we call the geometry \textit{defendable}, otherwise \textit{not defendable}. A trade-off does not occur, if the adversarial error is dominated by the generalisation error and if the generalisation error behaves roughly as a constant as a function of the defence strength \(\advtrainingcost\). 

To make a fair comparison between the geometries we are going to consider a normalised version of the matrices \(\Sigmaupsilon\) and \(\Sigmadelta\) and consider the total cost of the attack/defence to be tuned by \(\advgencost\)/\(\advtrainingcost\). 
We are asking how to spend a fixed budget among the defence directions, to protect most effectively against a given attack. Since both matrices are symmetric we can write them as \(\boldsymbol{\Sigma} = \sum_{i=1}^{d} \lambda_i \boldsymbol{v}_i \boldsymbol{v}_i^\top \) where \(\sum_{i=1}^{d} \lambda_i = 1\) and the eigenvectors are normalised \(\norm{\boldsymbol{v}_i}_2 = 1\).

To keep things simple, we can suppose that the attack matrix is \(\Sigmaupsilon = \boldsymbol{v} \boldsymbol{v}^\top\). We will refer to \(\boldsymbol{v}\) as the unit vector that is composed of a sum between \(\wstar\) and another vector \(\boldsymbol{u}\) perpendicular to \(\wstar\). 

We can thus decompose the attack matrix as
\begin{equation}
    \Sigmaupsilon = (\wstar + \boldsymbol{u}) (\wstar + \boldsymbol{u})^\top = \wstar \wstar^\top + \wstar \boldsymbol{u}^\top + \boldsymbol{u} \wstar^\top + \boldsymbol{u} \boldsymbol{u}^\top
\end{equation}
To reduce the attack strength, we want to minimise the the attack overlap \(A\) which is \(\w^\top \Sigmaupsilon \w\). 

To minimise the term \(\what^\top \boldsymbol{u} \boldsymbol{u}^\top \what\), we would like to make the ERM procedure chose a vector \(\w\) that is perpendicular to \(\boldsymbol{u}\). This can be done without affecting generalisation performances as \(\boldsymbol{u} \perp \wstar\). A way to remove this, is to regularise in the direction \(\boldsymbol{u}\). This can be done with a directional \(\ell_2\) regularisation or with adversarial training and choosing \(\Sigmadelta = \boldsymbol{u} \boldsymbol{u}^\top\). With this choice of \(\Sigmadelta\), the values \(\what^\top \wstar \boldsymbol{u}^\top \what\) and \(\what^\top \boldsymbol{u} \wstar^\top \what\) can be decreased in the same way.

For the last term \(\what^\top \wstar \wstar^\top \what\), this recipe does not work. We can interpret this term as being proportional to the square of the overlap \(m\) for isotropic data. Thus, it is obvious that protecting such a direction, can only be achieved by reducing the norm of \(\what\), which reduces the overlap \(m\), which in turn cannot be changed without hurting the angle between teacher and student and thus generalisation error. This term is thus at the heart of the trade-off between generalisation and boundary term.

Thus shown that all the components of the eigenvectors of \(\Sigmaupsilon\) that are perpendicular to the teacher can be ``regularised away''.

\section{ANALYSIS OF THE FAST GRADIENT METHOD ADVERSARIAL TRAINING}
\label{sec:fast-gradient-method}

Sometimes in practical adversarial one does not solve exactly the each-sample maximisation as it is to much computationally intensive for general machine learning models. Thus people turn to Fast Gradient Method (FGM) that corresponds to taking a single gradient step in the direction of max gradient with the biggest possible norm.
Unlike explicitly maximising the loss with respect to the data, which requires iterative and computationally intensive optimisation, FGM provides a more efficient, single-step approach. This efficiency makes FGM practical for training robust models on large datasets.

In the case of linear models we have that the objective is
\begin{equation}
    \sum_{\datidx = 1}^{n} g\qty( 
        y_\datidx \frac{\w^\top \x_\datidx}{\sqrt{d}} - \advtrainingcost \frac{\w^\top \Sigmadelta \w}{\sqrt{d} \norm{\w}_2}
    )
\end{equation}

Our theoretical toolkit is versatile enough to be adapted to this problem as well. We provide in \cref{sec:replica-computation} a statistical physics derivation of the two results in parallel. We also believe that also this results holds formally as a theorem.

\subsection{Class-preserving adversarial attacks}\label{sec:class-preserving-gen-attacks}

The ability of explicitly defining a direction allows us to reason about attacks that do not cross the teacher margin that we derive in this paragraph.

We propose a more refined metric called the class-preserving generalisation error. This metric only considers attacks that mislead the student model while not affecting the teacher model's classification or its confidence in that classification. The teacher's confidence is quantified by a margin, denoted as \(\gamma\), which represents the minimum allowable distance from the decision boundary to consider a classification confident.

This is an artificial metric, as it would be inaccessible to a student. We are interested in measuring the error with respect to the noisy labels by only considering attacks which are fair with respect to the ground truth.

We focus on attacks that attempt to deceive the student model within a specified norm bound but also ensure that the teacher model does not misclassify the perturbed data. The class-preserving generalisation error, incorporating these considerations, is defined as:
\begin{equation}\label{eq:def-class-preseriving-marign-error}
\begin{aligned}
    \classpresgenerr(\advgencost, \gamma) =& \EEb{y,\x}{
        \max_{
            \substack{
                \norm{\vecdelta}_{\Sigmaupsilon^{-2}} \leq \advgencost \\
                y \boldsymbol{\theta}_0^\top (\x + \vecdelta) > \sqrt{d} \gamma
            }
        } \mathbb{1}(y \neq \hat{y} (\what, \x + \vecdelta) )
        \mathbb{1}(y \wstar^\top \x > \sqrt{d} \gamma)
    } \\ 
    &+ \EEb{y,\x}{
        \mathbb{1}(y \neq \hat{y} (\what, \x ) )
        \mathbb{1}(y \wstar^\top \x \leq \sqrt{d} \gamma)
    }
\end{aligned}
\end{equation}

The constraint within the maximisation ensure that the perturbation does not cause the teacher model's confidence classification is not diminished beyond the margin \(\gamma\) and that the total length of the perturbation doesn't surpass the value \(\advgencost\). The two constraints can be rewritten as
\begin{align}
    & \vecdelta^\top \Sigmaupsilon^{-2} \vecdelta \leq \advgencost^2 \,, \label{eq:fair-gen-error-norm-cond} \\
    & y \wstar^\top \vecdelta > \gamma - y \wstar^\top \x \,, \label{eq:fair-gen-error-margin-perturbed-cond}
\end{align}
this means that the projection of the perturbation on the teacher should always be greater in absolute value than the projection of the data, \textit{i.e.} the data point classified from the teacher should not change.

To simplify further one can consider the vanishing margin case as in the main text, formally
\begin{equation}
    \classpresgenerr(\advgencost) = \lim_{\gamma \to 0^+} \classpresgenerr(\advgencost, \gamma) \,.
\end{equation}

We want to fool the student, so we consider a perturbation in the direction of \(\what\). Based on different conditions we have different attacks and thus different norm of this perturbation. We will indicate the norm as
\begin{equation}
    \vecdelta_{(\cdot)} = \alpha_{(\cdot)} \frac{\Sigmaupsilon \what}{\norm{\what}_2}
\end{equation}
where the subscript can indicate any of the cases and we will define them later. 

The first case is if the point \(\x\) already has a margin with the teacher that it is smaller than \(\gamma\). The condition for this to happen is that
\begin{equation}
    y \wstar^\top \x < \gamma \quad \implies \quad \alpha_{\mathrm{LM}} =0 \,,\quad 
    \vecdelta_{\mathrm{LM}} = \boldsymbol{0} \,, 
\end{equation}
in this case we do not perturb the input data point.

By just considering \cref{eq:fair-gen-error-norm-cond} we have that the maximum norm the perturbation can have is
\begin{equation}
    \alpha_{\mathrm{MAX}} = - y \advgencost \,,\quad \vecdelta_{\mathrm{MAX}} = - y \advgencost \frac{\Sigmaupsilon \what}{\norm{\what}_2}
\end{equation}
this is the strongest attack, but we can only use it if the perturbed image does not cross the margin \(\gamma\) of the teacher \(\wstar\). In equations, it means that \cref{eq:fair-gen-error-margin-perturbed-cond} is satisfied, which means that
\begin{equation}\label{eq:condition-pertubed-good-margin}
    y \frac{\wstar^\top (\x + \vecdelta_{\mathrm{MAX}})}{\sqrt{d}} \geq \gamma \, \iff \, y \frac{\wstar^\top \x}{\sqrt{d}} - \advgencost \frac{\wstar^\top \Sigmaupsilon \what}{\sqrt{d} \norm{\what}_2} \geq \gamma
\end{equation}

thus if this previous condition is satisfied we can perturb with \(\vecdelta_{\mathrm{MAX}}\).

If instead \cref{eq:fair-gen-error-margin-perturbed-cond,eq:condition-pertubed-good-margin} are not satisfied we can proceed the attack but with a smaller norm, that is found by imposing that the final margin is equal to \(\gamma\). By doing that we can solve for the final norm and obtain
\begin{equation}
    \alpha_{\gamma} = \frac{\sqrt{d} \norm{\what}_2}{\wstar^\top \Sigmaupsilon \what} \qty(y \gamma -  \frac{\wstar^\top \x}{\sqrt{d}}) \,, \quad 
    \vecdelta_{\gamma} = \frac{\sqrt{d} \Sigmaupsilon \what }{\wstar^\top \Sigmaupsilon \what} \qty(y \gamma -  \frac{\wstar^\top \x}{\sqrt{d}})
\end{equation}
where remember that \(y\) is just a sign.

One can thus rewrite explicitly the maximisation in \cref{eq:def-class-preseriving-marign-error} as follows
\begin{equation}\label{eq:class-preserving-error-erm}
\begin{aligned}
    \classpresgenerr(\advgencost, \gamma) &= \EEb{y,\x}{
        \mathbb{1}\qty(y \neq \hat{y} (\what, \x + \vecdelta_{\gamma}))
        \mathbb{1}\qty(y \frac{\wstar^\top \x}{\sqrt{d}} - \advgencost \frac{\wstar^\top \Sigmaupsilon \what}{\sqrt{d} \norm{\what}_2} < \gamma)
        \mathbb{1}(y \wstar^\top \x > \sqrt{d} \gamma)
    } \\
    &+ \EEb{y,\x}{
        \mathbb{1}\qty(y \neq \hat{y} (\what, \x + \vecdelta_{\mathrm{MAX}}))
        \mathbb{1}\qty(y \frac{\wstar^\top \x}{\sqrt{d}} - \advgencost \frac{\wstar^\top \Sigmaupsilon \what}{\sqrt{d} \norm{\what}_2} \geq \gamma)
        \mathbb{1}(y \wstar^\top \x > \sqrt{d} \gamma)
    } \\
    &+ \EEb{y,\x}{
        \mathbb{1}(y \neq \hat{y} (\what, \x ) )
        \mathbb{1}(y \wstar^\top \x \leq \sqrt{d} \gamma)
    }
\end{aligned}
\end{equation}
where each one of the lines correspond to one of the previously explained cases. To compute the expectations in \cref{eq:class-preserving-error-erm}, because of the presence of \(y\) one should consider the actual channel that generates the data. Nevertheless the form in \cref{eq:class-preserving-error-erm} is well suited for estimation from a given test dataset by replacing the expectations with empirical averages over that training dataset.

\subsubsection{Noiseless Channel}\label{sec:class-preserving-noiseless}

In the case that the channel is a noiseless sign channel we have 
\begin{equation}\label{eq:fair-err-overlaps-noiseless}
\begin{aligned}
    \classpresgenerr(\advgencost, \gamma) &= \int \dd{\mu(\nu,\lambda)}
    \mathbb{1}\qty[\sign (\nu) \neq \sign\qty(\lambda + \sign(\nu) \frac{A}{F} \qty(\gamma - \abs{\nu} ))] 
    \mathbb{1}\qty[\abs{\nu} - \advgencost \frac{F}{\sqrt{N}} < \gamma] 
    \mathbb{1}\qty[\abs{\nu} > \gamma] \\
    & + \int \dd{\mu(\nu,\lambda)}
    \mathbb{1}\qty[\sign (\nu) \neq \sign\qty(\lambda - \sign(\nu) \advgencost \frac{A}{\sqrt{N}})] 
    \mathbb{1}\qty[\abs{\nu} - \advgencost \frac{F}{\sqrt{N}} \geq \gamma] 
    \mathbb{1}\qty[\abs{\nu} > \gamma] \\
    & + \int \dd{\mu(\nu,\lambda)}
    \mathbb{1}\qty[\sign(\nu) \neq \sign(\lambda)] 
    \mathbb{1}\qty[\abs{\nu} \leq \gamma]
\end{aligned}
\end{equation}
where we have that the local fields \(\nu, \lambda\) are jointly Gaussian with zero mean and covariance \(\bigl( \begin{smallmatrix}\rho & m \\ m & q\end{smallmatrix}\bigr)\), this is the probability distribution \(\dd{\mu(\nu,\lambda)}\).

To simplify the double integral in \cref{eq:fair-err-overlaps-noiseless} it is a matter of integrating over \(\lambda\) and then computing the integration over \(\nu\) numerically, as it is in a single variable. In the end, we obtain
\begin{equation}
\begin{aligned}
    \classpresgenerr(\advgencost, \gamma) &= \int_0^\gamma \erfc\qty(\frac{m \nu }{\sqrt{2 \rho  \left(q \rho - m^2\right)}}) \mathcal{D}_\rho\qty[\nu] 
    + \int_{\gamma^\star}^{\infty} \erfc\qty(\frac{m \nu - \advgencost \sqrt{N} \rho }{\sqrt{2 \rho  \left(q \rho -m^2\right)}}) \mathcal{D}_\rho\qty[\nu] \\ 
    &+ \frac{1}{2} \qty[
        \int_{\gamma}^{\gamma^\star} \erfc\qty( \frac{A \rho \gamma + \nu (m F  - A \rho)}{ F \sqrt{2 \rho  \left(q \rho -m^2\right)}} ) \mathcal{D}_\rho\qty[\nu] +
        \int_{-\gamma^\star}^{-\gamma} \erfc\qty( \frac{A \rho \gamma - \nu (m F  - A \rho)}{ F \sqrt{2 \rho  \left(q \rho -m^2\right)}} ) \mathcal{D}_\rho\qty[\nu] 
    ] 
\end{aligned}
\end{equation}
where the notation \(\mathcal{D}_\rho\qty[\nu]\) indicates the p.d.f. of a random variable \(\nu \sim \mathcal{N}(0,\rho^2)\) and \(\gamma^\star = \max(\gamma, \gamma + \advgencost F / \sqrt{N}) \).

\subsubsection{Probit Channel}\label{sec:class-preserving-probit}
In the case where the channel is a probit channel we have that the class preserving error could be computed similarly at the addition of one integral. 
The integration is done over the measure \(\dd{\mu}(\nu, \lambda)\) as before but one should add the probability distribution of the channel. Thus we have that the integration becomes 
\begin{equation}\label{eq:fair-err-overlaps-noisy}
\begin{aligned}
    & \classpresgenerr(\advgencost, \gamma) \\ 
    &= \int \dd{y} \dd{\mu(\nu,\lambda)} P(y \mid \nu)
    \mathbb{1}\qty[y \neq \sign\qty(\lambda + \sign(\nu + z) \frac{A}{F} \qty(\gamma - \abs{\nu} ))] 
    \mathbb{1}\qty[y \nu - \advgencost \frac{F}{\sqrt{N}} < \gamma] 
    \mathbb{1}\qty[y \nu > \gamma] \\
    & + \int \dd{y} \dd{\mu(\nu,\lambda)} P(y \mid \nu)
    \mathbb{1}\qty[y \neq \sign\qty(\lambda - \sign(\nu) \advgencost \frac{A}{\sqrt{N}})] 
    \mathbb{1}\qty[y \nu - \advgencost \frac{F}{\sqrt{N}} \geq \gamma] 
    \mathbb{1}\qty[y \nu > \gamma] \\
    & + \int \dd{y} \dd{\mu(\nu,\lambda)} P(y \mid \nu)
    \mathbb{1}\qty[y\neq \sign(\lambda)] 
    \mathbb{1}\qty[y \nu \leq \gamma]
\end{aligned}
\end{equation}
where we have introduced the channel probability distribution.

\subsection{Comparison between full minimisation and FGM}

\begin{figure}
    \centering
    \includegraphics[width=0.88\textwidth]{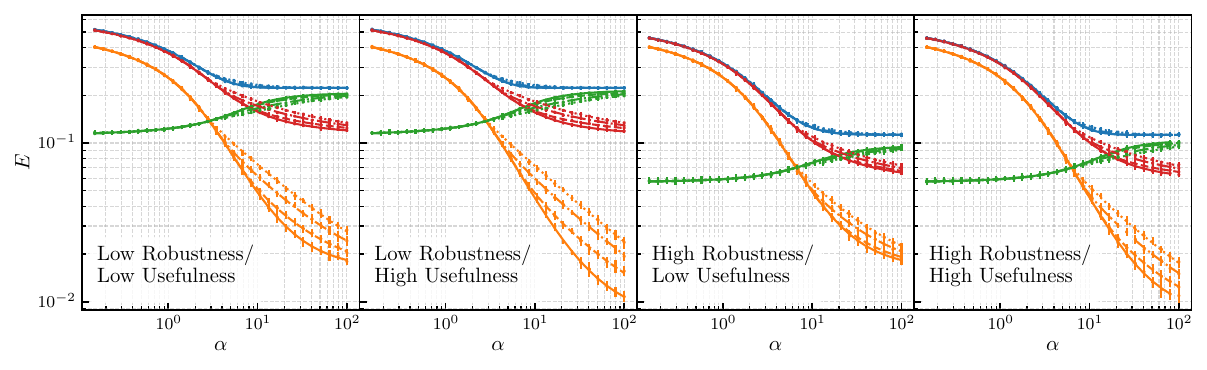}
    \hspace{-0.75em}
    \raisebox{0.3\height}{
        \includegraphics[width=0.09\textwidth]{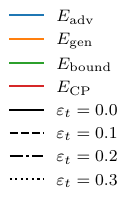}
    }
    \caption{FGM training for single block SWFM with the same setting as in \cref{fig:feature_comparison}. We see that the qualitative behaviour is very similar among the two figures.}
    \label{fig:FGM-feature-combination}
\end{figure}

\begin{figure}
    \centering
    \includegraphics[width=0.65\textwidth]{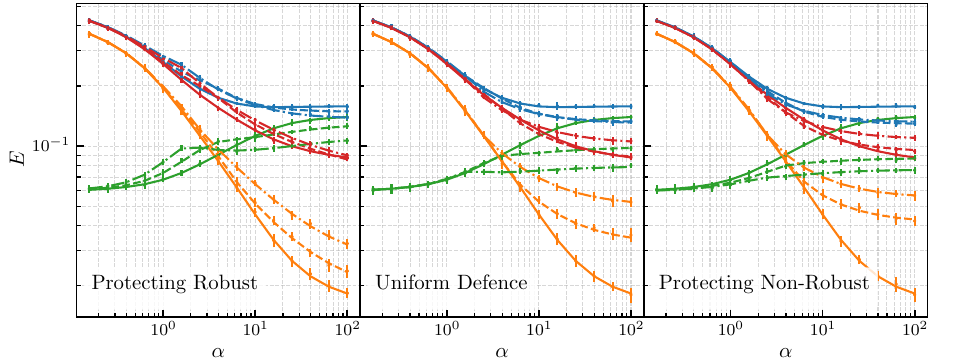}
    \caption{FGM training with the same setting as in \cref{fig:defence_sweep} (Left). We see that the qualitative behaviour is very similar among the two figures.}
    \label{fig:FGM-defence-sweep}
\end{figure}

From the look of \cref{fig:FGM-defence-sweep,fig:FGM-feature-combination}, taht correspond to the same setting as \cref{fig:defence_sweep,fig:feature_comparison}, we see that the qualitative behaviours are very similar among the two cases. 

\begin{itemize}
    \item In both cases we have that the adversarial error \(\advgenerr\) is dominated by generalisation error in the low sample complexity regime and it is dominated by boundary error in the high sample complexity regime. 
    \item In both cases we see that usefulness determines the plateau for generalisation error and robustness the one for boundary error.
    \item Among the different protection strategies protecting the non robust features leads to a smaller adversarial generalisation error in the high sample complexity regime.
\end{itemize}

A small difference that one can notice is that the difference in performances between the three cases considered in \cref{fig:FGM-defence-sweep} is more marked than the one considered in \cref{fig:defence_sweep}. 

In both plots we the red curves correspond to the class-preserving error explained in \cref{sec:class-preserving-gen-attacks}. We can see that this kind of error metric behaves as the adversarial training error: at high sample complexities it has the same value as the generalistion error while increasing the number of training points decreases its value.
One key difference is that the lowest value reached by \(\classpresgenerr\) is lower than the one reached by \(\advgenerr\). 
This behaviour can be explained as follows. As the sample complecity increases the student vector \(\what\) aligns with the teacher vector \(\wstar\) and in the case of \(\advgenerr\) a lot of points start crossing the boundary leading to an increase of the boundary error \(\bounderr\). In the case of the class preserving error these points will be perturbed to not misclassified as the boundary for the teacher and the students are almost the same and the adversarial perturbation cannot flip the teacher labels.
For this reason the class-preserving error is an intermediate error metric between generalisation and adversarial error.

\section{EXPERIMENTS ON REAL DATA}
\label{sec:real_data_experiments}

In this Appendix we demonstrate the robustness metric and defense-strategies evaluated the Cifar 10 \cite{krizhevsky_learning_nodate} and FashionMNIST \cite{xiao2017fashion}  datasets.

\begin{table}[H]
    \caption{We isolate robust features and demonstrate that attacking the non-robust features contributes most to boundary error. For CIFAR10, we keep the first 2500 principle components, and evaluate the target 1 vs 9. For FashionMNIST, we keep the first 500 principle components and evaluate odd versus even targets.}
    \label{tab:isolating_robust_features}
    \vskip 0.15in
    \begin{center}
        \begin{small}
            \begin{sc}
                \begin{tabular}{lcccr}
                    \toprule
                    \multicolumn{4}{c}{\textbf{Cifar 10}} \\
                    \midrule
                    Attack Direction & \(\trainerr\) & \(\generr\) & \(\advgenerr\) \\
                    \midrule
                    Uniform & 0.0749 & 0.2225 & 0.2645 \\
                    Robust & 0.0749 & 0.2225 & 0.2230 \\
                    Non-Robust & 0.0749 & 0.2225 & 0.2645 \\
                    \bottomrule
                \end{tabular}
                \vskip 0.1in
                \begin{tabular}{lcccr}
                    \toprule
                    \multicolumn{4}{c}{\textbf{Fashion MNIST}} \\
                    \midrule
                    Attack Direction & \(\trainerr\) & \(\generr\) & \(\advgenerr\) \\
                    \midrule
                    Uniform & 0.0326 & 0.0396 & 0.0635 \\
                    Robust & 0.0326 & 0.0396 & 0.0427 \\
                    Non-Robust & 0.0326 & 0.0396 & 0.0595 \\
                    \bottomrule
                \end{tabular}
            \end{sc}
        \end{small}
    \end{center}
    \vskip -0.1in
\end{table}

\begin{table}[H]
    \caption{We protect the non-robust features and find that identity and inverse proportionality to the explained variance is good compared to protecting proportional to the explained variance.}
    \label{tab:protecting_non_robust_features}
    \vskip 0.15in
    \begin{center}
        \begin{small}
            \begin{sc}
                \begin{tabular}{lcccr}
                    \toprule
                    \multicolumn{4}{c}{\textbf{Cifar 10}} \\
                    \midrule
                    Protection Direction & \(\trainerr\) & \(\generr\) & \(\advgenerr\) \\
                    \midrule
                    Uniform & 0.0802 & 0.1950 & 0.2255 \\
                    Proportional & 0.0736 & 0.2210 & 0.2625 \\
                    Inverse & 0.0873 & 0.1965 & 0.2230 \\
                    \bottomrule
                \end{tabular}
                \vskip 0.1in
                \begin{tabular}{lcccr}
                    \toprule
                    \multicolumn{4}{c}{\textbf{Fashion MNIST}} \\
                    \midrule
                    Protection Direction & \(\trainerr\) & \(\generr\) & \(\advgenerr\) \\
                    \midrule
                    Uniform & 0.0352 & 0.0395 & 0.0493 \\
                    Proportional & 0.0337 & 0.0391 & 0.0558 \\
                    Inverse & 0.0354 & 0.0392 & 0.0503 \\
                    \bottomrule
                \end{tabular}
            \end{sc}
        \end{small}
    \end{center}
    \vskip -0.1in
\end{table}

\subsection{Finding Non-Robust Features}

The margin-based robustness given in \cref{eq:mt-robustness-def} can be interpreted in terms of the student estimate \(\w\) and it can be written as
\begin{equation}\label{eq:empirical_robustness}
    \robustmetric^{\mathrm{emp.}}_j = \sum_{\datidx=1}^n y_\datidx \w_j (\x_\datidx)_j - \advgencost \sqrt{ \w^\top \Sigmadelta \w}
\end{equation}
which can be evaluated component-wise.

Our aim is to apply this measure on components obtained by PCA. The motivation for choosing PCA as a method of feature extraction and dimensionality reduction is twofold, on the one hand it creates uncorrelated variables and on the other hand, it creates them in order of explained variance. This produces a data-set similar to the BFM with a power-law distributed spectrum.

We begin our procedure by computing the principal components and projecting all test and training data. Next, we optimise the weights with \(\advtrainingcost = 0\). Next, we increase \(\advtrainingcost\) and evaluate the component-wise robustness-measure \cref{eq:empirical_robustness}. 

By hand, we choose a cutoff-value (\(2000\) for Cifar10 and for \(20000\) FashionMNIST) for the component-wise robustness-measure to produce a mask for robust features. All values higher than the cutoff are considered to be robust. We choose the constant such that the boundary error is explained to a large degree through the non-robust features. This is evaluated by constraining the attack to all features, the robust features and the non-robust features. The results of this experiment is shown in \cref{tab:isolating_robust_features}.

\subsection{Training the Non-Robust Features Adversarially}

A principle finding of ours is the discussion of different defence strategies. Here, we provide some simple experiments on real data. 

In the main text, we suggested to defend the non-robust features either proportional to their variance, inverse proportional to their variance or uniformly. We continue the experiment from the previous paragraph and consider the PCA explained variance as the relevant metric for the proportionality. To be explicit, we perform three different adversarial training runs with different directional choices for \(\Sigmadelta\). We define the uniform defence by having \(\Sigmadelta\) equal to the identity, the proportional defence by setting the directions proportional to their explained variance and the inverse defence by setting the directions proportional to the inverse of the explained variance. The results of this experiment are given in \cref{tab:protecting_non_robust_features}. Indeed, protecting inversely or uniformly appears to work well.

\section{SETTINGS OF MAIN TEXT FIGURES}\label{sec:figures-setting-general}

In this Appendix we provide the parameters for the figures presented in the main text.
For these experimental parts of the paper, we fix the function \(g(x) = \log( 1 + \exp(-x))\) to be the logistic loss.

\subsection{Setting Figure \ref{fig:feature_comparison}}\label{sec:settings-figure-2}

In \cref{fig:feature_comparison} we consider we consider a SWFM with a single block. Considering the figures from left to right, the entries for the parameter specifying the block are 

\begin{table}[H]
    \caption{Parameters for \cref{fig:feature_comparison}}
    \label{tab:parameters-figure-feature-comparison}
    \vskip 0.15in
    \begin{center}
        \begin{small}
            \begin{sc}
                \begin{tabular}{lccccccccr}
                    \toprule
                    Plot Label & \(\psi_\blockidx\) & \(t_\blockidx\) & \(\Delta_\blockidx\) & \(\Upsilon_\blockidx\) & \(\advgencost\) & \(\lambda\) & \(\tau\)  \\
                    \midrule
                    Low Robustness/Low Usefulness & 0.5 & 2 & 1 & 1 & 0.2 & \(10^{-3}\) & 0.05  \\
                    Low Robustness/High Usefulness & 0.5 & 8 & 1  & 1 & 0.2 & \(10^{-3}\) & 0.05 \\
                    High Robustness/Low Usefulness & 2 & 0.5 & 1  & 1 & 0.2 & \(10^{-3}\) & 0.05 \\
                    High Robustness/High Usefulness & 2 & 2 & 1  & 1 & 0.2 & \(10^{-3}\) & 0.05 \\
                    \bottomrule
                \end{tabular}
            \end{sc}
        \end{small}
    \end{center}
    \vskip -0.1in
\end{table}
The points describe the simulation using ERM routine, where we have used a dimension \(d=1000\) and averaged over 20 runs. We show the mean and the standard deviation on the mean.

\subsection{Setting Figure \ref{fig:defence_sweep}}\label{sec:settings-figure-3}

In \cref{fig:defence_sweep} there are three blocks of images, we start from the first block on the left. Here we show an SWFM consisting of two blocks \(k=2\). The values are filled as follows

\begin{table}[H]
    \caption{Parameters for \cref{fig:defence_sweep} (Left)}
    \label{tab:parameters-figure-defence-sweep}
    \vskip 0.15in
    \begin{center}
        \begin{small}
            \begin{sc}
                \begin{tabular}{lcccccccr}
                    \toprule
                    Plot Label & \((\psi_1,\psi_2)\) & \((t_1,t_2)\) & \((\Delta_1, \Delta_2)\) & \((\Upsilon_1, \Upsilon_2)\) & \(\advgencost\) & \(\lambda\) & \(\tau\)  \\
                    \midrule
                    Protecting Robust & (5,0.2) & (1,1) & (2,1) & (1,1) & 0.2 & \(10^{-3}\) & 0.05  \\
                    Uniform Defence & (5,0.2) & (1,1) & (1,1)  & (1,1) & 0.2 & \(10^{-3}\) & 0.05 \\
                     Protecting Non-Robust & (5,0.2) & (1,1) & (1,2)  & (1,1) & 0.2 & \(10^{-3}\) & 0.05 \\
                    \bottomrule
                \end{tabular}
            \end{sc}
        \end{small}
    \end{center}
    \vskip -0.1in
\end{table}
All matrices are normalised by their trace. The points describe the simulation using ERM, where we have used a dimension \(d=1000\) and averaged over 20 runs. We show the mean and the standard deviation on the mean.

The figure on the top right, shows a power-law spectrum in \(\psi_\blockidx\), we begin the indexing with 1 and choose a coefficient \(\beta=1.5\). Again, all the matrices are normalised by their trace. The values are constructed as follows, we consider the block dimension \(d_\blockidx = 1\), and use the following values
\begin{table}[H]
    \caption{Parameters for \cref{fig:defence_sweep} (Top, Right) and (Bottom, Left)}
    \label{tab:parameters-figure-power-law}
    \vskip 0.15in
    \begin{center}
        \begin{small}
            \begin{sc}
                \begin{tabular}{lcccccccr}
                    \toprule
                    \(\psi_\blockidx\) & \(t_\blockidx\) & \(\Delta_\blockidx\) & \(\Upsilon_\blockidx\) & \(\advgencost\) & \(\lambda\) & \(\tau\)  \\
                    \midrule
                     \(i^{-\beta}\) & 1 & 1 & 1 & 0.2 & \(10^{-3}\) & 0.05  \\
                    \bottomrule
                \end{tabular}
            \end{sc}
        \end{small}
    \end{center}
    \vskip -0.1in
\end{table}

In the figure on the bottom right, we show different power-law behaviour when \(\alpha \to \infty\). Again, we normalise each finite size power-law by their trace, we fix \(d=1000\). The parameters are filled exactly as in the previously described figure, the values can be found in \cref{tab:parameters-figure-power-law}.

\subsection{Setting Figure \ref{fig:regularisation-equivalence}}\label{sec:settings-figure-4}

In \cref{fig:regularisation-equivalence} for both the left and center plot we considered a BFM with a single block \(k=1\). We fix \(\tau = 0.05\) and \(\advgencost=0.2\). 

In both figures, we consider \(t_\blockidx = 1\), \(\psi_\blockidx  = 1\), \(\Delta_\blockidx = 1\), \(\Upsilon_i = 1\) and the values for the \(\advtrainingcost\) are shown in the legend.

In the left figure, we use our state evolution equations to optimise the regularisation strength in terms of generalisation error. Each simulation point hence is created at its respective optimal regularisation.

In the center figure, we compare the learning curves of standard adversarial training with the learning curves of the approximately equivalent loss, for this figure we fix the regularisation strength \(\lambda = 10^{-3}\).

In the right figure we build a SWFM model where we fix two different blocks of features. The percentage for the two features are \((\phi_1,\phi_2) = (0.01,0.99)\). The experiment is performed at a high sample complexity \(\alpha = 10^2\). The other values for the plots are
\begin{table}[H]
    \caption{Parameters for \cref{fig:optimal_defense}}
    \label{tab:parameters-figure-optimal-defence}
    \vskip 0.15in
    \begin{center}
        \begin{small}
            \begin{sc}
                \begin{tabular}{lcccccccr}
                    \toprule
                    Plot Label & \((\psi_1,\psi_2)\) & \((t_1,t_2)\) & \((\Delta_1, \Delta_2)\) & \((\Upsilon_1, \Upsilon_2)\) & \(\advgencost\) & \(\lambda\) & \(\tau\)  \\
                    \midrule
                    Top & (1,1) & (1,\(10^3\)) & (1,\(10^3\)) & (1,\(10^3\)) & 0.006 & \(10^{-3}\) & 0.05  \\
                    Bottom & (1,1) & (\(10^3\),1) & \(\operatorname{diag}(\Sigmadelta^b)\)  & \(\operatorname{diag}(\Sigmaupsilon^b)\) & 0.006 & \(10^{-3}\) & 0.05 \\
                    \bottomrule
                \end{tabular}
            \end{sc}
        \end{small}
    \end{center}
    \vskip -0.1in
\end{table}

In the bottom plot, we compose a vector orthogonal to the diagonal to the teacher covariance \(\singleattack \perp \operatorname{diag}\Sigmatheta \). Then, we choose \(\Sigmadelta^b = \Sigmaupsilon^b = a \singleattack \singleattack^\top + N \). We add a variance 1 Gaussian noise matrix \(N\) to make the matrix non-singular and we increase the importance of the first component through a factor \(a\). All matrices are normalised by their trace.

\section{STATISTICAL PHYSICS DERIVATION OF THE MAIN RESULT}\label{sec:replica-computation}

In this Section we give a full derivation of the results in \cref{thm:main-theorem-saddle-point-eqs} by means of the replica approach, a standard method developed in the realm of statistical physics of disordered systems.
Our computational approach can be found in \cite{Loureiro_2022, Gerace_2021,aubin_generalization_2020}. For a foundational understanding of this effective yet heuristic method, we suggest the following books \cite{mezard1987spin,mezard2009information}.

Between the two computations of the FG method and the true minimisation we chose to present the first one as it is the one more involved and one could obtain the other one by a little simplification when one has to derive the final saddle point equations.

\subsection{Gibbs minimisation}

The starting point is to define the following Gibbs measure over weights \(\w \in \RR^d\). We want the most probable states to be the ones that minimises the ERM problem in the first place. Then to select only these states we will consider the zero temperature limit by taking the parameter \(\beta \to \infty\).
The measure that we are interested in is
\begin{equation}\label{eq:Gibbs_measure}
\mu_\beta(\mathrm{d} \w) =
\frac{1}{\mathcal{Z}_\beta} e^{ -\beta
    \left[
        \sum_{\mu=1}^n g\left(y^\mu, \w^{\top} \x^\mu, \w, \Sigmadelta, \advtrainingcost \right)
        +\frac{\lambda}{2} \w^T \Sigmaw \w 
    \right]}
    \mathrm{d} \w
    = \frac{1}{\mathcal{Z}_\beta} 
        \underbrace{\prod_{\mu=1}^n e^{-\beta g\left(y^\mu, \w^{\top} \x^\mu, \w, \Sigmadelta, \advtrainingcost\right)}}_{P_g} 
        \underbrace{ e^{-\frac{\beta \lambda}{2} \w^T \Sigmaw \w} 
    \mathrm{~d} w_i}_{P_w}    
\end{equation}
Where \(P_g\) is the probability distribution associated with the channel and \(P_w\) is the prior probability distribution.

Here, $\mathcal{Z}_\beta$, is the partition function that normalises the Gibbs measure and it is given by
\begin{equation}\label{eq:Partition_Function}
\mathcal{Z}_\beta= \int_{\RR^d} 
    \mathrm{d} \w
    e^{-\frac{\beta \lambda}{2} \w^T \Sigmaw \w}
    \prod_{\mu=1}^n e^{-\beta g\left(y^\mu, \w^{\top} \x^\mu, \w,\Sigmadelta,  \advtrainingcost\right)}    
\end{equation}

You do need attention, but the free energy density is sufficient. In the zero temperature limit, $\beta \to \infty$ the Gibbs measure in \cref{eq:Gibbs_measure} concentrates around the solutions of the ERM problem. With the replica method, we can compute the free energy density, it is given by:
\begin{equation}\label{eq:free_energy_density}
\beta f_\beta=-\lim _{d \rightarrow \infty} \frac{1}{d} \mathbb{E}_{\mathcal{D}} \log \mathcal{Z}_\beta    
\end{equation}

To evaluate the quenched average of the free energy is to use the replica trick
\begin{equation}\label{eq:replica-trick}
    \lim _{d \rightarrow \infty} \frac{1}{d} \mathbb{E}_{\mathcal{D}} \log \mathcal{Z}_\beta =
    \lim _{r \rightarrow 0} \lim _{d \rightarrow \infty} \frac{1}{d} \frac{\partial_r \mathbb{E}_{\mathcal{D}} \mathcal{Z}^r}{1} 
\end{equation}

Note that we introduced three limits up to here. The first is the zero temperature limit ensuring that we find the ground state of our Gibbs measure which corresponds to the minimum of our ERM problem. The second is the thermodynamic limit of very large dimension whilst keeping the sampling ratio fixed. And the third limit stems from the replica trick allowing us to compute the logarithm of the partition function, it corresponds to setting the number of replicated systems to zero.

This computation follows for the first part the one in \cite{Loureiro_2022}. So we start with the initial definition of replicated partition function the difference we have in our case is that we have a dependence on $\advtrainingcost$ on the output probability.
\begin{equation}
\begin{aligned}
    \mathbb{E}_{\mathcal{D}} \mathcal{Z}_\beta^r & 
    = \prod_{\mu=1}^n \mathbb{E}_{\x^\mu} \prod_{a=1}^r \int_{\RR^d} P_w\left( \dd{\w^a} \right) P\left(y^\mu \mid \frac{\x^\mu \cdot \w^a}{\sqrt{d}}\right) \\ & 
    = \prod_{\mu=1}^n \int_{\RR} \dd{y}^\mu 
    \int_{\RR^p} P_{\boldsymbol{\theta}_0} \left(\dd{\wstar} \right) 
    \int_{\RR^{d \times r}}\left(\prod_{a=1}^r P_w\left(\dd{\w^a}\right)\right) 
    \mathbb{E}_{\x^\mu} \left[
        P_0\left(y^\mu \mid \frac{ \x^\mu \cdot \wstar }{\sqrt{d}}\right) \prod_{a=1}^r P_g\left(y^\mu \mid \frac{\x^\mu \cdot \w^a}{\sqrt{d}}, \Sigmadelta, \w^a, \advtrainingcost \right)
    \right]
\end{aligned}
\end{equation}
explicitly we have that the term in $P_g$ is in the case of the true adversarial attack
\begin{equation}
    P_g\left(y^\mu \mid \frac{\x^\mu \cdot \w^a}{\sqrt{d}}, \Sigmadelta, \w^a, \advtrainingcost \right) = \frac{\sqrt{\beta}}{\sqrt{2\pi}} \exp\qty( 
        - \beta g\qty(
            y \frac{\x^\mu \cdot \w^a}{\sqrt{d}} 
            - \frac{\advtrainingcost}{\sqrt{d}} \sqrt{\w^a \Sigmadelta \w^a}
        ) 
    )
\end{equation}
or in the case of a FGM attack
\begin{equation}
    P_g\left(y^\mu \mid \frac{\x^\mu \cdot \w^a}{\sqrt{d}}, \Sigmadelta, \w^a, \advtrainingcost \right) = \frac{\sqrt{\beta}}{\sqrt{2\pi}} \exp\qty( 
        - \beta g\qty(
            y \frac{\x^\mu \cdot \w^a}{\sqrt{d}} 
            - \frac{\advtrainingcost}{\sqrt{d}} \frac{\w^a \Sigmadelta \w^a}
            {\norm{\w^a}_2}
        ) 
    )
\end{equation}
note that $P_0$ can be a general noisy channel distribution.
the expectation part is equal to:
\begin{equation}
\begin{aligned}
    & \mathbb{E}_{\x^\mu} \left[
        P_0\left(y^\mu \mid \frac{\x^\mu \cdot \boldsymbol{\theta}_0}{\sqrt{d}}\right) \prod_{a=1}^r P_g\left(y^\mu \mid \frac{\x^\mu \cdot \w^a}{\sqrt{d}}, \Sigmadelta, \w^a, \advtrainingcost \right)
    \right] \\ 
    & = \int_{\RR} \mathrm{d} \nu_\mu P_0\left(y \mid \nu_\mu\right) 
    \int_{\RR^r}\left(\prod_{a=1}^r \mathrm{~d} \lambda_\mu^a P_g\left(y^\mu \mid \lambda_\mu^a, \Sigmadelta, \w^a, \advtrainingcost \right)\right) 
    \mathbb{E}_{\x^\mu} 
    \left[
        \delta\left(\nu_\mu - \frac{\x^\mu \cdot \boldsymbol{\theta}_0}{\sqrt{d}}  \right) \prod_{a=1}^r \delta\left(\lambda_\mu^a - \frac{\x^\mu \cdot \w^a}{\sqrt{d}}\right)
    \right]
\end{aligned}
\end{equation}
We can still perform the average over the dataset. We have that the new variables will behave again as Gaussians with the following covariances:
\begin{equation}
    \rho \equiv \mathbb{E}\left[\nu_\mu^2\right] = \frac{1}{d} \boldsymbol{\theta}_0^{\top} \Sigmax \boldsymbol{\theta}_0 , \quad 
    m^a \equiv \mathbb{E}\left[\lambda_\mu^a \nu_\mu\right] = \frac{1}{d} \boldsymbol{\theta}_0^{\top} \Sigmax \w^a, \quad 
    Q^{a b} \equiv \mathbb{E}\left[\lambda_\mu^a \lambda_\mu^b\right] = \frac{1}{d} \w^{a \top} \Sigmax \w^b
\end{equation}
where one can organise them in a single covariance matrix.

Now we want to perform several change of variables. The first one is the one in the matrix of overlaps:
\begin{equation}
\begin{aligned}
    1 \propto & \int_{\RR} \dd{\rho} \delta\left( d \rho-\boldsymbol{\theta}_0^{\top} \Sigmax \boldsymbol{\theta}_0 \right)
    \int_{\RR^r} \prod_{a=1}^r \dd{m^a} \delta\left( d m^a - \boldsymbol{\theta}_0^{\top} \Sigmax \w^a \right) 
    \int_{\RR^{r \times r}} \prod_{1 \leq a \leq b \leq r} \dd{Q^{a b}} \delta\left( d Q^{a b}-\w^{a \top} \Sigmax \w^b \right) \\
    = & \int_{\RR} \frac{\dd{\rho} \mathrm{d} \hat{\rho}}{2 \pi} e^{-i \hat{\rho}\left(d \rho-\boldsymbol{\theta}_0^{\top} \Sigmax \boldsymbol{\theta}_0\right)} 
    \int_{\RR^r} \prod_{a=1}^r \frac{\dd{m^a} \dd{\hat{m}^a}}{2 \pi} e^{-i \sum_{a=1}^r \hat{m}^a\left(d m^a-\boldsymbol{\theta}_0^{\top} \Sigmax \w^a\right)} 
    \int_{\RR^{r \times r}} \\ 
    &\prod_{1 \leq a \leq b \leq r} \frac{\mathrm{d} Q^{a b} \dd{\hat{Q}^{a b}} }{2 \pi} e^{-i \hat{Q}^{a b}\left(d Q^{a b}-\w^{a \top} \Sigmax \w^b\right)}
\end{aligned}
\end{equation}

We also would like to define new overlaps which are
\begin{equation}
    P = \frac{1}{d} \w^\top \Sigmadelta \w \,,\quad 
    A = \frac{1}{d} \w^\top \Sigmaupsilon \w \,,\quad 
    F = \frac{1}{d} \boldsymbol{\theta}_0^{\top} \Sigmaupsilon \w \,,\quad 
    N = \frac{1}{d} \w^{\top} \w \,,
\end{equation}
which enter into the computation as follows
\begin{equation}
\begin{aligned}
    1 \propto & \int \prod_{a=1}^r \dd{P^{a}} \delta\qty(d P^{a} - \w^a \Sigmadelta \w^a )
    \int \prod_{a=1}^r \dd{A^{a}} \delta\qty(d A^{a} - \w^a \Sigmaupsilon \w^a ) \\
    &\int_{\RR^r} \prod_{a=1}^r \dd{F^a} \delta\left( d F^a - \boldsymbol{\theta}_0^{\top} \Sigmaupsilon \w^a \right) 
    \int \prod_{a=1}^r \dd{N^{a}} \delta\qty(d N^{a} - \w^a \cdot \w^a ) \\
    = & \int \prod_{a=1}^r \frac{\dd{P^{a}} \dd{\hat{P}^{a}}}{2\pi} e^{ -i \hat{P}^{a} (d P^{a} - \w^a \Sigmadelta \w^a ) }
    \int \prod_{a=1}^r \frac{\dd{A^{a}} \dd{\hat{A}^{a}}}{2\pi} e^{ -i \hat{A}^{a} (d A^{a} - \w^a \Sigmaupsilon \w^a ) } \\
    & \int_{\RR^r} \prod_{a=1}^r \frac{\dd{F^a} \dd{\hat{F}^a}}{2 \pi} e^{-i \sum_{a=1}^r \hat{F}^a\left(d F^a-\boldsymbol{\theta}_0^{\top} \Sigmaupsilon \w^a\right)} 
    \int \prod_{a=1}^r \frac{\dd{N^{a}} \dd{\hat{N}^{a}}}{2\pi} e^{ -i \hat{N}^{a} (d N^{a} - \w^a \cdot \w^a) }
\end{aligned}
\end{equation}
notice that the overlap \(N\) enters the computation only if we are considering the FGM attack and not for the true minimisation attack. In the following of this computation we will try to remain as general as possible considering both cases.

We finally can write our replicated partition function as the integral of a functional as follows
\begin{equation}\label{eq:integral-replicated-partition-fun}
    \mathbb{E}_{\mathcal{D}} \mathcal{Z}_\beta^r = 
    \int \frac{\dd{\rho} \mathrm{d} \hat{\rho}}{2 \pi} 
    \prod_{a=1}^{r} \frac{\dd{m^a} \dd{\hat{m}^a}}{2 \pi} \frac{\dd{P^{a}} \dd{\hat{P}^{a}}}{2\pi} \frac{\dd{A^{a}} \dd{\hat{A}^{a}}}{2\pi} \frac{\dd{F^{a}} \dd{\hat{F}^{a}}}{2\pi} \frac{\dd{N^{a}} \dd{\hat{N}^{a}}}{2\pi}
    \prod_{1 \leq a \leq b \leq r} \frac{\mathrm{d} Q^{a b} \dd{\hat{Q}^{a b}} }{2 \pi} 
    e^{d \Phi^{(r)}}
\end{equation}
where the $r$ times replicated functional \(\Phi^{(r)}\) is
\begin{equation}\label{eq:replicated-free-energy}
\begin{aligned}
    \Phi^{(r)} &= - \rho \hat{\rho} - \sum_{a=1}^{r} m^a \hat{m}^a - \sum_{1 \leq a \leq b \leq r} Q^{a b} \hat{Q}^{a b} - \sum_{a=1}^{r} N^{a} \hat{N}^{a} - \sum_{a=1}^{r} A^{a} \hat{A}^{a} - \sum_{a=1}^{r} P^{a} \hat{P}^{a} - \sum_{a=1}^{r} F^{a} \hat{F}^{a} \\
    &+ \alpha \Psi_y^{(r)}\left(\rho, m^a, Q^{a b}, A^a, N^{a}, P^A, F^a \right)
    + \Psi_w^{(r)}\left(\hat{\rho}, \hat{m}^a, \hat{Q}^{a b}, \hat{A}^a, \hat{N}^{a}, \hat{P}^a, \hat{F}^a \right)
\end{aligned}
\end{equation}
we will refer to the elements in the first line of \cref{eq:replicated-free-energy} as the trace term. Note in \cref{eq:integral-replicated-partition-fun} we factored out $d$ such that we can later evaluate the partition function in the thermodynamic limit using Laplace's method.
We also have defined the prior part of the free energy \(\Psi_w\) to be
\begin{equation}\label{eq:prior-part}
\begin{aligned}
    \Psi_w^{(r)} = & \frac{1}{d} \log \left[
        \int_{\RR^d} P_{\boldsymbol{\theta}_0} \left(\dd{\boldsymbol{\theta}_0}\right) e^{\hat{\rho} \boldsymbol{\theta}_0^{\top} \Sigmax \boldsymbol{\theta}_0} \right.\\
        & \left. \int_{\RR^{d \times r}} \prod_{a=1}^r P_w\left(\dd{\w^a}\right) e^{
            \sum_{a=1}^{r} \left(
                \hat{m}^a \boldsymbol{\theta}_0^{\top} \Sigmax \w^a +
                \hat{A}^{a} \w^{a \top} \Sigmaupsilon \w^a + 
                \hat{P}^{a} \w^{a \top} \Sigmadelta \w^a + 
                \hat{F}^{a} \w^{a \top} \Sigmaupsilon \w^a + 
                \hat{N}^{a} \w^a \cdot \w^a
            \right)            
            +\sum_{1 \leq a \leq b \leq r} \left(
                \hat{Q}^{a b} \w^{a \top} \Sigmax \w^b 
            \right)    }
            \right] \\
\end{aligned}
\end{equation}
and the channel part of the free energy \(\Psi_y\) as
\begin{equation}\label{eq:channel-part}
    \Psi_y^{(r)} = \log \qty[
        \int_{\RR} \dd{y} \int_{\RR} \mathrm{d} \nu P_0(y \mid \nu) \int \prod_{a=1}^r \mathrm{~d} \lambda^a P_g\left(y \mid \lambda^a, P^{a}, N^{a}, \advtrainingcost \right) \mathcal{N}\left(\nu, \lambda^a ; \mathbf{0}, \Sigma^{a b}\right)
    ]
\end{equation}
where we have used the fact that \((\nu_\mu, \lambda_\mu) \: \mu = 1, \dots n\) factors over all the data points.

In the thermodynamic limit where $d\to \infty$ with $n/d$ fixed, the integral in \cref{eq:integral-replicated-partition-fun} concentrates around the values of the overlap parameters that extremize the free entropy \(\Phi^{(r)}\) and hence we can get the free energy density as:
\begin{equation}
    \beta f_\beta = - \lim_{r\to 0^+} \frac{1}{r} \operatorname{extr} \Phi^{(r)} = - \lim_{r\to 0^+} \partial_r \operatorname{extr} \Phi^{(r)}
\end{equation}

\subsection{Replica Symmetric Ansatz}

We propose the following Ansatz for the variables that we have to extremise over
\begin{equation}
\begin{array}{rrr}
    m^a=m & \hat{m}^a=\hat{m} & \text { for } a=1, \cdots, r \\
    q^{a a} = Q & \hat{q}^{a a} = -\frac{1}{2} \hat{Q} & \text { for } a=1, \cdots, r \\
    q^{a b} = q & \hat{q}^{a b}=\hat{q} & \text { for } 1 \leq a<b \leq r \\ 
    P^{a} = P & \hat{P}^{a} = -\frac{1}{2} \hat{P} & \text { for } a=1, \cdots, r \\
    N^{a} = N & \hat{N}^{a} = -\frac{1}{2} \hat{N} & \text { for } a=1, \cdots, r \\
    A^{a} = A & \hat{A}^{a} = -\frac{1}{2} \hat{A} & \text { for } a=1, \cdots, r \\
    F^{a} = F & \hat{F}^{a} = \hat{F} & \text { for } a=1, \cdots, r \\
\end{array}
\end{equation}
Before we take the replica zero limit, let's check that our ansatz above is well-defined and does not have an order one term in $\Phi^{(r)}$ that diverges. For this, we need to ensure that $\lim _{r \rightarrow 0^{+}} \Phi^{(r)}=0$. The trace terms depends on $r$ except for $\rho \hat{\rho}$, $\lim _{r \rightarrow 0^{+}} \Psi_y^{(r)} = \lim _{r \rightarrow 0^{+}} \Psi_w^{(r)} =0$ hold. Note that $\lim_{r \rightarrow 0^{+}} \log C^r = 0$ but $\lim_{r \rightarrow 0^{+}} \partial_r \log C^r = \log C $, and thus all we need to check is the prior part of the free energy in the zero replica limit.

\begin{equation}
\lim _{r \rightarrow 0^{+}} \Phi^{(r)}= -\rho \hat{\rho}
\end{equation}
For this limit to be zero, we must fix $\hat{\rho} = 0$ and note that $\rho$ is a constant we fixed earlier. 

Plugging in the Ansatz, the trace term becomes
\begin{equation}
     -\rho \hat{\rho} - r m \hat{m} - \frac{r(r-1)}{2} q \hat{q} + \frac{r}{2} Q \hat{Q} + \frac{r}{2} A \hat{A} + \frac{r}{2} N \hat{N} + \frac{r}{2} P \hat{P} - r F \hat{F}
\end{equation}

Now we take the limit \(r \to 0\) after dividing the trace term $T$ (which is no longer an actual trace as we introduced overlaps beyond the traditional replica matrix ansatz) by $r$
\begin{equation}
\begin{aligned}
   T = &\frac{1}{2}( q\hat{q} + Q\hat{Q} ) + \frac{1}{2}P\hat{P} + \frac{1}{2}A\hat{A} + \frac{1}{2}N\hat{N}  - m \hat{m} - F \hat{F}\\
    \end{aligned}
\end{equation}
We now define \(V = Q - q\) and \(\hat{V} = \hat{Q} + \hat{q}\) and rewrite the trace term as follows (by replacing \(q\) and \(\hat{Q}\)).
\begin{equation}
\begin{aligned}
   T & = \frac{1}{2}( q\hat{q} + (V+q)(\hat{V} - \hat{q}) ) + \frac{1}{2}P\hat{P} + \frac{1}{2}A\hat{A} + \frac{1}{2}N\hat{N}  - m \hat{m} - F \hat{F}\\
    & = \frac{1}{2}(V\hat{V} + q\hat{V} - V\hat{q} ) + \frac{1}{2}P\hat{P} + \frac{1}{2}A\hat{A} + \frac{1}{2}N\hat{N}  - m \hat{m} - F \hat{F}\\
\end{aligned}
\end{equation}

\subsubsection{Prior Replica Zero Limit}

Thus we can proceed plug these ansätze inside \cref{eq:channel-part,eq:prior-part} we obtain the following for the prior term

\begin{equation}\label{eq:prior-part-rs-ansatz}
\begin{aligned}
    \Psi_w^{(r)} = \frac{1}{d} \log 
    \left[
        \int_{\RR^d} 
        P_{\boldsymbol{\theta}_0} 
        \left(\dd{\boldsymbol{\theta}_0}\right) 
    e^{\hat{\rho} \boldsymbol{\theta}_0^{\top} \Sigmax \boldsymbol{\theta}_0} 
    \right. &
    \int_{\RR^{d \times r}} 
    \prod_{a=1}^r P_w\qty(\dd{\w^a}) 
    e^{
        \sum_{a=1}^{r} \qty(
            \hat{m} \boldsymbol{\theta}_0^{\top} \Sigmax \w^a
            \hat{F} \w^{a \top} \Sigmaupsilon \w^a
        )            
        +\sum_{1 \leq a < b \leq r} \qty(
            \hat{q} \w^{a \top} \Sigmax \w^b
        ) 
    } \\
    &\left. e^{ - \frac{1}{2} \sum_{a=1}^{r} \qty(
        \hat{Q} \w^{a \top} \Sigmax \w^a + 
        \hat{A} \w^{a \top} \Sigmaupsilon \w^a +                  
        \hat{P} \w^{a \top} \Sigmadelta \w^a + 
        \hat{N} \w^a \cdot \w^a 
    ) }
    \right]
\end{aligned}
\end{equation}

to perform in the following the \(r \to 0^+\) limit we can change a bit the integral by factoring out all the terms.

To perform this simplification we will use the multidimensional Hubbard-Stratonovic identity which reads
\begin{equation}
    e^{ \frac{1}{2} \sum_{a, b=1}^{r} \w^{a \top} \left[ \hat{q}  \Sigmax \right] \w^b} = 
    \mathbb{E}_{\boldsymbol{\xi}} \qty[
        e^{\boldsymbol{\xi}^{\top} \, \sqrt{ \hat{q}  \Sigmax} \sum_{a=1}^{r} \w^a}
    ]
\end{equation}
where \(\boldsymbol{\xi} \sim \mathcal{N}(\boldsymbol{0}, 1_d)\). 

Thus by calling the part inside the \(\log\) in \cref{eq:prior-part-rs-ansatz} with the letter \(\mathcal{A}\) we have that (putting in $\hat{\rho} = 0$)
\begin{equation}
\begin{aligned}
    \mathcal{A} =& \mathbb{E}_{\boldsymbol{\theta}_0} 
    \int_{\RR^{d \times r}} \prod_{a=1}^r P_w\left(\dd{\w^a}\right) 
    e^{
        -\sum_{a=1}^{r} \left(
            \hat{m} \boldsymbol{\theta}_0^{\top} \Sigmax \w^a 
            -\hat{F} \boldsymbol{\theta}_0^{\top} \Sigmaupsilon \w^a 
        \right)
         + \frac{1}{2} \sum_{1 \leq a, b \leq r} 
            \hat{q} \w^{a \top} \Sigmax \w^b
            } \\
            & \cdot e^{
         - \frac{1}{2} \sum_{a=1}^{r} \left(
            \hat{V} \w^{a \top} \Sigmax \w^a + 
            \hat{A} \w^{a \top} \Sigmaupsilon \w^a +
            \hat{P} \w^{a \top} \Sigmadelta \w^a +
            \hat{N} \w^a \cdot \w^a 
        \right)
    } \\
    =&  \mathbb{E}_{\boldsymbol{\theta}_0} 
    \int_{\RR^{d \times r}} \prod_{a=1}^{r} P_w\left(\dd{\w^a}\right) 
    e^{
        -\sum_{a=1}^{r} \qty( 
            \frac{\hat{V}}{2} \wavec \Sigmax \wavec + \frac{\hat{P}}{2}  \wavec \Sigmadelta \wavec + \frac{\hat{A}}{2}  \wavec \Sigmaupsilon \wavec + \frac{\hat{N}}{2}  \wavec \wavec + \hat{m} \boldsymbol{\theta}_0 \Sigmax \wavec + \hat{F} \boldsymbol{\theta}_0 \Sigmaupsilon \wavec
        )
        + \frac{1}{2} \sum_{1 \leq a, b \leq r} 
            \hat{q} \w^{a \top} \Sigmax \w^b
    } \\ 
    =&  \mathbb{E}_{\boldsymbol{\theta}_0} 
    \int_{\RR^{d \times r}} \prod_{a=1}^{r} P_w\left(\dd{\w^a}\right) \mathbb{E}_{\boldsymbol{\xi}} \qty[
        e^{
            - \frac{\hat{V}}{2} \wavec \Sigmax \wavec - \frac{\hat{P}}{2}  \wavec \Sigmadelta \wavec - \frac{\hat{A}}{2}  \wavec \Sigmaupsilon \wavec - \frac{\hat{N}}{2}  \wavec \wavec - \wavec \qty( \hat{m} \Sigmax \boldsymbol{\theta}_0 + \hat{F} \Sigmaupsilon \boldsymbol{\theta}_0 - \sqrt{ \hat{q}  \Sigmax }  \boldsymbol{\xi} ) 
        }
    ] \\ 
    =& \mathbb{E}_{\boldsymbol{\xi},\boldsymbol{\theta}_0} \qty[
        \qty[
            \int_{\RR^{d}} P_w\left(\dd{\w}\right)
            e^{
                -\frac{\hat{V}}{2} \w \Sigmax \w - \frac{\hat{P}}{2}  \w \Sigmadelta \w - \frac{\hat{A}}{2}  \w \Sigmaupsilon \w - \frac{\hat{N}}{2}  \w \w - \w \qty( \hat{m} \Sigmax \boldsymbol{\theta}_0 + \hat{F} \Sigmaupsilon \boldsymbol{\theta}_0 - \sqrt{ \hat{q}  \Sigmax }  \boldsymbol{\xi} ) 
            }
        ]^r
    ]
\end{aligned}
\end{equation}
Then we can take the derivative and limit and obtain 
\begin{equation}\label{eq:prior-term-after-r-limit}
    \Psi_w = \lim_{r \to 0^+} \partial_r \Psi_w^{(r)} = \frac{1}{d} \mathbb{E}_{\boldsymbol{\xi}, \boldsymbol{\theta}_0} \qty[
        \log \int_{\RR^{d}} P_w\left(\dd{\w}\right)
        e^{
            -\frac{\hat{V}}{2} \w \Sigmax \w - \frac{\hat{P}}{2}  \w \Sigmadelta \w - \frac{\hat{A}}{2}  \w \Sigmaupsilon \w - \frac{\hat{N}}{2}  \w \w - \w \qty( \hat{m} \Sigmax \boldsymbol{\theta}_0 + \hat{F} \Sigmaupsilon \boldsymbol{\theta}_0 - \sqrt{ \hat{q}  \Sigmax  }  \boldsymbol{\xi} ) 
        }
    ]
\end{equation}
where we still need to take the limit \(d \to \infty\). 

\subsubsection{Channel Replica Zero Limit}

Now we can focus on the channel term and rewrite it in a more suitable way for taking the \(r \to 0^+\) limit. In a very similar fashion as before we would like to simplify
\begin{equation}
    \Psi_y^{(r)} = \log \qty[
        \int_{\RR} \dd{y} \int_{\RR} \dd{\nu} P_0(y \mid \nu) \int \prod_{a=1}^r \dd{\lambda^a} P_g\left(y \mid \lambda^a, P, N, \advtrainingcost \right) \mathcal{N}\left(\nu, \lambda^a ; \mathbf{0}, \Sigma^{a b}\right)
    ]
\end{equation}
We will indicate the argument of the \(\log\) with \(\mathcal{B}\). Additionally we have that the martix of covariances is
\begin{equation}
    \Sigma = \begin{pmatrix}
        \rho & m & m & \dots & m \\
        m & Q & q & \dots & q \\
        m & q & Q & \dots & q \\
        \vdots & \vdots & \vdots & \ddots & \vdots \\
        m & q & q & \dots & Q \\
    \end{pmatrix}
\end{equation}
and in addition also the inverse matrix has a Replica Symmetric Structure which is given from the following elements
\begin{equation}
\begin{aligned}
    \qty(\Sigma^{-1})_{00} \equiv \tilde{\rho} &= \frac{Q + (r-1) q}{\rho(Q + (r-1) q)- r m^2} \,, & 
    \qty(\Sigma^{-1})_{0a} \equiv \tilde{m} &= \frac{m}{r m^2 - \rho (Q + (r-1) q)} \,,\\
    \qty(\Sigma^{-1})_{aa} \equiv \tilde{Q} &= \frac{\rho(Q + (r-2) q) - (r-1) m^2}{(Q-q) \left(\rho (Q + (r-1) q) - r m^2 \right)} \,,& 
    \qty(\Sigma^{-1})_{ab} \equiv \tilde{q} &= \frac{m^2 - \rho q}{(Q - q)\left(\rho (Q + (r-1) q) - r m^2 \right)} \\
\end{aligned}
\end{equation}
and thus there is an implicit dependence on $r$ in the covariance. To check that the inverse matrix has a RS structure as well one can think of the formula that is used to evaluate the inverse of a matrix from the cofactors.

Also we look at the determinant of the matrix. There are three different eigenvalue types

\begin{equation}
\begin{aligned}
    \lambda_1 &= Q - q\,, \quad & \lambda_2 &= \frac{1}{2}\qty(-Q - q(r-1) - \rho - \tilde{\Delta}) \,, \quad & \lambda_3 &= \frac{1}{2}\qty(-Q - q(r-1) - \rho + \tilde{\Delta}) \,, \\
    d_1 &= r - 1 \,, \quad & d_2 &= 1\,, \quad & d_3 &= 1\,, 
\end{aligned}
\end{equation}
with \(\tilde{\Delta} = \sqrt{4 m^2 r +(Q + q(r-1) - \rho)^2}\) and thus one obtains the determinant. More explicitly we have that
\begin{equation}
\begin{aligned}
    \det \qty(2\pi \Sigma) & = (2\pi)^{r+1} (Q - q)^{r-1} \frac{1}{4} (-Q - q(r-1) - \rho - \tilde{\Delta}) (-Q - q(r-1) - \rho + \tilde{\Delta}) \\
    & = (2\pi)^{r+1} (Q - q)^{r-1} ( \rho ( Q + (r-1)q) - rm^2 )
\end{aligned}
\end{equation}

Thus we have that
\begin{equation}
\begin{aligned}
    \mathcal{B} &= \int_{\RR} \dd{y} \int_{\RR} \dd{\nu} P_0(y \mid \nu) 
    e^{-\frac{1}{2} \tilde{\rho} \nu^2}
    \int \prod_{a=1}^{r} \dd{\lambda^a} P_g\left(y \mid \lambda^a, P, N, \advtrainingcost \right) \\
    & e^{
        - \tilde{m} \nu \sum_{a=1}^{r} \lambda^a - \frac{1}{2} \tilde{Q} \sum_{a=1}^{r} (\lambda^a)^2 - \frac{1}{2} \tilde{q} \sum_{1 \leq a,b \leq r, a \neq b} \lambda^a \lambda^b - \frac{1}{2} \log\det(2\pi \Sigma)
    } \\
    &= \mathbb{E}_{\xi} \int_{\RR} \dd{y} e^{- \frac{1}{2} \log\det(2\pi \Sigma)}
    \int_{\RR} \dd{\nu} P_0(y \mid \nu) 
    e^{-\frac{1}{2} \tilde{\rho} \nu^2}
    \qty[
        \int \dd{\lambda} P_g \left(y \mid \lambda, P, N, \advtrainingcost \right)
        e^{-\frac{\tilde{Q}-\tilde{q}}{2} \lambda^2 + (\sqrt{-\tilde{q}} \xi - \tilde{m} \nu) \lambda}
    ]^r
\end{aligned}
\end{equation}

Now we can follow a similar procedure as before and define \(V = Q - q\) we have that
and the limit is
\begin{equation}
\begin{aligned}
    \Psi_y &= \lim_{r \to 0^+} \partial_r \Psi_y^{(r)} = \mathbb{E}_{\xi} \qty[
        \int_{\RR} \dd{y} \int \frac{\dd{\nu}}{\sqrt{2 \pi \rho}} P_0(y \mid \nu) e^{-\frac{1}{2 \rho} \nu^2}
        \log \qty[
            \int \frac{\dd{\lambda}}{\sqrt{2 \pi}} P_g(y \mid \lambda, P, N, \advtrainingcost ) e^{-\frac{1}{2} \frac{\lambda^2}{V} + \left(\frac{\sqrt{q-m^2 / \rho}}{V} \xi + \frac{m / \rho}{V} \nu \right) \lambda}
        ]
    ] \\
    &-\frac{1}{2} \log V - \frac{1}{2} \frac{q}{V}
\end{aligned}
\end{equation}

We would like to rewrite the quantities with the help of the following definition
\begin{equation}\label{eq:partition-function-definitions}
    \mathcal{Z}_0(y, \omega, V) = \int \frac{\mathrm{d} x}{\sqrt{2 \pi V}} e^{-\frac{1}{2 V}(x-\omega)^2} P_0(y \mid x)\,, \quad 
    \mathcal{Z}_y(y, \omega, V, P, n) = \int \frac{\mathrm{d} x}{\sqrt{2 \pi V}} e^{-\frac{1}{2 V}(x-\omega)^2} P_g(y \mid x, P, N, \advtrainingcost)\,,
\end{equation}

The result becomes thus
\begin{equation}
    \mathbb{E}_{\xi} \qty[
        \int_{\RR} \dd{y} \mathcal{Z}_0 \qty(y, \frac{m}{\sqrt{q}} \xi, \rho - \frac{m^2}{q})
        \log \mathcal{Z}_y (y, \sqrt{q} \xi, V, P, N)
    ]
\end{equation}

Now there are two things that we still need to do : find the form for the prior term and take the limit \(\beta \to \infty\).

\subsection{Prior term for \texorpdfstring{\(\ell_2\)}{L2} regularisation}

To be as general as possible we would like to include the case of a possible non isotropic regularisation. Thus
\begin{equation}
    P_w(\dd{\w}) = \frac{1}{(2 \pi)^{d / 2}} \exp(-\frac{\beta \lambda}{2}\w \Sigmaw \w ) \dd{\w}
\end{equation}
We want to calculate the term inside the \(\log\) in \cref{eq:prior-term-after-r-limit}
\begin{equation}
\begin{aligned}
    & \int_{\RR^d} P_w(\dd{\w}) 
    e^{
        -\frac{\hat{V}}{2} \w \Sigmax \w - \frac{\hat{P}}{2}  \w \Sigmadelta \w - \frac{\hat{A}}{2}  \w \Sigmaupsilon \w - \frac{\hat{N}}{2}  \w \w - \w \qty( \hat{m} \Sigmax \boldsymbol{\theta}_0 + \hat{F} \Sigmaupsilon \boldsymbol{\theta}_0 -  \sqrt{ \hat{q}  \Sigmax }  \boldsymbol{\xi} ) 
    } \\ & = \frac{
        \exp 
        \left(
            \frac{1}{2}\left(-\hat{m} \Sigmax^{\top} \boldsymbol{\theta}_0 - \hat{F} (\Sigmaupsilon)^{\top} \boldsymbol{\theta}_0 +  \sqrt{ \hat{q}  \Sigmax }  \boldsymbol{\xi} \right)^{\top}\boldsymbol{\Lambda}^{-1} \left(-\hat{m} \Sigmax^{\top} \boldsymbol{\theta}_0 - \hat{F} (\Sigmaupsilon)^{\top} \boldsymbol{\theta}_0  +\sqrt{ \hat{q}  \Sigmax  }  \boldsymbol{\xi} \right)^{\top}
        \right)
    }{
        \sqrt{\det \boldsymbol{\Lambda}}
    }
\end{aligned}
\end{equation}
where we defined \(\boldsymbol{\Lambda} = \beta \lambda \Sigmaw + \hat{V} \Sigmax + \hat{P} \Sigmadelta + \hat{A} \Sigmaupsilon + \hat{N} \mathbb{1}\). Now the prior term becomes after taking the log and using the identity $\log \operatorname{det}=\tr \log$
\begin{equation}
\begin{aligned}
    \Psi_w &= \frac{1}{d} \mathbb{E}_{\boldsymbol{\xi},\boldsymbol{\theta}_0}
    \qty[
        \frac{1}{2}\left(-\hat{m} \Sigmax^{\top} \boldsymbol{\theta}_0 - \hat{F} (\Sigmaupsilon)^{\top} \boldsymbol{\theta}_0 +  \sqrt{ \hat{q}  \Sigmax  }  \boldsymbol{\xi} \right)^{\top}\boldsymbol{\Lambda}^{-1} \left(-\hat{m} \Sigmax^{\top} \boldsymbol{\theta}_0 - \hat{F} (\Sigmaupsilon)^{\top} \boldsymbol{\theta}_0 +  \sqrt{ \hat{q}  \Sigmax  }  \boldsymbol{\xi}\right)^{\top} ]- \frac{1}{2d} \tr\log \boldsymbol{\Lambda} \\
    &= \frac{1}{2 d} \tr \qty[
        \left(\hat{m}^2 \Sigmax^{\top} \boldsymbol{\theta}_0 \boldsymbol{\theta}_0^{\top} \Sigmax + \hat{m}\hat{F} \Sigmax^{\top} \boldsymbol{\theta}_0 \boldsymbol{\theta}_0^{\top} \Sigmaupsilon + \hat{m}\hat{F} (\Sigmaupsilon)^{\top} \boldsymbol{\theta}_0 \boldsymbol{\theta}_0^{\top} \Sigmax + \hat{F}^2 (\Sigmaupsilon)^{\top} \boldsymbol{\theta}_0 \boldsymbol{\theta}_0^{\top} \Sigmaupsilon + \hat{q}  \Sigmax \right) \boldsymbol{\Lambda}^{-1}
    ] \\
    & - \frac{1}{2d} \tr\log \boldsymbol{\Lambda} 
\end{aligned}
\end{equation}
The factor $\frac{1}{d}$ comes from the required scaling on $d$ for the free entropy and the expectation from our replica zero limit of the prior term.

\subsection{Zero temperature limit}

We now need to take the zero temperature limit for this case. The explicit scalings of the parameters are
\begin{equation}\label{eq:temperature-scalings}
\arraycolsep=7.5pt\def\arraystretch{1.4}
\begin{array}{rrrrrrr}
    V \rightarrow \beta^{-1} V & 
    q \rightarrow q & 
    m \rightarrow m & 
    A \rightarrow A &
    N \rightarrow N &
    P \rightarrow P &
    F \rightarrow F \\
    \hat{V} \rightarrow \beta \hat{V} & 
    \hat{q} \rightarrow \beta^2 \hat{q} & 
    \hat{m} \rightarrow \beta \hat{m} & 
    \hat{A} \rightarrow \beta \hat{A} &
    \hat{N} \rightarrow \beta \hat{N} & 
    \hat{P} \rightarrow \beta \hat{P} &
    \hat{F} \rightarrow \beta \hat{F}\\
\end{array}
\end{equation}

The limit of the prior term is
\begin{equation}
\begin{aligned}
    &\Psi_w = \lim_{\beta\to\infty} \frac{1}{\beta} \Psi_w = \\
    &\frac{1}{2 d} \tr \qty[
        \left(\hat{m}^2 \Sigmax^{\top} \boldsymbol{\theta}_0 \boldsymbol{\theta}_0^{\top} \Sigmax  + \hat{m}\hat{F} \Sigmax^{\top} \boldsymbol{\theta}_0 \boldsymbol{\theta}_0^{\top} \Sigmaupsilon + \hat{m}\hat{F} (\Sigmaupsilon)^{\top} \boldsymbol{\theta}_0 \boldsymbol{\theta}_0^{\top} \Sigmax + \hat{F}^2 (\Sigmaupsilon)^{\top} \boldsymbol{\theta}_0 \boldsymbol{\theta}_0^{\top} \Sigmaupsilon + \hat{q} \Sigmax\right) \boldsymbol{\Lambda}^{-1}
    ]
\end{aligned}
\end{equation}
To understand the limit of the channel term, we need to get the following insight for the limit of the channel partition function
\begin{equation}
\begin{aligned}
    \mathcal{Z}_y(y, \omega, V, P, N) & = \int \frac{\mathrm{d} x}{\sqrt{2 \pi V}} e^{-\frac{\beta}{2 V}(x-\omega)^2} P_g(y \mid x, P, N, \advtrainingcost) \\
    & = \sqrt{\beta} \int \frac{\mathrm{d} x}{\sqrt{2 \pi V}} e^{-\frac{\beta}{2 V}(x-\omega)^2} \frac{1}{\sqrt{2 \pi}} e^{( 
        - \beta g(
            y x - \frac{\advtrainingcost}{\sqrt{d}} \frac{P}{\sqrt{N}}) )} \\    
    & \underset{\beta \rightarrow \infty}{=} e^{-\beta \mathcal{M}_{V g(y, \cdot)}(\omega)} 
\end{aligned}
\end{equation}
where we introduced the Moreau envelope defined in \cref{eq:moreau-proximal-definitions}. Notice that the previous computation is done for the case of the FGM attack and it doesn't change much in the case of the true minimisation.

Then the limit of the channel term becomes
\begin{equation}
    \Psi_y = \lim_{\beta\to\infty} \frac{1}{\beta} \Psi_y = - \mathbb{E}_{\xi} 
    \left[
        \int \dd{y} \mathcal{Z}_0\left(y, \frac{m}{\sqrt{q}} \xi, \rho-\frac{m^2}{q}\right) \mathcal{M}_{V g(y, \cdot; P, N, \advtrainingcost)}(\sqrt{q} \xi)
    \right]
\end{equation}
where \(\mathcal{M}_{V g(y, \cdot; A, N, \advtrainingcost)}\) is the Moreau envelope of the modified loss function defined in \cref{eq:modified-minimisation-problem} with the relevant quantities changed for their overlaps and \(\xi \sim \mathcal{N}(0,1)\). 

After taking the zero temperature limit, we are left with the following expression for the free energy density
\begin{equation}\label{eq:free-energy-density-after-limit}
\begin{aligned}
    \lim _{\beta \rightarrow \infty} f_\beta &= \mathop{\operatorname{extr}}_{\substack{V, q, m, A, N, P, F,\\ \hat{V}, \hat{q}, \hat{m}, \hat{A}, \hat{N}, \hat{P}, \hat{F}}}
    \left\{-\frac{1}{2}(q \hat{V}-\hat{q} V) - \frac{1}{2}P\hat{P} - \frac{1}{2}A\hat{A} - \frac{1}{2}N\hat{N} + m \hat{m} + F\hat{F} +\alpha \mathbb{E}_{\xi}\left[\int \dd{y} \mathcal{Z}_0 \mathcal{M}_{V g(y, \cdot)}\right]\right. \\
    &\left.-\frac{1}{2 d} \tr\left(\hat{m}^2 \Sigmax^{\top} \boldsymbol{\theta}_0 \boldsymbol{\theta}_0^{\top} \Sigmax  + \hat{m}\hat{F} \Sigmax^{\top} \boldsymbol{\theta}_0 \boldsymbol{\theta}_0^{\top} \Sigmaupsilon + \hat{m}\hat{F} (\Sigmaupsilon)^{\top} \boldsymbol{\theta}_0 \boldsymbol{\theta}_0^{\top} \Sigmax + \hat{F}^2 (\Sigmaupsilon)^{\top} \boldsymbol{\theta}_0 \boldsymbol{\theta}_0^{\top} \Sigmaupsilon + \hat{q} \Sigmax\right)\boldsymbol{\Lambda}^{-1}\right\}
\end{aligned}
\end{equation}

\subsection{Saddle-point equations}

The extremisation condition of \cref{eq:free-energy-density-after-limit} can be translated into the overlap needing to satisfy the following
\begin{equation}\label{eq:fixed_point_ set of equations.equations}
\begin{aligned}
\hat{V} & = 2 \alpha \partial_q \Psi_y, & & q= - 2 \partial_{\hat{V}} \Psi_w \\
\hat{q} & = - 2 \alpha \partial_{V} \Psi_y, & & V= 2 \partial_{\hat{q}} \Psi_w, \\
\hat{N} & = 2 \alpha \partial_N \Psi_y, & & N= -2 \partial_{\hat{N}} \Psi_w \\
\hat{P} & = 2 \alpha \partial_P \Psi_y, & & P= -2 \partial_{\hat{P}} \Psi_w \\
\hat{A} & = 2 \alpha \partial_A \Psi_y, & & A= -2 \partial_{\hat{A}} \Psi_w \\
\hat{m} & = - \alpha \partial_m \Psi_y, & & m= \partial_{\hat{m}} \Psi_w \\
\hat{F} & = - \alpha \partial_F \Psi_y, & & F= \partial_{\hat{F}} \Psi_w .
\end{aligned}    
\end{equation}

As we pre-announced we would like to find the stationary values that dominate the integral and to do so we should derive the exponent with respect to all the order parameters.
The saddle points that depend on \(m,q,V,\hat{m},\hat{q}\) and \(\hat{V}\) are of a similar form as those found already in \cite{Loureiro_2022}. We need thus to derive with respect to \(A,N,P,F,\hat{A},\hat{P},\hat{F}\) and \(\hat{N}\).

\subsubsection{The Channel Saddle-Point Equations}

Let us begin by looking at the derivatives with respect to \(P\) and \(N\). These derivatives amount to computing the derivative of the Moreau-envelope with respect to \(P\) and \(N\) since we have that
\begin{equation}
\begin{aligned}
    \partial_P \Psi_{y} & = \mathbb{E}_{y, \xi}\left[
        \mathcal{Z}_0 \qty(y, \frac{m}{\sqrt{q}} \xi, \rho - \frac{m^2}{q})
        \partial_P \mathcal{M}_{V g(y, \cdot; P, N, \advtrainingcost)}(\sqrt{q} \xi) 
    \right] \,, \\
    \partial_N \Psi_{y} & = \mathbb{E}_{y, \xi}\left[
        \mathcal{Z}_0 \qty(y, \frac{m}{\sqrt{q}} \xi, \rho - \frac{m^2}{q})
        \partial_N \mathcal{M}_{V g(y, \cdot; P, N, \advtrainingcost)}(\sqrt{q} \xi) 
    \right] \,, \\
\end{aligned}
\end{equation}

Let's focus on the true minimisation. In this specific case we have that 
\begin{equation}\label{eq:moreau-envelope-shift-propriety}
    \mathcal{M}_{V g(y, \cdot, P, N, \advtrainingcost)}(\omega) = 
    \inf _{x \in \RR}\left[
        \frac{(x-\omega)^2}{2 V} + 
        \ell \qty(y x - \advtrainingcost \sqrt{P})
    \right] = \mathcal{M}_{V \ell(y, \cdot)} \qty(
        \omega - y \advtrainingcost \sqrt{P}
    )
\end{equation}
where we remind that this specific form is possible since \(y \in \qty{+ 1, -1}\). With this we can relate these function the the derivative of the Moreau envelope with respect to its input as
\begin{equation}
    \partial_P \mathcal{M}_{V g(y,\cdot, P, N, \advtrainingcost)}(\omega) = 
    -y \frac{\advtrainingcost}{2\sqrt{P}} \mathcal{M}^\prime_{V \ell(y, \cdot)}\qty(\omega - y \advtrainingcost \sqrt{P}) 
\end{equation}

With this, we can write the new equation as
\begin{equation}
    \hat{P} = \alpha \frac{\advtrainingcost}{2\sqrt{P}}\mathbb{E}_{\xi}\left[
        \int_{\RR} \dd{y} y \mathcal{Z}_0 f_g
    \right] 
\end{equation}
where we have defined
\begin{equation}
    f_g(y, \omega, V, P, N, \advtrainingcost) = 
    - \mathcal{M}^\prime_{V g(y, \cdot, P, N, \advtrainingcost)}(\omega)
\end{equation}
In the case of the FGM we will also have an equation for \(N\) that can be derived similarly.

As the channel term does not depend on the overlaps $F$ and $A$ the hat equations are trivially zero and 
\begin{equation}
    \hat{A} = 0 \,, \quad \hat{F} = 0 \,.
\end{equation}

The remaining three equations can be found as in \cite{Loureiro_2022}, where the only difference lies in the dependence of \(f_g\) on \(\hat{P}\) and \(\hat{N}\). Note that here and above we denote by $z^*$ the value of the proximal at any given point of integration.
\begin{equation}
    \hat{V} = -\alpha \mathbb{E}_{\xi}\left[
        \int_{\RR} \dd{y} \mathcal{Z}_0 \partial_\omega f_g
    \right] \,, \quad 
    \hat{q} = \alpha \mathbb{E}_{\xi}\left[
        \int_{\RR} \dd{y} \mathcal{Z}_0 f_g^2
    \right] \,, \quad 
    \hat{m} = \alpha \mathbb{E}_{\xi}\left[
        \int_{\RR} \dd{y} \partial_\omega \mathcal{Z}_0 f_g
    \right] \,,
\end{equation}

\subsubsection{The Prior Saddle-Point Equations}

For the prior saddle-point equations our starting point is 
\begin{equation}
\begin{aligned}
    \Psi_w &= \frac{1}{2 d} \tr \qty[
        \left(\hat{m}^2 \Sigmax^{\top} \boldsymbol{\theta}_0 \boldsymbol{\theta}_0^{\top} \Sigmax  + \hat{m}\hat{F} \Sigmax^{\top} \boldsymbol{\theta}_0 \boldsymbol{\theta}_0^{\top} \Sigmaupsilon + \hat{m}\hat{F} (\Sigmaupsilon)^{\top} \boldsymbol{\theta}_0 \boldsymbol{\theta}_0^{\top} \Sigmax + \hat{F}^2 (\Sigmaupsilon)^{\top} \boldsymbol{\theta}_0 \boldsymbol{\theta}_0^{\top} \Sigmaupsilon + \hat{q}  \Sigmax \right) \boldsymbol{\Lambda}^{-1}
    ]  
\end{aligned}
\end{equation}
where for simplicity of notation we define 
\(\boldsymbol{\Lambda} = \beta \lambda \Sigmaw + \hat{V} \Sigmax + \hat{P} \Sigmadelta + \hat{A} \Sigmaupsilon + \hat{N} \mathbb{1}\) 
and we will use 
\(\boldsymbol{H} = \hat{m}^2 \Sigmax^{\top} \boldsymbol{\theta}_0 \boldsymbol{\theta}_0^{\top} \Sigmax + \hat{m}\hat{F} \Sigmax^{\top} \boldsymbol{\theta}_0 \boldsymbol{\theta}_0^{\top} \Sigmaupsilon + \hat{m}\hat{F} (\Sigmaupsilon)^{\top} \boldsymbol{\theta}_0 \boldsymbol{\theta}_0^{\top} \Sigmax + \hat{F}^2 (\Sigmaupsilon)^{\top} \boldsymbol{\theta}_0 \boldsymbol{\theta}_0^{\top} \Sigmaupsilon + \hat{q}  \Sigmax\).

As the channel equations for $\hat{A}$ and $\hat{F}$ are trivially zero, we want to start with these derivatives as the following expressions will simplify considerably.
\begin{equation}
    \begin{aligned}
        A &= \partial_{\hat{A}} \Psi_w = \frac{1}{d} \tr \qty[
        \boldsymbol{H} \Sigmaupsilon \boldsymbol{\Lambda}^{-2}
    ] \\
     F & = \partial_{\hat{F}} \Psi_w = \frac{1}{d} \tr \qty[
        \hat{m} \left( \Sigmax^{\top} \boldsymbol{\theta}_0 \boldsymbol{\theta}_0^{\top} \Sigmaupsilon\right) \boldsymbol{\Lambda}^{-1}
    ] \\
    \end{aligned}
\end{equation}

We want to compute a few derivatives of the term \(\Psi_w\) to obtain equations for the overlap \(P\).

We begin with the hat-variable
\begin{equation}
    P = \partial_{\hat{P}} \Psi_w = \frac{1}{d} \tr \qty[
        \boldsymbol{H} \Sigmadelta \boldsymbol{\Lambda}^{-2}
    ]
\end{equation}
Again in the case of the FGM we will have an additional equation that is very similar to the one of \(P\).

As before the derivative w.r.t. \(\hat{q}, \hat{a}, \hat{n}\) follow from previous literature as
\begin{equation}
    V = \partial_{\hat{q}} \Psi_w = \frac{1}{d} \tr \qty[
        \Sigmax \boldsymbol{\Lambda}^{-1}
    ] \,, \quad 
    q = \partial_{\hat{V}} \Psi_w = \frac{1}{d} \tr \qty[
        \boldsymbol{H} \Sigmax \boldsymbol{\Lambda}^{-2}
    ] \,, \quad
    m = \partial_{\hat{m}} \Psi_w = \frac{1}{d} \tr \qty[ \hat{m} \Sigmax^{\top} \boldsymbol{\theta}_0 \boldsymbol{\theta}_0^{\top} \Sigmax \boldsymbol{\Lambda}^{-1} ] \,.
\end{equation}

Note that for the numerical evaluation $\frac{1}{d} \tr$ is just the mean of the eigenspectrum.

\subsection{Final set of saddle point equations for \texorpdfstring{\(\ell_2\)}{L2} regularisation}

We state here our final set of saddle point equations for reference
\begin{equation}\label{eq:final-saddle-point-eqs-appendix}
\begin{aligned}
    &\begin{cases}
        \hat{m} = \alpha \mathbb{E}_{\xi}\left[
            \int_{\RR} \dd{y} \partial_\omega \mathcal{Z}_0 f_g(\sqrt{q} \xi, P, N, \advtrainingcost)
        \right] \\
        \hat{q} = \alpha \mathbb{E}_{\xi}\left[
            \int_{\RR} \dd{y} \mathcal{Z}_0 f_g^2(\sqrt{q} \xi, P, N, \advtrainingcost)
        \right] \\
        \hat{V} = -\alpha \mathbb{E}_{\xi}\left[
            \int_{\RR} \dd{y} \mathcal{Z}_0 \partial_\omega f_g(\sqrt{q} \xi, P, N, \advtrainingcost)
        \right] \\
        \hat{P} = \advtrainingcost \frac{1}{2 \sqrt{P}} \alpha \mathbb{E}_{\xi} \qty[ 
            \int_{\RR} \dd{y} \mathcal{Z}_0 y f_g (\sqrt{q} \xi, P, N, \advtrainingcost)
        ] 
    \end{cases} \\
    &\begin{cases}
        m = \frac{1}{d} \tr \qty[ 
            \hat{m} \Sigmax^{\top} \boldsymbol{\theta}_0 \boldsymbol{\theta}_0^{\top} \Sigmax \boldsymbol{\Lambda}^{-1} 
        ] \\
        q = \frac{1}{d} \tr \qty[
            \qty( \hat{m}^2 \Sigmax^{\top} \boldsymbol{\theta}_0 \boldsymbol{\theta}_0^{\top} \Sigmax + \hat{q}  \Sigmax ) \Sigmax \boldsymbol{\Lambda}^{-2}
        ] \\
        V = \frac{1}{d} \tr \qty[
            \Sigmax \boldsymbol{\Lambda}^{-1}
        ] \\
        P = \frac{1}{d} \tr \qty[
            \qty( \hat{m}^2 \Sigmax^{\top} \boldsymbol{\theta}_0 \boldsymbol{\theta}_0^{\top} \Sigmax + \hat{q}  \Sigmax ) \Sigmadelta \boldsymbol{\Lambda}^{-2}
        ] 
    \end{cases}
\end{aligned}
\end{equation}

where we remember the definitions of \(\boldsymbol{\Lambda} = \lambda \Sigmaw + \hat{V} \Sigmax + \hat{P} \Sigmadelta\). Note \(\hat{A}, \hat{F}\) are exactly zero because the channel is not dependent on \(A, F\). The values for \(A\) and \(F\) becomes
\begin{equation}
    A = \frac{1}{d} \tr \qty[
        \qty( \hat{m}^2 \Sigmax^{\top} \boldsymbol{\theta}_0 \boldsymbol{\theta}_0^{\top} \Sigmax + \hat{q}  \Sigmax ) \Sigmaupsilon \boldsymbol{\Lambda}^{-2}
    ]\,, \quad
    F = \frac{1}{d} \tr \qty[
        \hat{m} \left( \Sigmax^{\top} \boldsymbol{\theta}_0 \boldsymbol{\theta}_0^{\top} \Sigmaupsilon\right) \boldsymbol{\Lambda}^{-1}
    ] \,.
\end{equation}
where we remind that \(f_g\) is defined in \cref{eq:moreau-proximal-definitions}.

\end{document}